\documentclass[acmtog]{acmart}

\acmSubmissionID{1988}
\copyrightyear{2025}
\acmYear{2025}
\setcopyright{cc}
\setcctype{by}
\acmConference[SA Conference Papers '25]{SIGGRAPH Asia 2025 Conference Papers}{December 15--18, 2025}{Hong Kong, Hong Kong}
\acmBooktitle{SIGGRAPH Asia 2025 Conference Papers (SA Conference Papers '25), December 15--18, 2025, Hong Kong, Hong Kong}\acmDOI{10.1145/3757377.3763957}
\acmISBN{979-8-4007-2137-3/2025/12}

\usepackage{xcolor} 
\usepackage{colortbl} 

\newcommand{\boldparagraph}[1]{\paragraph{#1}}

\makeatletter
\DeclareRobustCommand\onedot{\futurelet\@let@token\@onedot}
\def\@onedot{\ifx\@let@token.\else.\null\fi\xspace}
\def\eg{e.g\onedot}

\makeatother

\makeatletter
\DeclareRobustCommand\onedot{\futurelet\@let@token\@onedot}
\def\@onedot{\ifx\@let@token.\else.\null\fi\xspace}
\def\eg{e.g\onedot}

\makeatother

\usepackage[table]{xcolor} % Load FIRST
\usepackage{xcolor}
\usepackage{arydshln}      % Load SECOND
\usepackage{booktabs}      % Keep, but avoid mixing with \hdashline
\usepackage{multirow}
\usepackage{graphicx}
\usepackage{url}
\usepackage{caption}
\usepackage{subcaption}
\usepackage{xspace}

\definecolor{colorfirst}{RGB}{255,153,153}
\definecolor{colorsecond}{RGB}{255,204,153}
\definecolor{colorthird}{RGB}{255,255,153}

\newcommand{\colorfirst}{255,153,153}
\newcommand{\colorsecond}{255,204,153}
\newcommand{\colorthird}{255,255,153}

\begin{document}

% \title{Neural Localized Triplane: A Universal Enhancer for 3D Gaussian Splatting}
% \title{3D Neural Triplane Splatting: Expressive 3D Gaussian Splatting for View Synthesis, Geometry, and Dynamic Reconstruction
\title{Neural Texture Splatting: Expressive 3D Gaussian Splatting for View Synthesis, Geometry, and Dynamic Reconstruction
}

\author{Yiming Wang}
\email{wangyim@ethz.ch}
\affiliation{
\institution{ETH Zurich}%\looseness=-1}
\country{Switzerland}
}
\author{Shaofei Wang}
\email{sfwang0928@gmail.com}
\affiliation{
\institution{ETH Zurich}%\looseness=-1}
\country{Switzerland}
}
\author{Marko Mihajlovic}
\email{marko.mihajlovic@inf.ethz.ch}
\affiliation{
\institution{ETH Zurich}%\looseness=-1}
\country{Switzerland}
}
\author{Siyu Tang}
\email{siyu.tang@inf.ethz.ch}
\affiliation{
\institution{ETH Zurich}%\looseness=-1}
\country{Switzerland}
}

\begin{abstract}
3D Gaussian Splatting (3DGS) has emerged as a leading approach for high-quality novel view synthesis, with numerous variants extending its applicability to a broad spectrum of 3D and 4D scene reconstruction tasks.
Despite its success, the representational capacity of 3DGS remains limited by the use of 3D Gaussian kernels to model local variations. 
Recent works have proposed to augment 3DGS with additional per-primitive capacity, such as per-splat textures, to enhance its expressiveness. 
However, these per-splat texture approaches primarily target dense novel view synthesis with a reduced number of Gaussian primitives, and their effectiveness tends to diminish when applied to more general reconstruction scenarios.
In this paper, we aim to achieve concrete performance improvement over state-of-the-art 3DGS variants across a wide range of reconstruction tasks, including novel view synthesis, geometry and dynamic reconstruction, under both sparse and dense input settings.
To this end, we introduce Neural Texture Splatting (NTS).
At the core of our approach is a global neural field (represented as a hybrid of a tri-plane and a neural decoder) that predicts local appearance and geometric fields for each primitive.
By leveraging this shared global representation that models local texture fields across primitives, we significantly reduce model size and facilitate efficient global information exchange, demonstrating strong generalization across tasks. 
Furthermore, our neural modeling of local texture fields introduces expressive view- and time-dependent effects, a critical aspect that existing methods fail to account for.
Extensive experiments show that Neural Texture Splatting consistently improves models and achieves state-of-the-art results across multiple benchmarks. 
Project website: \url{https://19reborn.github.io/nts/}.
\end{abstract}

% \keywords{character animation, human-object interaction}

\begin{teaserfigure}
  \includegraphics[width=\textwidth]{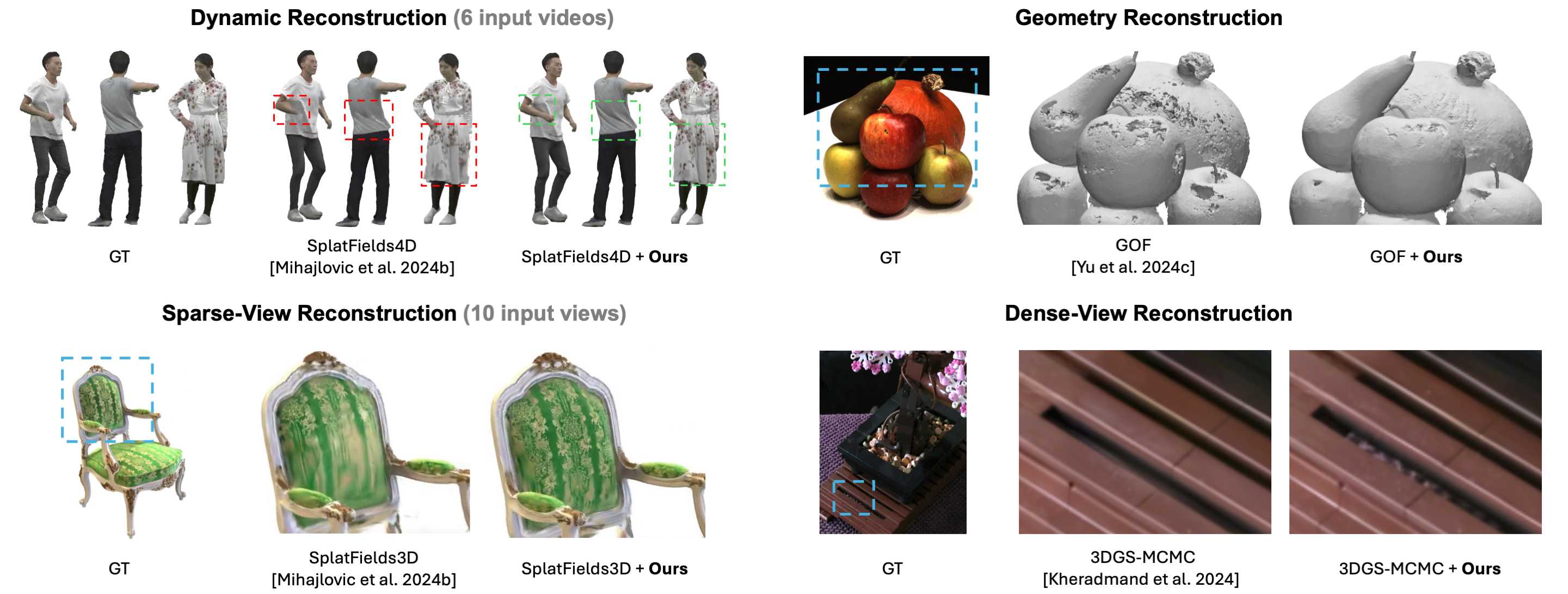}
  \vspace{-1.5em}
  \caption{\textbf{Neural Texture Splatting (NTS)} augments 3D Gaussian Splatting (3DGS) with neural RGBA fields per primitive, offering a plug-and-play module that boosts performance across multiple applications. 
  We show that NTS significantly enhances the quality of sparse-view reconstruction for static and dynamic scenes, while also achieving noticeable improvements on geometry reconstruction and
    novel view synthesis under dense training views.}
  \Description{Neural Texture Splatting (NTS)}
  \label{fig:teaser}
\end{teaserfigure}

\maketitle

\section{Introduction}
\label{sec:intro}
% Radiance field representations~\cite{mildenhall-ECCV2021-nerf,kerbl20233dgs} have demonstrated success in reconstructing static scenes from calibrated RGB captures.
% Radiance Fields has been widely explored as an leading representations~\cite{mildenhall-ECCV2021-nerf,kerbl20233dgs} have demonstrated success in reconstructing scenes from calibrated RGB captures.
Neural radiance fields (NeRFs~\cite{mildenhall2021nerf}) have demonstrated remarkable success in reconstructing scenes from calibrated RGB captures. They have been widely applied across various 3D reconstruction tasks, showing promise in applications such as AR/VR, robotics, gaming, and 3D content creation \cite{tewari2022advances}.
% , \yiming{more..} SLAM~\cite{Zhu2022nice_slam}. 
%
Recently, 3D Gaussian Splatting (3DGS)~\cite{kerbl20233dgs} has emerged as an efficient alternative to NeRFs by representing radiance fields as collections of 3D Gaussian primitives. 
This explicit, point-based representation enables both high-quality and real-time rendering.  It has also been extended to tasks such as surface reconstruction~\cite{Huang2DGS2024, Yu2024GOF}, sparse-view reconstruction~\cite{SplatFields}, and dynamic scene modeling~\cite{Wu2024CVPR, sun20243dgstream, lei2024mosca, zhu2024motiongs, qian20243dgs}.

However, the representational capacity of each primitive in 3DGS is inherently limited by the simplistic assumption of Gaussian kernels, making it difficult to capture high-frequency details and complex variations in appearance and geometry. Recent works have sought to enhance this capacity by incorporating per-splat 2D texture mappings~\cite{chao2024textured, rong2024gstex} or modifying the Gaussian kernel distribution~\cite{Li2025CVPR, Zhu2025CVPR}, showing promising improvements in novel view synthesis through increased local expressiveness.

Despite these advances, such methods are typically designed for novel view synthesis using dense training views and a reduced number of Gaussian primitives, while
% fail to consider their effectiveness on
forgoing other tasks such as sparse-view reconstruction, dynamic scene modeling, and geometry reconstruction.
For instance, our experiments show that a straightforward per-splat texture mapping implementation similar to ~\cite{chao2024textured} does not yield consistent improvements beyond the dense view setup.
%This modeling strategy
In general, directly adding per-splat texture mapping suffers from two fundamental limitations.
\emph{(1)} The texture mapping lacks view- and time-dependent variations, which are essential for capturing specular reflections in novel view synthesis and temporal appearance variations in dynamic scene modeling. This shortcoming also hinders surface reconstruction, as the model may resort to incorrect geometry to compensate for insufficient view-dependent appearance modeling. 
\emph{(2)} Locally optimizing per-primitive texture mappings can lead to redundant capacity and poor spatial consistency across neighboring splats, thereby increasing the risk of overfitting, as evidenced by the geometry reconstruction results in Table~\ref{tab:overfit}. This issue becomes particularly detrimental for both static and dynamic reconstruction under sparse-view settings, where global information is essential for robust optimization and improved generalization ability~\cite{SplatFields}.

To this end, we introduce Neural Texture Splatting (NTS), a unified framework that enhances 3DGS-based methods across a broad range of tasks. NTS augments each Gaussian primitive with local 2D or 3D RGBA texture fields to capture complex geometric and appearance variations. More importantly, we take inspiration from~\cite{SplatFields} and leverage the implicit regularization of neural networks to enforce global consistency across primitives, thereby improving robustness and generalization across tasks, while also enabling efficient modeling of view- and time-dependent spatial variations within each splat. 
Specifically, we utilize a global tri-plane~\cite{Peng2020ECCV,Chan2022CVPR} and a neural network to decode each Gaussian primitive's local RGBA texture fields based on its position. The neural network can take both viewing direction and timestep as inputs, hence producing view/time-dependent effects for local RGBA texture fields.  During rendering, we adopt the ray-Gaussian intersection formula from \cite{Yu2024GOF} to compute the 3D contribution point of each primitive, which is then used to query the local texture fields for additional color and opacity added to the original Gaussian splatting form of volume rendering. 

Our contributions can be summarized as follows:
\begin{itemize}
    % \item We introduce \textbf{3D Neural Triplane Splatting (3DNTS)}, a novel representation that enahcens 3D Gaussian Splatting splats neural per-primitive expressive triplane radiance field, 
    \item  We introduce Neural Texture Splatting (NTS), a novel representation that augments 3D Gaussian Splatting with neural per-primitive RGBA texture fields, serving as a robust representation that enhances 3DGS-based methods across a wide range of tasks.

    \item We propose to leverage a global tri-plane network to predict localized texture fields, enabling expressive modeling of geometry and appearance while ensuring robustness across tasks.

    \item Extensive experiments demonstrate that our representation consistently improves existing state-of-the-art methods across diverse benchmarks, including dense and sparse-view novel view synthesis, surface reconstruction, and dynamic reconstruction.
\end{itemize}
    % \item We propose a global triplane network to predict localized triplane radiance fields combined with the acorrding splatting algorithm, enabling expressive modeling of color and opacity variations within each Gaussian primitive while using neural network to ensure  robustness across different setup.

\section{Related Works}
\subsection{Novel View Synthesis}
Recent advances in novel view synthesis (NVS) from calibrated multi-view images have introduced powerful scene representations, including multi-plane images~\cite{flynn2016deepstereo,zhou2018rel10k,broxton2020immersive}, voxel grids~\cite{sun2022direct,muller2022instantngp,reiser2024binary}, and neural fields~\cite{mildenhall2021nerf,sitzmann2019scene}. Among these, 3D Gaussian Splatting (3DGS)~\cite{kerbl20233dgs} stands out as an efficient volumetric rendering approach that combines real-time rasterization with high-quality synthesis. Its flexibility has enabled extensions to surface~\cite{Huang2DGS2024, Yu2024GOF}, sparse~\cite{SplatFields}, and dynamic scene reconstruction~\cite{yang2024deformable, wu20244dgaussian}. 

\subsection{Splatting Beyond Gaussians}
While 3DGS achieves remarkable efficiency and quality, its representational power is fundamentally limited by the fixed Gaussian kernel shape per splat. This restricts the model’s ability to encode high-frequency details, fine geometry, and complex view-dependent effects. To address these limitations, recent works explore augmenting Gaussians with per-splat texture maps \cite{rong2024gstex, chao2024textured, xu2024supergaussians, svitov2024billboard, huang2024texturedGS,song2024hdgstextured2dgaussian, zhang2025nest}, introducing local 2D fields that modulate color and opacity. Texture-GS \cite{xu2024texture} employs an explicit cubemap texture and a global 3D-to-2D mapping to achieve appearance editing.
Others propose modifying the Gaussian kernel itself to better match surface geometry or appearance distributions \cite{Li2025CVPR, Zhu2025CVPR}.  

Although these methods improve expressiveness for dense-input NVS, they often suffer from overfitting and lack robustness when generalized to sparse-view inputs, surface reconstruction, or dynamic scenes. In contrast, our Neural Texture Splatting (NTS) offers a unified and modular enhancement: we predict per-splat RGBA texture fields using a shared global neural representation, allowing for efficient, view- and time-conditioned modeling. This approach substantially improves expressiveness and generalization, making NTS a plug-and-play upgrade to existing 3DGS pipelines across diverse tasks. 

\subsection{Surface Reconstruction}
While the vanilla NeRF can be used to reconstruct surfaces, its resulting geometry is often noisy due to the lack of explicit regularization.  A number of works have been proposed to improve the geometry reconstruction of NeRF, mainly by replacing the underlying density field with a geometrically regularized field, such as an occupancy field~\cite{Oechsle2021ICCV} or a signed distance field (SDF)~\cite{Wang2021NEURIPS,Yariv2021NEURIPS,Yariv2023SIGGRAPH,Li2023CVPR,Wang2023ICCV,Alexandru2023CVPR}.
With the advent of 3DGS, a number of works have also been proposed to extend it to high-quality surface reconstruction.  Initial efforts have focused on jointly optimizing a neural SDF with 3DGS, using the learned SDF to regularize the geometry of 3DGS~\cite{Chen2023ARXIV,Yu2024NEURIPS}.  While effective, these methods suffer from slow optimization and excessive triangle counts when extracting the surface mesh.
More recent works optimize only the Gaussian splatting parameters to maintain efficiency, while introducing additional regularization terms~\cite{Huang2DGS2024,Guedon2024CVPR,Turkulainen2024WACV}, novel re-parameterization of Gaussians~\cite{Huang2DGS2024,Dai2024SIGGRAPH,Yu2024SIGGRAPHASIA}, or advanced mesh extraction algorithms~\cite{Guedon2024CVPR,Dai2024SIGGRAPH,Huang2DGS2024,Yu2024SIGGRAPHASIA} to achieve faithful surface reconstruction.
Our proposed approach is compatible with these 3DGS surface reconstruction methods.  We demonstrate the effectiveness of our approach by building on top of one of the latest methods~\cite{Yu2024SIGGRAPHASIA}.

\subsection{Sparse-view Static 3DGS Reconstruction}
Recovering high-quality geometry and appearance from a small number of input views remains a major limitation for 3DGS, which typically relies on optimizing a large number of independent splats. This independence requires dense observations to constrain each primitive effectively, often leading to overfitting or poor generalization in sparse-view settings. Recent approaches tackle this issue by incorporating deep learned priors~\cite{charatan2024pixelsplat, chen2024mvsplat, chung2024depthreg, zhang2025transplat,tang2024mv}. In contrast, SplatFields~\cite{SplatFields} avoids the sensitivity of such priors and introduces neural spatial regularization by regressing splat features from their 3D spatial positions. This induces a spatial autocorrelation bias that significantly improves performance in sparse-view scenarios. Our Neural Texture Splatting integrates seamlessly with SplatFields and further enhances reconstruction quality, as demonstrated in our experiments.

\subsection{Sparse-view Dynamic 3DGS Reconstruction} 
Recent methods have extended Gaussian splatting to dynamic scenes by learning time-varying deformation fields with MLPs~\cite{Wu_2024_CVPR, yang2023deformable3dgs, yu2024cogs, lu20243d}. However, these approaches typically rely on densely sampled views from varying camera positions to fully capture scene dynamics—an assumption that is rarely practical in real-world scenarios, where fixed camera rigs are commonly used to observe a scene from limited viewpoints. For a comprehensive overview of this growing area, we refer the reader to the recent survey on dynamic 3DGS~\cite{fan2025advances}.

To support dynamic reconstruction under sparse-view conditions, we build on SplatFields4D~\cite{SplatFields}, which employs the expressive ResFields architecture~\cite{mihajlovic2024ResFields} to model scene deformation from limited inputs. Our Neural Texture Splatting is fully compatible with this setup and further improves reconstruction quality, as shown in our experiments.

\begin{table}[!t]
  \centering
  % \footnotesize
  \small
  \setlength{\tabcolsep}{2pt} % Adjust column spacing if needed
  \caption{\textbf{Overfitting Issue for Per-primitive Texture Baseline.} We use GOF~\cite{Yu2024GOF} as the backbone. Results on the DTU~\cite{aanaes2016DTU} dataset show that using a per-primitive texture representation without our proposed global neural texture is more prone to overfitting on training views, as evidenced by worse test Chamfer Distance (CD). In contrast, our global neural texture mitigates this issue and enhances generalization.
}
  \begin{tabular}{l| cc cc}
    \toprule
    \multirow{2}{*}{Method} 
    & \multicolumn{2}{c}{Scan 24} 
    & \multicolumn{2}{c}{Scan 105} \\
    & Train PSNR$\uparrow$  & Test CD$\downarrow$  
    & Train PSNR$\uparrow$ & Test CD$\downarrow$ \\
    \midrule
    W/o Neural Texture 
    & \textbf{36.05} & 0.48 
    & \textbf{39.42} & 0.70 \\
    W/ Neural Texture
    & 34.73 & \textbf{0.44} 
    & 38.99 & \textbf{0.64} \\
    \bottomrule
  \end{tabular}

  % \vspace{-2.0em}
  \label{tab:overfit}
\end{table}

\section{Background}
3D Gaussian Splatting (3DGS)~\cite{kerbl20233dgs} represents scenes with a set
of 3D Gaussian primitives, where each Gaussian primitive is defined by a 3D
covariance matrix $\mathbf{\Sigma} \in \mathbb{R}^{3\times3} $ and center
position $\mathbf{\mu} \in \mathbb{R}^3$.  To ensure positive semi-definiteness,
the covariance matrix is factorized into the rotation matrix $\mathbf{R} \in
\mathbb{R}^{3\times3}$ and the scale matrix $\mathbf{S} \in
\mathbb{R}^{3\times3}$ as  $\mathbf{\Sigma} =
\mathbf{R}\mathbf{S}\mathbf{S}^\top\mathbf{R}^\top$ .  The 3D Gaussian kernel of the $k$-th primitive, denoted as $\mathcal{G}_k$, is defined as:
\begin{equation}
    \label{eqn:3d_gaussian}
    \begin{aligned}
        \mathcal{G}_k(\mathbf{x}) &= \text{exp}(-\frac{1}{2}(\mathbf{x} - \mathbf{\mu}_k)^T\mathbf{\Sigma}_k^{-1}(\mathbf{x} - \mathbf{\mu}_k)) \\
        \text{s.t.} &\quad \mathbf{\Sigma}_k = \mathbf{R}_k\mathbf{S}_k\mathbf{S}_k^\top\mathbf{R}_k^\top
    \end{aligned}
\end{equation}

%
%%%%%%%%%%%%%%%%%%%%%%%%%%%%%%%%%%%%%
%
\begin{figure*}[t]
	\centering
	\includegraphics[width=0.9\linewidth]{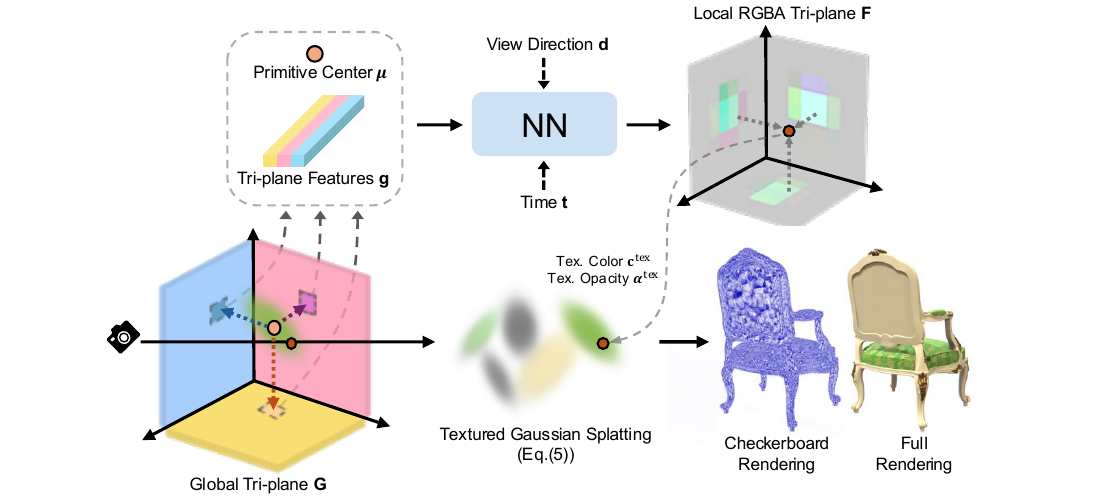}
	\vspace{-10pt}
	\caption
	{
            \textbf{Method Overview.} 
            Our method enhance 3D Gaussian Splatting (3DGS) by introducing a local RGBA tri-plane texture to each splat (top right). During rendering with our proposed textured Gaussian Splatting (bottom, Sec.~\ref{subsec:textured_gaussian_splatting}), each ray computes the intersection point with splats and queries the corresponding RGBA textures, which are combined with the original Gaussian attributes to produce the final rendering result via volume rendering (Eq.\eqref{eqn:textured_vol_rendering}). We show checkerboard and learned RGB texture renderings for visualization.  
            While this local texture field improves the representational capacity of color and opacity, it also increases the risk of overfitting and lacks the ability to capture view- and time-dependent variations. To address these limitations, we introduce a global tri-plane neural field that models the local texture fields in a compact and shared manner (top, Sec.~\ref{subsec:neural_global_encoding}).
            }
	\label{fig:overview}
 \vspace{-6pt}
\end{figure*}
%
%%%%%%%%%%%%%%%%%%%%%%%%%%%%%%%%%%%%%
%

%
% consists of center position $\mathbf{\mu}_k$, scaling
% $\mathbf{S}_k$, rotation $\mathbf{R}_k$, color $\mathbf{c}_k \in
% \mathbb{R}^{3}$, and opacity $\alpha_k \in [0, 1]$. 
To differentially render
the color of a pixel $\mathbf{p}\in \mathbb{R}^2$, the colors associated with
Gaussian primitives are alpha-composited from front to back, following the
conventional volume rendering equation:
\begin{equation}
    \label{eqn:vol_rendering}
    \mathbf{c}(\mathbf{p}) = \sum_{k=1}^K \mathbf{c}_k \alpha_k \mathcal{P}(\mathcal{G}_k, \mathbf{p}) \prod_{j=1}^{k-1}(1 - \alpha_k \mathcal{P}(\mathcal{G}_k, \mathbf{p}))
\end{equation}
where $\mathbf{c}_k \in \mathbb{R}^{3}$, $\alpha_k \in [0, 1]$ are the view-dependent color and
opacity of the primitive, respectively.  $\mathcal{P}(\mathcal{G}_k,
\mathbf{p})$ evaluates the 3D Gaussian kernel $\mathcal{G}_k$ to the camera ray
corresponding to the pixel $\mathbf{p}$.  This evaluation can be done by either
EWA splatting~\cite{zwicker2002ewa}, exact projection of 2D
surfels~\cite{Huang2DGS2024}, or finding the ray-primitive
intersection~\cite{Yu2024SIGGRAPHASIA}.
% where $\alpha_i$ is the opacity associated with each Gaussian, and $\mathcal{G}^{\text{2D}}_i$ is the 2D Gaussian value evaluated at pixel location $\mathbf{p}$.
% \begin{equation}
% \label{eqn:alpha_gaussian}
%     \alpha_i = \mathcal{G}^{\text{2D}}_i(\mathbf{p})\cdot o_i
% \end{equation}
The attributes of the Gaussians, including their center position, rotation,
scale, opacity, and color (modeled using spherical harmonic coefficients to
account for view-dependent effects), are optimized through gradient descent by
minimizing photometric losses computed on the rendered 2D images.

\section{Methods}
In this section, we describe our method in detail (illustrated in Fig.~\ref{fig:overview}).

\subsection{Textured Gaussian Splatting}
\label{subsec:textured_gaussian_splatting}
The expressiveness of the 3DGS primitive (hereafter referred to as \textit{splat}) is
inherently limited, as it models color and opacity distributions using a single
3D Gaussian kernel.  However, real-world scenarios often exhibit significantly
more complex spatial variations in both geometry and appearance, such as high-frequency view-dependent effects.
In this work, we introduce a learnable RGBA texture field for each splat, complementing the original Gaussian kernel $\mathcal{G}_k$, to more effectively capture  detailed local variations in appearance and geometry.

We define the RGBA texture field $\mathcal{F}$ in the local coordinate system of each splat.
Given a world space point $\mathbf{\hat{x}}$, its
corresponding point $\mathbf{x}_k = \{
    x_k, y_k, z_k \}$ in the local coordinate of the $k$th splat is calculated using the splat's center position $\boldsymbol{\mu}_k$, scale matrix $\mathbf{S}_k$ and rotation matrix $\mathbf{R}_k$ as
\begin{equation}
    \mathbf{x}_k = \mathbf{S}^{-1}_k \mathbf{R}_k ( \mathbf{\hat{x}} - \boldsymbol{\mu}_k)
\end{equation}
The RGBA texture values of $\mathbf{c}^{\text{tex}}_k(\mathbf{\hat{x}})$ and $\alpha^{\text{tex}}_k(\mathbf{\hat{x}})$, for the world point $\mathbf{\hat{x}}$, are queried from the texture field $\mathcal{F}$ using the local coordinate point $\mathbf{x}_k$ as
\begin{equation}
    \mathbf{c}^{\text{tex}}_k(\mathbf{\hat{x}}), \mathbf{\alpha}^{\text{tex}}_k(\mathbf{\hat{x}}) = \mathcal{F}(\mathbf{x}_k) 
\end{equation}

We interpret the queried RGBA texture values $\mathbf{c}^{\text{tex}}_k(\hat{\mathbf{x}})$ and $\alpha^{\text{tex}}_k(\hat{\mathbf{x}})$ as approximations of the integrated contributions from the entire local texture field during volume rendering. 
These contributions are added to the original color and opacity terms in the standard rendering formulation of conventional Gaussian Splatting (Eq.\eqref{eqn:vol_rendering}), enhancing its expressiveness. Accordingly, we define the modified volume rendering equation for our Textured Gaussian Splatting as follows:
\begin{equation}
\begin{aligned}
\label{eqn:textured_vol_rendering}
\mathbf{c}(\mathbf{p}) = \sum_{k=1}^K \; &(\mathbf{c}_k + \mathbf{c}^{\text{tex}}_k(\hat{\mathbf{x}}_k)) 
(\alpha_k \mathcal{P}(\mathcal{G}_k, \mathbf{p}) + \alpha^{\text{tex}}_k(\hat{\mathbf{x}}_k))\cdot \\
& \prod_{j=1}^{k-1} \left(1 - \alpha_j \mathcal{P}(\mathcal{G}_j, \mathbf{p}) - \alpha^{\text{tex}}_j(\hat{\mathbf{x}}_j) \right)
\end{aligned}
\end{equation}
% where $\hat{\mathbf{x}}_k$ denotes the intersection point between the rendering camera ray and the $k$-th Gaussian primitive.
Here, $\hat{\mathbf{x}}_k$ denotes the query point used to evaluate the RGBA texture field $\mathcal{F}$ for the $k$-th Gaussian primitive. 
The opacity term in volume rendering, 
$\alpha_k \mathcal{P}(\mathcal{G}_k, \mathbf{p}) + \alpha^{\text{tex}}_k(\hat{\mathbf{x}}_k)$, 
is clamped to $[0,0.99]$ to ensure that the resulting transmittance remains non-negative.
We adopt the ray–Gaussian intersection algorithm in~\cite{Yu2024SIGGRAPHASIA} to obtain the query point for the RGBA texture field. 
Since the integrated texture value is primarily learned through the RGBA texture fields, our representation remains compatible with alternative ray–Gaussian intersection algorithms~\cite{huang2024texturedGS, chao2024textured}.

\boldparagraph{Modeling the RGBA Texture Field}
% Instead of using a continuous field, such as a small MLP~\cite{reiser2021kilonerf} to represent each local texture field implicitly, we adopt an explicit representation for the texture field to improve efficiency.
Instead of employing a continuous implicit representation, such as a small MLP~\cite{reiser2021kilonerf}, for each local texture field, we adopt an explicit representation to improve computational efficiency.
We use a fixed resolution $\tau$ to discretize the RGBA texture field $\mathcal{F}$. While a straightforward approach would be to represent it using a $\tau \times \tau \times \tau$ voxel grid, we found this to be memory-inefficient, especially at higher resolutions.
%, as it requires allocating a dense grid for each primitive.
This motivates us to employ a tri-plane representation~\cite{Chen2022ECCV,Peng2020ECCV,Chan2022CVPR} to compress the RGBA volume. 

Specifically, the RGBA volume of the $k$-th splat is represented by three orthogonal learnable planes, $\mathbf{F}_{xy}^k$, $\mathbf{F}_{xz}^k$, and $\mathbf{F}_{yz}^k$, each in $\mathbb{R}^{\tau \times \tau \times 4}$. Each plane has a resolution of $\tau \times \tau$ and stores 4-channel RGBA values.
Given a point in the
splat's local coordinate $\mathbf{x}_k = \{ x_k, y_k, z_k \}$, its corresponding
local color and opacity values can be obtained by querying the three planes:
\begin{equation}
\label{eqn:local_triplane}
\begin{aligned}
\centering
    \mathbf{f}_{xy}^k (\mathbf{x}_k) &= \mathbf{F}_{xy}^k(x_k, y_k) \\
    \mathbf{f}_{xz}^k (\mathbf{x}_k) &= \mathbf{F}_{xz}^k(x_k, z_k) \\
    \mathbf{f}_{yz}^k (\mathbf{x}_k) &= \mathbf{F}_{yz}^k(y_k, z_k) \\
    \mathbf{c}_k^{\text{tex}} (\mathbf{x}_k), \alpha_k^{\text{tex}} (\mathbf{x}_k) &= \frac{1}{3}(\mathbf{f}_{xy}^k (\mathbf{x}_k) + \mathbf{f}_{xz}^k (\mathbf{x}_k) + \mathbf{f}_{yz}^k (\mathbf{x}_k))
\end{aligned}
\end{equation}

% \begin{equation}
% \label{eqn:local_triplane}
% \begin{aligned}
% \mathbf{f}_{xy}^k (\mathbf{x}_k) &= \mathbf{F}_{xy}^k(x_k, y_k), \quad
% \mathbf{f}_{xz}^k (\mathbf{x}_k) = \mathbf{F}_{xz}^k(x_k, z_k), \quad
% \mathbf{f}_{yz}^k (\mathbf{x}_k) = \mathbf{F}_{yz}^k(y_k, z_k), \\
% \mathbf{c}_k^{\text{tex}} (\mathbf{x}_k), \alpha_k^{\text{tex}} (\mathbf{x}_k) &= \frac{1}{3}\left(\mathbf{f}_{xy}^k (\mathbf{x}_k) + \mathbf{f}_{xz}^k (\mathbf{x}_k) + \mathbf{f}_{yz}^k (\mathbf{x}_k)\right).
% \end{aligned}
% \end{equation}

%
We employ bilinear interpolation when sampling from each of the three planes and discard query points outside a local bounding box of size 3. 
An alternative approach, which approximates the texture field using a 2D RGBA texture, is described in the supplementary material.
%

%
% We also explore a simplified approximation of the RGBA volume using a single 2D plane, aligned with the splat's principal axis. 
% As shown in Tab.~\ref{tab:dense_view_blender}, when combined with our Neural Texture method (introduced in the next section), both representations improve upon the baseline.
% We adopt the 3D tri-plane representation for modeling the RGBA texture field throughout this paper.

%%%%%%
% \input{Figures/fig_merged_curve_wrt_views}
% \input{Figures/fig_dense_view_mipnerf360_new}
% \input{Tables/sparse_view_owlii_4views}
% \input{Tables/Sparse_view_blender_mean}

\subsection{Neural Texture Splatting}
\label{subsec:neural_global_encoding}

While the local RGBA textures capture spatial variations of color and opacity within a splat, these variations are view-independent and time-independent, making them suboptimal for modeling complex view-dependent effects and temporal dynamics.
%
% A straightforward extension for incorporating view dependence is to replace the RGBA textures with textures of spherical harmonics (SH) coefficients per splat. However, this leads to a significant increase in parameter count (e.g., 9$\times$ more for three SH levels), which often results in unstable training and overfitting.
%
Furthermore, per-splat RGBA textures often introduce redundant representational capacity and lack spatial consistency across neighboring splats, resulting in reduced generalization and robustness. 
% We observe that directly optimizing these textures yields limited improvements across tasks: while they may enhance novel view synthesis, their effectiveness carry limitedly over to surface reconstruction or sparse-view reconstruction, as demonstrated in Table~\ref{tab:ablation}.
We observe that directly optimizing these textures yields limited improvements across tasks: although they result in noticeable gains for novel view synthesis, their effectiveness degrades significantly in other scenarios, such as surface reconstruction and sparse-view reconstruction, as demonstrated in Table~\ref{tab:overfit}.

To address this issue, we propose using a global neural field to model local texture fields in a compact and shared manner, enabling information sharing to achieve robust optimization.
We begin by leveraging a global tri-plane representation, denoted as $\mathbf{G}_{xy}, \mathbf{G}_{xz}, \mathbf{G}_{yz} \in \mathbb{R}^{H \times W \times C}$, to encode features $\mathbf{g}_k \in \mathbb{R}^C$ for the $k$-th splat. 
For each splat, the global tri-plane features $\mathbf{g}_k$ are extracted at its center position $\boldsymbol{\mu}_k = (\mu_x, \mu_y, \mu_z)$ as follows:

\vspace{-0.5em}

\begin{equation}
\begin{aligned}
\mathbf{g}_{xy}^k = \mathbf{G}_{xy}(\mu_x^k, \mu_y^k), \quad
\mathbf{g}_{xz}^k &= \mathbf{G}_{xz}(\mu_x^k, \mu_z^k), \quad
\mathbf{g}_{yz}^k = \mathbf{G}_{yz}(\mu_y^k, \mu_z^k), \\
\mathbf{g}_k &= \text{Concat}(\mathbf{g}_{xy}^k, \mathbf{g}_{xz}^k, \mathbf{g}_{yz}^k).
\end{aligned}
\end{equation}

The global tri-plane features $\mathbf{g}_k$ are concatenated with the splat's center position $\boldsymbol{\mu}_k$ and additional attributes, such as the viewing direction $\mathbf{d}_k$,
% pointing from the camera origin to the splat center
and an optional time step $\mathbf{t}$ for dynamic scenes. The combined feature is then passed through a shallow global neural network to predict the RGBA volume, represented by the local RGBA tri-planes textures $\mathbf{F}_{xy}^k$, $\mathbf{F}_{xz}^k$, and $\mathbf{F}_{yz}^k$ for the $k$-th splat.

\vspace{-0.5em}
\begin{equation}
\label{eqn:triplane_mlp}
\{ \mathbf{F}_{xy}^k, \mathbf{F}_{xz}^k, \mathbf{F}_{yz}^k \} = \mathrm{NN}(\mathbf{g}_k,  \boldsymbol{\mu}_k, \mathbf{d}_k, \mathbf{t})
\end{equation}
where $\mathbf{t}$ is included only for dynamic scenes and omitted for static ones.  
We employ distinct global triplanes $G$ and Neural decoders $\mathrm{NN}$ for color and opacity textures. The view direction serves as input only to the color network.

While the neural network could directly predict the RGBA tri-plane textures with a total dimension of $\tau \times \tau \times 3 \times 4$, we wish to further reduce computational cost and enable the use of higher-resolution local texture planes.  To this end, we utilize Canonical Polyadic (CP) decomposition to represent each of the three local planes.
%
%Instead of predicting the full-resolution textures directly,
Specifically, the network outputs 1D vectors that produce planes via outer products. 
For the $k$-th splat, the neural network predicts a pair of RGBA vectors $\mathbf{v}^{0,k}_{uv} \in \mathbb{R}^{4\tau}$ and $\mathbf{v}^{1,k}_{uv} \in \mathbb{R}^{4\tau}$ for each plane $uv \in \{xy, xz, yz\}$. Each local RGBA tri-plane texture is then computed as the outer product of the corresponding vector pair:
\begin{equation}
\begin{aligned}
\{\mathbf{v}_{xy}^{0,k}, \mathbf{v}_{xy}^{1,k}, \mathbf{v}_{xz}^{0,k}, \mathbf{v}_{xz}^{1,k}, \mathbf{v}_{yz}^{0,k}, \mathbf{v}_{yz}^{1,k} \} = \mathrm{NN}(\mathbf{g}_k, \mathbf{R}_k, \boldsymbol{\mu}_k, \mathbf{r}, t), \\
\mathbf{F}_{xy}^k = \mathbf{v}_{xy}^{0,k} \otimes \mathbf{v}_{xy}^{1,k}, \quad
\mathbf{F}_{xz}^k = \mathbf{v}_{xz}^{0,k} \otimes \mathbf{v}_{xz}^{1,k}, \quad
\mathbf{F}_{yz}^k = \mathbf{v}_{yz}^{0,k} \otimes \mathbf{v}_{yz}^{1,k}.
\label{eqn:cp_decompose}
\end{aligned}
\end{equation}

\subsection{Optimization}
\label{subsec:optimization}
We adopt the original optimization objective of 3DGS, \eg, a combination of L1 distance and D-SSIM
term as
\begin{equation} \label{eqn:rendering_loss}
    \mathcal{L}_c = \mathcal{L}_{\text{MSE}}(\hat{I}, I) + \lambda \mathcal{L}_{\text{D-SSIM}} (\hat{I}, I)
\end{equation}
where $\hat{I}$ denotes ground-truth target image and $I$ the corresponding
rendered target image. 
We set $\lambda=0.2$, and when integrating with other baselines such as GOF~\cite{Yu2024GOF}, we retain the auxiliary loss terms as defined in their original implementations.

% We adopt the same photometric rendering loss as used in 3D Gaussian Splatting, defined in Eqn.~\ref{eqn:rendering_loss}.
In addition, we introduce an L1 regularization term on the tri-plane RGBA textures to encourage sparsity:
\begin{equation}
\mathcal{L}_{\text{norm}} = \frac{1}{N \cdot \tau^2} \sum_{u,v} \left( \left\| \mathbf{F}_{xy}(u,v) \right\| + \left\| \mathbf{F}_{xz}(u,v) \right\|
+ \left\| \mathbf{F}_{yz}(u,v) \right\|
\right)
\label{eqn:sparsity_loss}
\end{equation}
where \(N\) is the number of splats, \(\tau\) denotes the resolution of each texture plane, and \((u,v)\) refers to the grid coordinates on the texture planes. $\mathcal{L}_{\text{norm}}$ is combined with $\mathcal{L}_c$ using a loss weight of 0.01.
A detailed ablation study of the components and hyperparameters is shown in Table~\ref{tab:ablation_components}.

\begin{table}[]
 \centering
  \scriptsize
  \setlength{\tabcolsep}{4pt}

\caption{
\textbf{Dynamic Reconstruction.} We report averaged metrics on the Owlii dataset with 4, 6, 8 and 10 training views. Detailed breakdowns are shown in Fig.~\ref{fig:owlii_sparse_compar_curve} and in the appendix tables. SSIM/LPIPS values are scaled by 100 for clarity. \colorbox{colorfirst}{1st}, \colorbox{colorsecond}{2nd} and \colorbox{colorthird}{3rd} denote the best results.
}
\resizebox{\linewidth}{!}{%
\begin{tabular}{@{}lccccc@{}}
\toprule
\multirow{2}{*}{Method} & \multicolumn{5}{c}{PSNR $\uparrow$} \\
& \textit{Mean} & Dancer & Exercise & Model & Basketball \\
\midrule
4D-GS & 23.27 & 23.92 & 23.27 & 24.11 & 21.78 \\
Deformable3DGS & 24.58 & 25.36 & 24.72 & 23.98 & 24.26 \\
4DGaussians & \cellcolor[RGB]{\colorthird}26.45 & \cellcolor[RGB]{\colorthird}25.86 & \cellcolor[RGB]{\colorthird}26.41 & \cellcolor[RGB]{\colorthird}27.38 & \cellcolor[RGB]{\colorthird}26.15 \\
SplatFields4D & \cellcolor[RGB]{\colorsecond}27.87 & \cellcolor[RGB]{\colorsecond}28.16 & \cellcolor[RGB]{\colorsecond}27.16 & \cellcolor[RGB]{\colorsecond}28.02 & \cellcolor[RGB]{\colorsecond}28.13 \\
SplatFields4D + \textbf{Ours} & \cellcolor{colorfirst}29.25 & \cellcolor{colorfirst}29.54 & \cellcolor{colorfirst}28.74 & \cellcolor{colorfirst}29.21 & \cellcolor{colorfirst}29.50 \\
\midrule
& \multicolumn{5}{c}{SSIM $\uparrow$} \\
& \textit{Mean} & Dancer & Exercise & Model & Basketball \\
\midrule
4D-GS & 88.81 & 89.96 & 88.92 & 87.87 & 88.48 \\
Deformable3DGS & 91.05 & 92.03 & 91.88 & 89.02 & 91.29 \\
4DGaussians & \cellcolor[RGB]{\colorthird}93.57 & \cellcolor[RGB]{\colorthird}93.38 & \cellcolor[RGB]{\colorthird}94.26 & \cellcolor[RGB]{\colorthird}92.79 & \cellcolor[RGB]{\colorthird}93.86 \\
SplatFields4D & \cellcolor[RGB]{\colorsecond}95.09 & \cellcolor[RGB]{\colorsecond}95.49 & \cellcolor[RGB]{\colorsecond}95.24 & \cellcolor[RGB]{\colorsecond}93.99 & \cellcolor[RGB]{\colorsecond}95.66 \\
SplatFields4D + \textbf{Ours} & \cellcolor{colorfirst}95.86 & \cellcolor{colorfirst}96.28 & \cellcolor{colorfirst}95.91 & \cellcolor{colorfirst}94.97 & \cellcolor{colorfirst}96.28 \\
\midrule
& \multicolumn{5}{c}{LPIPS $\downarrow$} \\
& \textit{Mean} & Dancer & Exercise & Model & Basketball \\
\midrule
SplatFields4D        &        7.18 &        6.78 &        7.74 &        7.84 &        6.38 \\
SplatFields4D + \textbf{Ours} & \textbf{6.70} & \textbf{6.10} & \textbf{7.42} & \textbf{7.17} & \textbf{6.08} \\
\bottomrule
\end{tabular}
}

 % \vspace{-12pt}
\label{tab:owlii_mean}
\end{table}

\begin{table}[t]
\centering
\scriptsize
\setlength{\tabcolsep}{2.5pt}
\caption{
\textbf{Sparse-view Reconstruction.} We report averaged metrics on the Blender dataset with 4, 6, 8 and 10 training views. Detailed breakdowns are shown in Fig.~\ref{fig:blender_sparse_compar_curve} and in the appendix tables. SSIM/LPIPS values are scaled by 100 for clarity. \colorbox[RGB]{\colorfirst}{1st}, \colorbox[RGB]{\colorsecond}{2nd} and \colorbox[RGB]{\colorthird}{3rd} denote the best results.
}
\resizebox{\linewidth}{!}{%
\begin{tabular}{@{}lccccccccc@{}}
\toprule
\multirow{2}{*}{Method} & \multicolumn{9}{c}{PSNR $\uparrow$} \\
& \textit{Mean} & Toy & Ficus & Hotdog & Chair & Mic & Ship & Drums & Materials \\
\midrule
SparseNeRF & 20.88 & \cellcolor[RGB]{\colorthird}22.47 & 17.96 & \cellcolor[RGB]{\colorthird}26.01 & 23.61 & 19.64 & \cellcolor[RGB]{\colorthird}20.16 & 16.76 & \cellcolor{colorfirst}20.42 \\
SuGaR & 19.43 & 20.39 & 21.16 & 21.69 & 22.14 & 18.07 & 18.22 & 17.34 & 16.47 \\
ScaffoldGS & 20.17 & 19.14 & 21.08 & 21.23 & 21.88 & 24.75 & 17.61 & 17.77 & 17.89 \\
Mip-Splatting & 20.75 & 20.29 & 22.40 & 21.55 & 22.04 & 24.75 & 18.24 & 18.49 & 18.23 \\
3DGS & 21.26 & 20.65 & 22.80 & 22.88 & 22.80 & \cellcolor[RGB]{\colorsecond}25.43 & 18.71 & 18.63 & 18.16 \\
LightGaussian & 21.41 & 20.95 & \cellcolor{colorfirst}23.85 & 22.41 & 22.99 & \cellcolor[RGB]{\colorthird}25.12 & 18.98 & \cellcolor[RGB]{\colorthird}18.81 & 18.21 \\
2DGS & \cellcolor[RGB]{\colorthird}21.44 & 20.49 & 22.61 & 24.80 & \cellcolor[RGB]{\colorthird}23.77 & 24.86 & 18.34 & 18.73 & 17.96 \\
SplatFields3D & \cellcolor[RGB]{\colorsecond}22.58 & \cellcolor[RGB]{\colorsecond}23.13 & \cellcolor[RGB]{\colorthird}23.00 & \cellcolor[RGB]{\colorsecond}26.66 & \cellcolor[RGB]{\colorsecond}23.98 & 25.11 & \cellcolor[RGB]{\colorsecond}20.36 & \cellcolor[RGB]{\colorsecond}19.57 & \cellcolor[RGB]{\colorthird}18.87 \\
SplatFields3D + \textbf{Ours} & \cellcolor{colorfirst}23.05 & \cellcolor{colorfirst}23.74 & \cellcolor[RGB]{\colorsecond}23.02 & \cellcolor{colorfirst}26.87 & \cellcolor{colorfirst}24.18 & \cellcolor{colorfirst}25.71 & \cellcolor{colorfirst}21.14 & \cellcolor{colorfirst}20.01 & \cellcolor[RGB]{\colorsecond}19.77 \\
\midrule
& \multicolumn{9}{c}{SSIM $\uparrow$} \\
& \textit{Mean} & Toy & Ficus & Hotdog & Chair & Mic & Ship & Drums & Materials \\
\midrule
SparseNeRF & 84.08 & \cellcolor[RGB]{\colorthird}85.30 & 82.95 & \cellcolor[RGB]{\colorthird}92.70 & 87.81 & 86.87 & \cellcolor[RGB]{\colorthird}73.76 & 78.76 & \cellcolor{colorfirst}84.52 \\
SuGaR & 80.51 & 79.01 & 86.27 & 87.51 & 85.50 & 84.05 & 69.86 & 76.92 & 74.92 \\
ScaffoldGS & 81.10 & 76.50 & 87.52 & 84.60 & 84.34 & 93.53 & 65.31 & 78.69 & 78.32 \\
Mip-Splatting & 83.75 & 79.44 & 89.81 & 88.02 & 86.75 & 94.55 & 68.37 & 82.68 & 80.38 \\
3DGS & 84.15 & 80.27 & 90.24 & 88.85 & 87.64 & \cellcolor[RGB]{\colorthird}94.71 & 68.40 & 83.09 & 79.95 \\
LightGaussian & 84.90 & 81.08 & \cellcolor{colorfirst}91.46 & 89.63 & 87.82 & \cellcolor[RGB]{\colorsecond}94.76 & 69.36 & 83.67 & 81.40 \\
2DGS & \cellcolor[RGB]{\colorthird}85.23 & 81.15 & 90.28 & 90.97 & \cellcolor[RGB]{\colorsecond}88.94 & 94.68 & 70.70 & \cellcolor[RGB]{\colorthird}84.35 & 80.78 \\
SplatFields3D & \cellcolor[RGB]{\colorsecond}87.04 & \cellcolor[RGB]{\colorsecond}85.41 & \cellcolor[RGB]{\colorthird}90.28 & \cellcolor[RGB]{\colorsecond}93.87 & \cellcolor[RGB]{\colorthird}88.86 & 94.46 & \cellcolor[RGB]{\colorsecond}75.89 & \cellcolor[RGB]{\colorsecond}85.51 & \cellcolor[RGB]{\colorthird}82.01 \\
SplatFields3D + \textbf{Ours} & \cellcolor{colorfirst}87.70 & \cellcolor{colorfirst}86.00 & \cellcolor[RGB]{\colorsecond}90.67 & \cellcolor{colorfirst}93.99 & \cellcolor{colorfirst}89.08 & \cellcolor{colorfirst}94.80 & \cellcolor{colorfirst}77.12 & \cellcolor{colorfirst}86.21 & \cellcolor[RGB]{\colorsecond}83.70 \\
\midrule
& \multicolumn{9}{c}{LPIPS $\downarrow$} \\
& \textit{Mean} & Toy & Ficus & Hotdog & Chair & Mic & Ship & Drums & Materials \\
\midrule
SplatFields3D        & \textbf{13.49} & \textbf{13.86} &        8.97 & \textbf{9.23} &        \textbf{11.63} &        5.62 &        \textbf{28.07 }&        13.06 &        17.44 \\
SplatFields3D + \textbf{Ours} &        13.50   &        14.03   & \textbf{8.22} &        9.55   & 11.89 & \textbf{5.46} & 28.75 & \textbf{12.92} & \textbf{17.20} \\
\bottomrule
\end{tabular}
}

\label{tab:sparse_view_mean}

 % \vspace{-12pt}
\end{table}

\section{Experiments}
In this section, we demonstrate the effectiveness of our approach by integrating it into several state-of-the-art 3D Gaussian Splatting (3DGS) variants and evaluating its performance across a diverse set of reconstruction tasks. 
% \yiming{a bit lenghty and repeating here}
% In this section, we demonstrate the effectiveness of our method by integrating it into several state-of-the-art 3D Gaussian Splatting (3DGS) variants, including GOF, 3DGS-MCMC, and SplatFields. We evaluate performance across a wide range of reconstruction tasks, such as dense-view novel view synthesis, surface reconstruction, and sparse-view reconstruction for both static and dynamic scenes. Our approach serves as a plug-and-play module that consistently improves reconstruction quality across different 3DGS pipelines~\yiming{add sec ref}.
% In this section, we integrate our representation into state-of-the-art 3D Gaussian Splatting (3DGS) variants, GOF, 3DGS-MCMC, SpaltFields across various reconstruction tasks, including dense-view novel view synthesis, surface reconstruction, and sparse-view static and dynamic scene reconstruction. This demonstrates the plug-and-play capability of our method and its effectiveness in improving performance across different 3DGS pipelines~\yiming{add sec ref}.
% Our representation can be integrated into 3DGS-based models to enhance performance across different reconstruction tasks. We conduct evaulation on both surface reconstruction and novel view synthesis under diverse scene configurations.

\subsection{Setup}
% \subsection{{Setup}
\boldparagraph{Datasets}  
Our evaluation spans multiple standard benchmarks. 
% Blender~\cite{mildenhall2021nerf} and Mip-NeRF360~\cite{barron2022mipnerf360} are used to assess dense-view novel view synthesis on both synthetic and real-world scenes. Sparse-view performance is measured following the SplatFields~\cite{SplatFields} protocol, utilizing Blender for static 3D scenes and Owlii~\cite{xu2017owlii} for dynamic 4D sequences. Surface reconstruction quality is benchmarked using the DTU dataset~\cite{aanaes2016DTU}.
We first benchmark sparse-view reconstruction on the Blender dataset~\cite{mildenhall2021nerf} and the Owlii dataset~\cite{xu2017owlii} for static and dynamic scenes, respectively. We then evaluate surface reconstruction on the DTU dataset~\cite{aanaes2016DTU} and dense-view novel view synthesis on the Blender~\cite{mildenhall2021nerf} and Mip-NeRF360~\cite{barron2022mipnerf360} datasets.  We use standard metrics for novel view synthesis, including PSNR, SSIM~\cite{SSIM}, and LPIPS~\cite{zhang2018lpips}. 
For surface reconstruction, we measure geometric accuracy using Chamfer Distance on the extracted mesh, following~\cite{Huang2DGS2024, Yu2024GOF}.

\boldparagraph{Backbone choice}
We adopt GOF~\cite{Yu2024GOF} and 3DGS-MCMC~\cite{kheradmand20243dgc_mcmc} for dense-view reconstruction tasks, and employ SplatFields~\cite{SplatFields} as our backbone model for sparse-view scenarios. Each of these methods demonstrates state-of-the-art performance in different domains: GOF excels in surface reconstruction, 3DGS-MCMC in dense-view novel view synthesis, and SplatFields in sparse-view 3D reconstruction and dynamic 4D scene modeling. 
\subsection{Sparse-view Scene Reconstruction}
% We demonstrate that our model effectively increases the model expressiveness while keeping generalization ability under the sparse-view setting, significantly improving the current state-of-the-art method~\cite{SplatFields} for sparse-view reconstruction of both static and dynamic scenes.
% Please refer to the supplementary document and video for the complete benchmark results and additional qualitative visualizations.
We demonstrate that our model significantly enhances representational expressiveness while maintaining strong generalization under sparse-view settings, outperforming the current state-of-the-art method~\cite{SplatFields} in reconstructing both static and dynamic scenes.
Comprehensive benchmark results and additional qualitative visualizations are provided in the supplementary document and accompanying video.

\begin{figure*}
    \centering
    \includegraphics[width=1.0\linewidth]{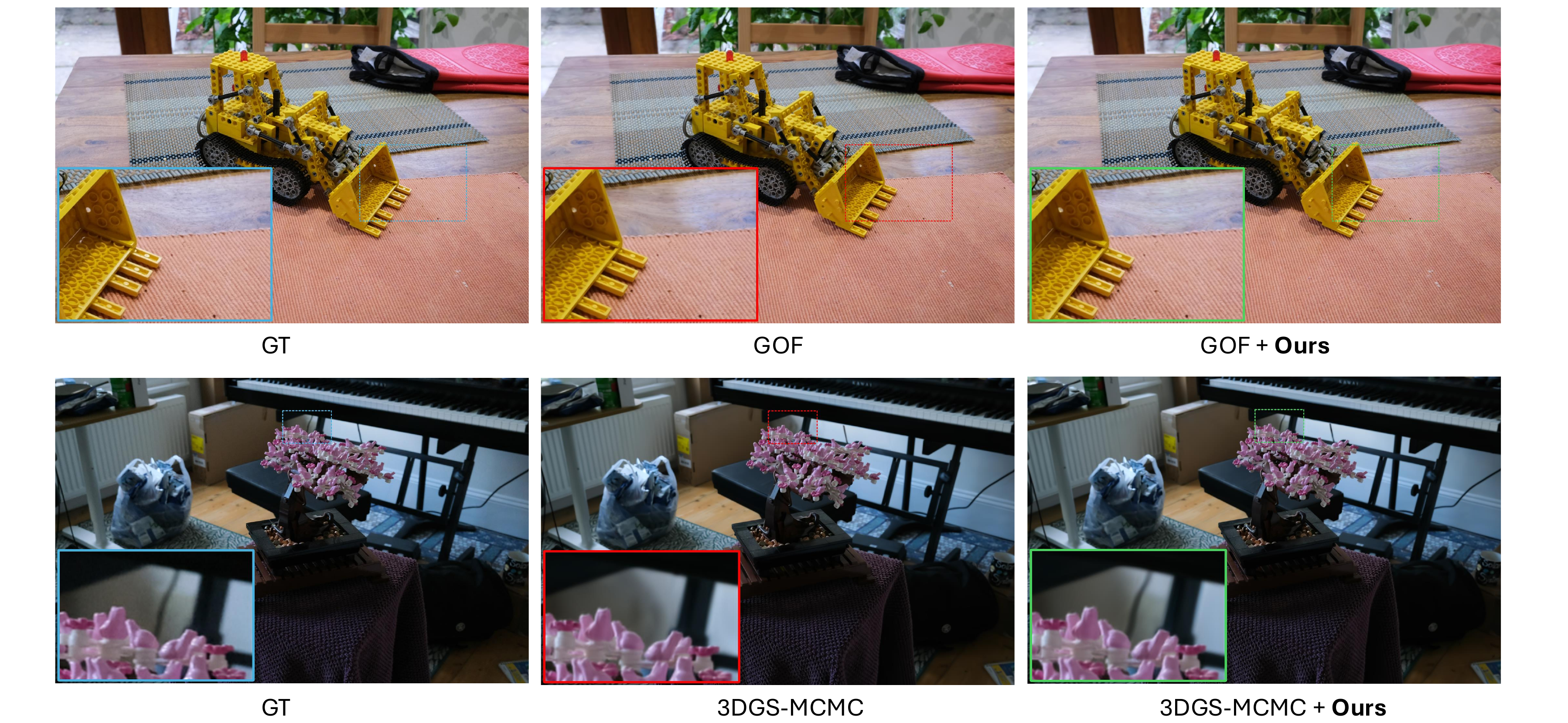}
     \vspace{-1.0em}
    \caption{\textbf{Qualitative comparison of novel view synthesis on the MipNeRF360 dataset.} Our method enhances baselines by better handling view-dependent effects and preserving fine-grained structures.}
    \label{fig:mipnerf360}
    \vspace{-0.5em}
\end{figure*}
\begin{figure*}[t]
\centering
\begin{minipage}{0.48\textwidth}
    \centering
	\includegraphics[width=1.0\linewidth]{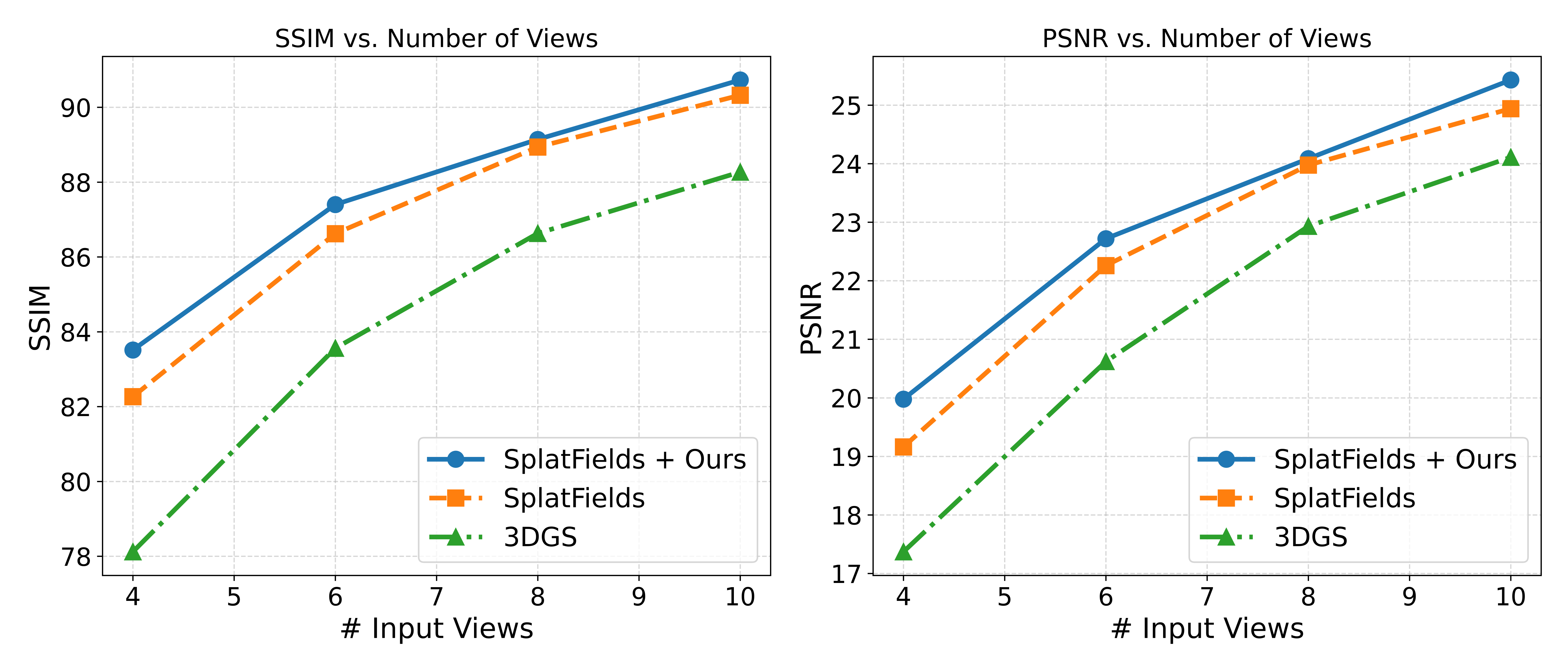}
	\vspace{-16pt}
	\caption
	{
\textbf{Average PSNR and SSIM curves with varying input views on Blender under the sparse-view setup.} Our method consistently improves upon SplatFields across input views ranging from 4 to 10.
}
\label{fig:blender_sparse_compar_curve}
\end{minipage}
\hfill
\begin{minipage}{0.48\textwidth}
    \centering
	\includegraphics[width=1.0\linewidth]{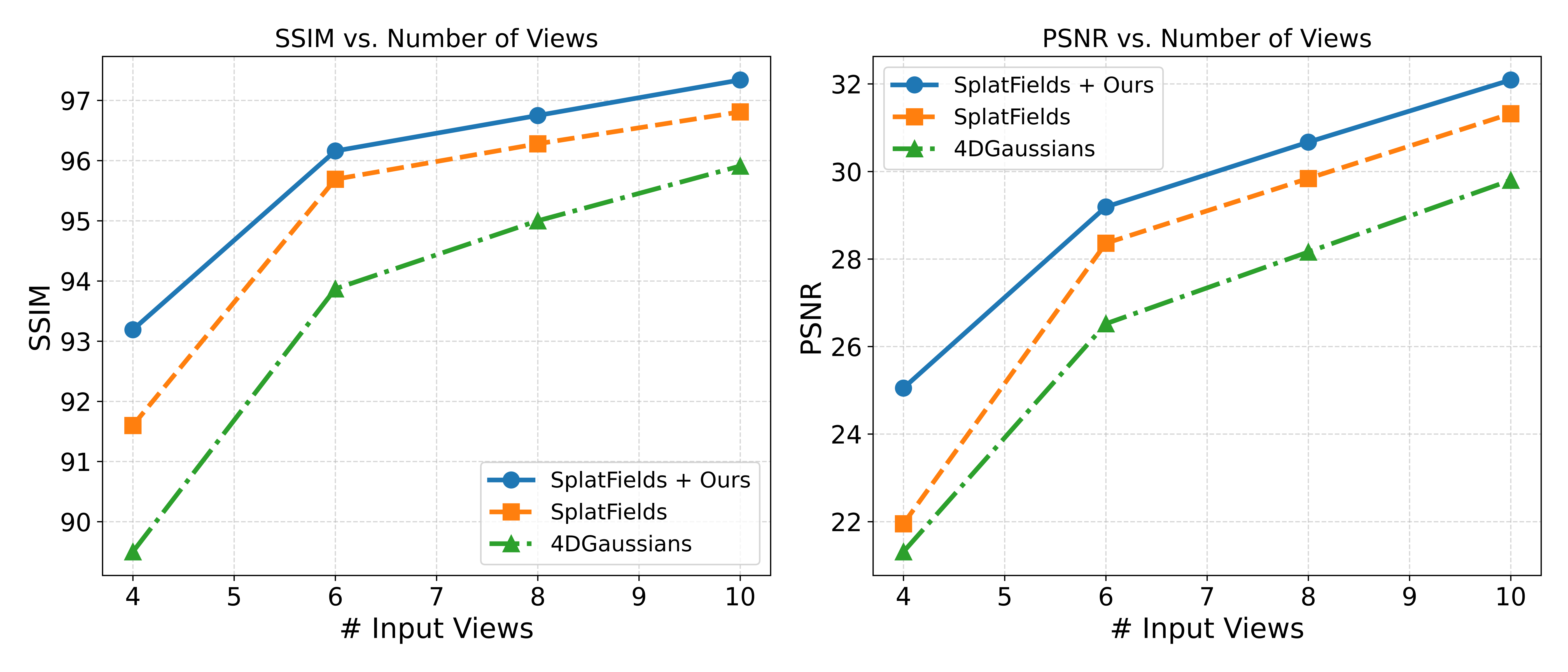}
	\vspace{-16pt}
	\caption
	{
\textbf{Average PSNR and SSIM curves with varying input views on Owlii under the sparse-view dynamic setup.} Our method consistently improves upon SplatFields across input views ranging from 4 to 10.
	}
\label{fig:owlii_sparse_compar_curve}
\end{minipage}
% \caption{Comparison of two different viewpoints.}

% \label{fig:two_views}

\vspace{-0.5em}
\end{figure*}
\begin{figure*}
    \centering
    \includegraphics[width=0.95\linewidth]{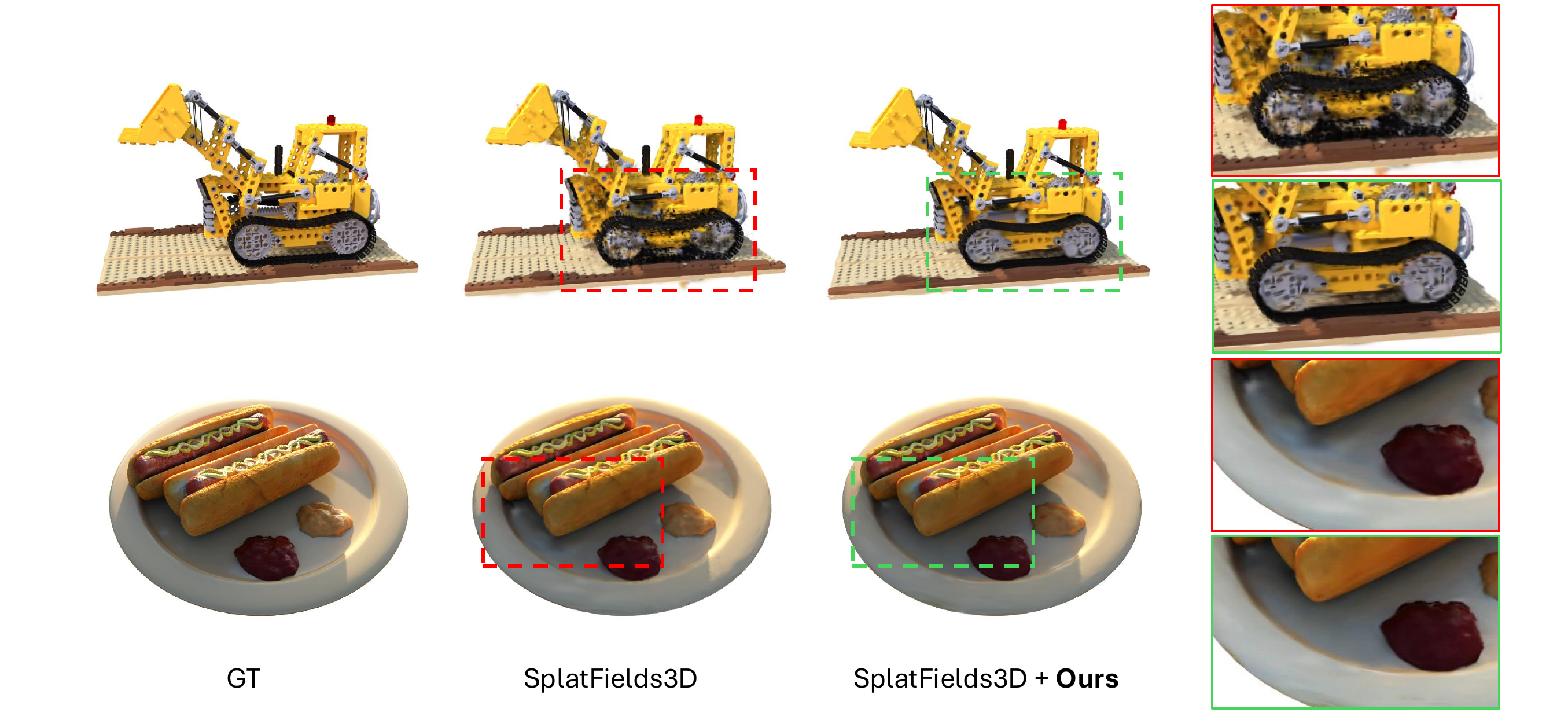}
    \vspace{-1.0em}\caption{\textbf{Qualitative results for sparse-view reconstruction on Blender using 10 input views.} Our method yields sharper reconstructions with more accurate geometry.}
    \label{fig:blender_sparse}
\end{figure*}

\begin{table}[!t]
\centering
\footnotesize
\setlength{\tabcolsep}{3.5pt}

\renewcommand{\arraystretch}{1.0}  % 30% more vertical space
\caption{\textbf{Dense-view Reconstruction.}
Our method consistently improves the performance of both Gaussian Opacity Field (GOF) and 3DGS-MCMC across novel view synthesis and surface reconstruction tasks.
}
\begin{tabular}{lccc ccc c}
\toprule
\multirow{2}{*}{Method} & \multicolumn{3}{c}{Blender} & \multicolumn{3}{c}{MipNeRF360} & \multicolumn{1}{c}{DTU} \\
 & PSNR$\uparrow$ & SSIM$\uparrow$ & LPIPS$\downarrow$ & PSNR$\uparrow$ & SSIM$\uparrow$ & LPIPS$\downarrow$ & CD$\downarrow$ \\
\midrule
MipNeRF360 & 30.34 & 0.9510 & 0.0600 & 27.69 & 0.7920 & 0.2370 & -------- \\
Instant-NGP & 32.20 & 0.9590 & 0.0550 & 25.30 & 0.6710 & 0.3710 & -------- \\
3DGS & 33.08 & 0.9671 & 0.0440 & 27.26 & \cellcolor[RGB]{\colorthird}0.8318 & 0.1871 & -------- \\
% 3DGS-MCMC & \cellcolor[RGB]{\colorsecond}33.81 & \cellcolor[RGB]{\colorfirst}0.9718 & \cellcolor[RGB]{\colorfirst}0.0284 & \cellcolor[RGB]{\colorfirst}28.14 & \cellcolor[RGB]{\colorfirst}0.8427 & \cellcolor[RGB]{\colorfirst}0.1755 & -------- \\
TexturedGaussians & 33.24 & 0.9674 & 0.0428 & 27.35 & 0.8274 & \cellcolor[RGB]{\colorthird}0.1858 & -------- \\
2DGS & 33.08 & 0.9680 & 0.0332 & 26.81 & 0.7967 & 0.2520 & \cellcolor[RGB]{\colorthird}0.80 \\
% \hdashline
% \hline
GOF & 33.44 & 0.9698 & 0.0307 & 27.45 & 0.8272 & 0.1945 & \cellcolor[RGB]{\colorsecond}0.74 \\
GOF + \textbf{Ours} & \cellcolor[RGB]{\colorfirst}34.09 & \cellcolor[RGB]{\colorthird}0.9710 & 
\cellcolor[RGB]{\colorthird}0.0286 & 
\cellcolor[RGB]{\colorthird}27.71 &
0.8257 & 
0.1964 & 
\cellcolor[RGB]{\colorfirst}0.67 \\
3DGS-MCMC & \cellcolor[RGB]{\colorthird}33.81 &
\cellcolor[RGB]{\colorsecond}0.9718 & 
\cellcolor[RGB]{\colorsecond}0.0284 & 
\cellcolor[RGB]{\colorsecond}28.14 & 
\cellcolor[RGB]{\colorsecond}0.8427 & 
\cellcolor[RGB]{\colorsecond}0.1755 & -------- \\
3DGS-MCMC + \textbf{Ours} & \cellcolor[RGB]{\colorsecond}34.06
& \cellcolor[RGB]{\colorfirst}0.9724 & 
\cellcolor[RGB]{\colorfirst}0.0275 & 
\cellcolor[RGB]{\colorfirst}28.24 & 
\cellcolor[RGB]{\colorfirst}0.8429 & 
\cellcolor[RGB]{\colorfirst}0.1748 & -------- \\
% \hdashline
% \textbf{+ Ours} (Mean)  & 32.20 & 0.9590 & 0.0550 & 25.30 & 0.6710 & 0.3710 & + 0.07 \\
\bottomrule
\end{tabular}

% \vspace{-2.5em}
\label{tab:dense_view_all}
\end{table}
\boldparagraph{Static Scene} We compare our method against recent 3DGS~\cite{kerbl20233dgs,Huang2DGS2024,Yu2023MipSplatting,fan2023lightgaussian,scaffoldgs,guedon2024sugar,SplatFields} and NeRF-based~\cite{wang2023sparsenerf} approaches for sparse-view static scene reconstruction. Following the SplatFields protocol, we conduct experiments on the Blender dataset~\cite{mildenhall2021nerf}, varying the number of training views from 4 to 10. 
% Both SplatFields and our method are trained for 40k iterations using identical configurations. Our model is initialized from the SplatFields checkpoint at 20k iterations, followed by 20k iterations of joint training. 
%
Table~\ref{tab:sparse_view_mean} presents a comprehensive comparison averaged over the 4, 6, 8, and 10-view settings, where our method achieves the best performance among the baselines.
As shown in Fig.~\ref{fig:blender_sparse_compar_curve}, which illustrates the performance of SplatFields, SplatFields+Ours, and 3DGS with respect to the number of input views, our method consistently outperforms the state-of-the-art SplatFields across varying view counts. 
Qualitative comparison between ours and SplatFields under the 10-view input setting are presented in Fig.~\ref{fig:blender_sparse}, where our method produces sharper reconstructions with fewer floater artifacts.
% and improved structural consistency compared to the baseline.

%
%
% To further illustrate the trend across varying view counts, we present a performance curve in Figure~\ref{fig:blender_sparse_compar_curve}, comparing SplatFields, SplatFields+Ours, and 3DGS. Our method demonstrates more stable improvements as the number of input views increases, with significant gains in both PSNR and perceptual quality in the low-view regime. This highlights the strength of our approach in leveraging sparse observations to reconstruct accurate scene geometry and appearance.
%
%

\boldparagraph{Dynamic Scene}
To evaluate dynamic scene reconstruction under real-world conditions, we adopt four sequences from the Owlii dataset~\cite{xu2017owlii} as benchmarks. We use 100 frames for training and 100 test frames captured from varying viewpoints along a rotating camera trajectory. We compare our method with recent dynamic 3D Gaussian Splatting approaches -- 4D-GS~\cite{yang2023gs4d}, Deformable3DGS~\cite{yang2024deformable}, and 4DGaussians~\cite{wu20244dgaussian} -- under varying numbers of input views, ranging from 4 to 10. Complete benchmark results and additional comparisons with dynamic NeRF methods~\cite{li2022neural3DV, pumarola2021dnerf, park2021nerfies, park2021hypernerf} are provided in the supplementary material.
% Our approach builds upon the state-of-the-art method SplatFields~\cite{SplatFields}. 

% To ensure a fair comparison, both SplatFields and our method are trained for 200k iterations. Specifically, we initialize our model from the SplatFields checkpoint at 100k iterations and continue with an additional 100k iterations of joint training.

Quantitative results averaged over the 4, 6, 8, and 10-view input settings are reported in Table~\ref{tab:owlii_mean}. Fig.~\ref{fig:owlii_sparse_compar_curve} further shows performance trends as the number of input views increases from 4 to 10, comparing SplatFields, SplatFields+Ours, and the second-best baseline, 4DGaussians.
Across all view counts, our method consistently achieves superior performance. 

\begin{figure*}
    
    \centering
    \includegraphics[width=0.95\linewidth]{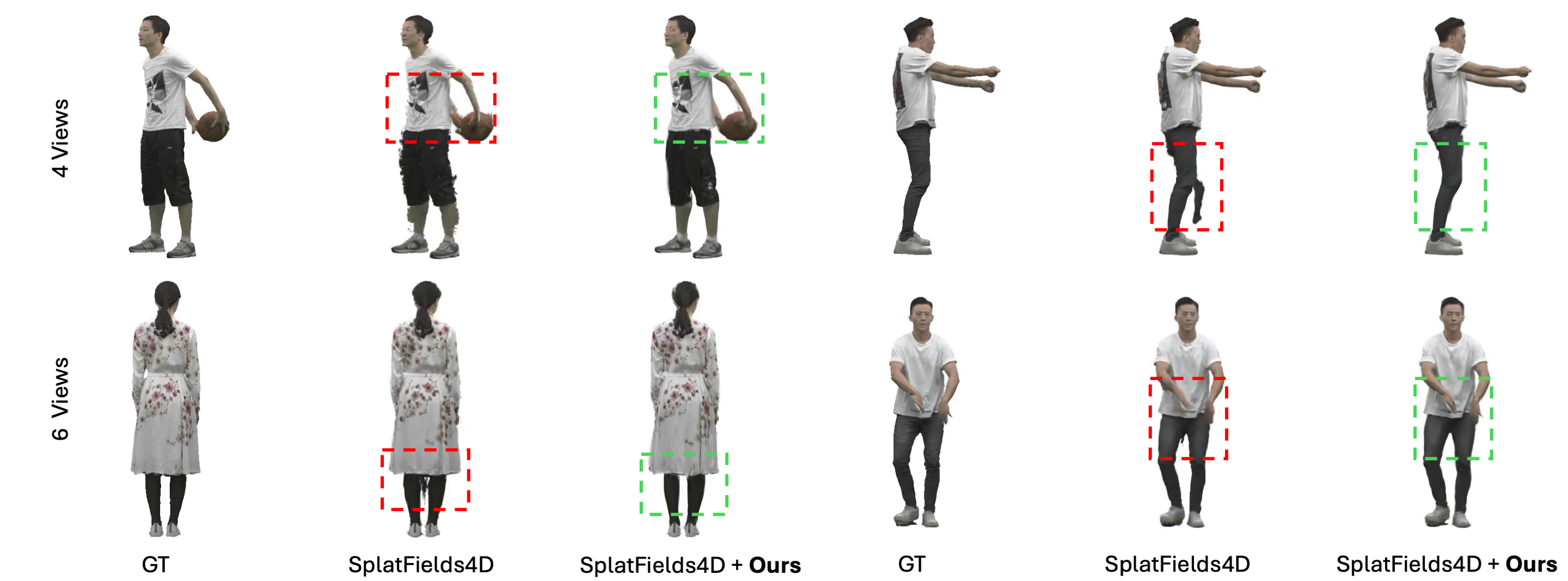}
    \vspace{-1.0em}
    \caption{\textbf{Qualitative results for sparse-view dynamic reconstruction on Owlii using 4 and 6 input views.} Our method effectively reduces floaters and inaccurate boundaries compared to SplatFields.}
    \label{fig:owlii_4views}
    \vspace{-1.0em}
    % \vspace{-0.5em}
\end{figure*}

\begin{figure*}
    
    \centering
    \includegraphics[width=0.95\linewidth]{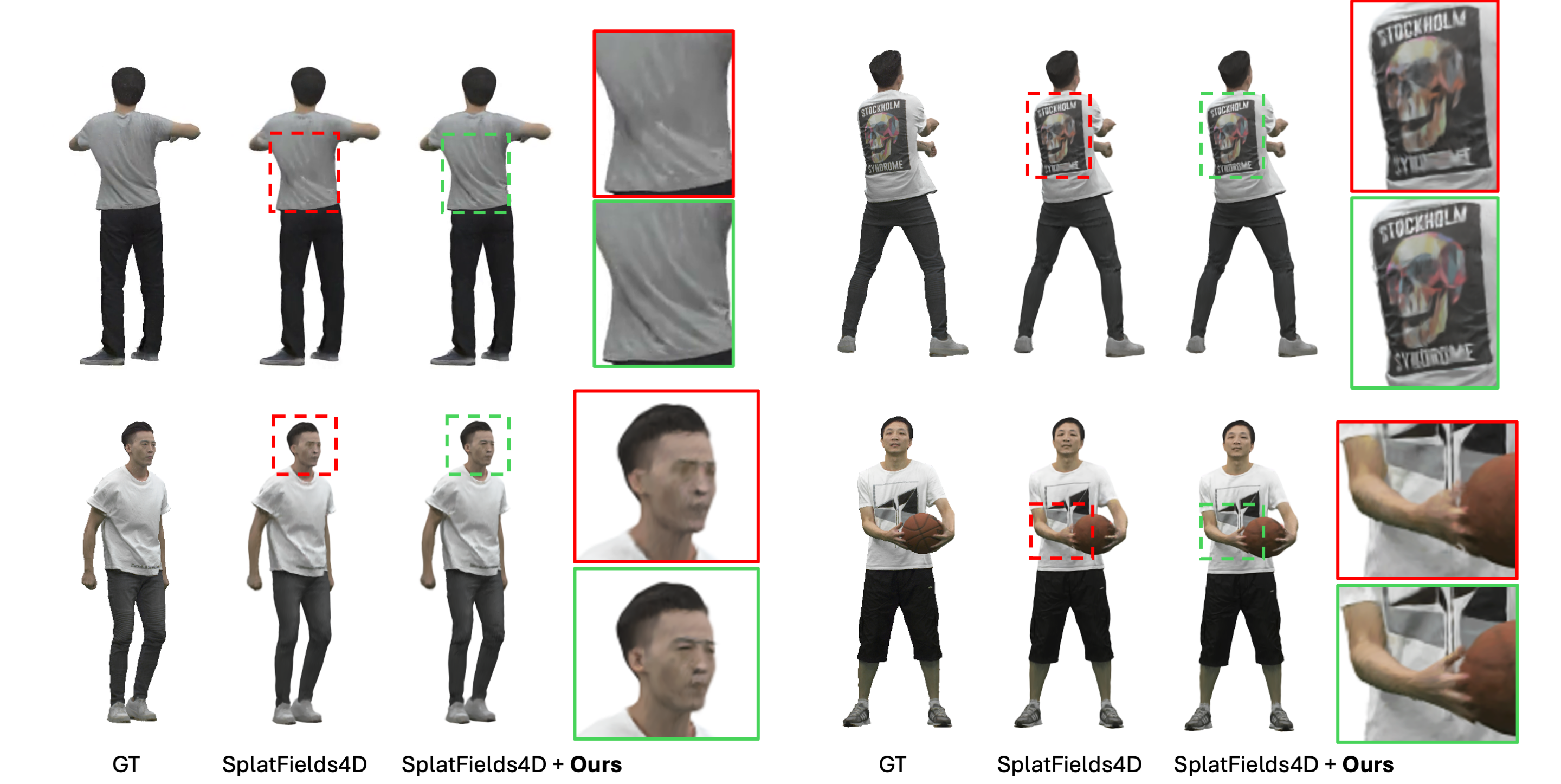}
    \vspace{-1.0em}
    \caption{\textbf{Qualitative results for sparse-view dynamic reconstruction on Owlii using 10 input views.} Our method can capture finer details compared to SplatFields.}
    \label{fig:owlii_10views}
    \vspace{-1.0em}
    % \vspace{-0.5em}
\end{figure*}

Fig.~\ref{fig:owlii_4views} demonstrates that under the challenging sparse-view setup, our method effectively mitigates floaters and boundary artifacts. With 10 input views, as illustrated in Fig.~\ref{fig:owlii_10views}, our approach yields qualitatively superior reconstructions, recovering fine-grained details and producing sharper, more geometrically faithful structures, particularly in challenging regions such as human faces and hands.

% As shown in Fig.~\ref{fig:owlii_4views}, our method alleviates floaters and boundary artifacts under the challenging sparse-view setup.
% Fig.~\ref{fig:owlii_10views} illustrates qualitative results with 10 input views, highlighting the advantages of our approach in capturing fine-grained structures. In particular, our method produces sharper reconstructions and more faithful geometry in challenging regions such as human faces and hands.

% \subsection{Novel View Synthesis}

% \paragraph{Dense-view Static Scene  Reconstruction}

% \paragraph{Sparse-view Static Scene Reconstruction
% }
% \paragraph{Sparse-view Dynamic Scene Novel View Synthesis}

% % \paragraph{Surface Reconstruction}

% \subsection{Surface Reconstruction}

% \input{Figures/fig_owlii_sparse_psnr_ssim_views}

% \input{Tables/sparse_view_owlii_4views}
% \input{Tables/Sparse_view_blender_4views}

\subsection{Dense-view Scene Reconstruction}
We evaluate our model’s effectiveness in dense-view reconstruction on both synthetic and real-world benchmarks. For comparison, the 3DGS-based methods are trained for 30k iterations. In our setup, we first pretrain our backbone model for 5k iterations and then jointly train with our module for an additional 25k iterations, using the same hyperparameters throughout to ensure a fair comparison.
\begin{table}[t]
\centering
\footnotesize
\setlength{\tabcolsep}{2.5pt} 
\caption{\textbf{Efficiency Benchmark.} 
Comparison of baseline methods (3DSG-MCMC, SplatFields) and our variants across reconstruction quality (PSNR), number of Gaussians (\#G), model size, training VRAM, training time, and rendering FPS.}
\begin{tabular}{@{}l l c c c c c c@{}}
\toprule
Dataset & Method & PSNR & \#G & Size & VRAM & Time & FPS \\
\midrule
\multirow{2}{*}{Blender} 
  & 3DSG-MCMC             & 33.81 & 0.29M &  69MB &  6GB &  5min   & 380 \\
  & 3DSG-MCMC + Ours      & 34.06 & 0.29M &  87MB & 15GB & 36min   &  85 \\
\midrule
\multirow{2}{*}{MipNeRF360}
  & 3DSG-MCMC             & 28.14 & 3.56M & 842MB & 15GB & 15min   & 135 \\
  & 3DSG-MCMC + Ours      & 28.24 & 3.56M & 860MB & 43GB & 75min   &  37 \\
\midrule
\multirow{2}{*}{Blender} 
  & SplatFields           & 22.58 & 0.30M &  74MB &  8GB & 80min   & 430 \\
  & SplatFields + Ours    & 23.05 & 0.33M &  86MB & 17GB & 135min  & 150 \\
\midrule
\multirow{2}{*}{Owlii}
  & SplatFields           & 27.87 & 0.10M &  28MB &  7GB &  14h    &  38 \\
  & SplatFields + Ours    & 29.25 & 0.10M &  31MB & 11GB &  23h    &  15 \\
\bottomrule
\end{tabular}

\label{tab:efficiency_benchmark}
\vspace{-2.50em}
\end{table}

\begin{figure*}
    \centering
    \includegraphics[width=0.9\linewidth]{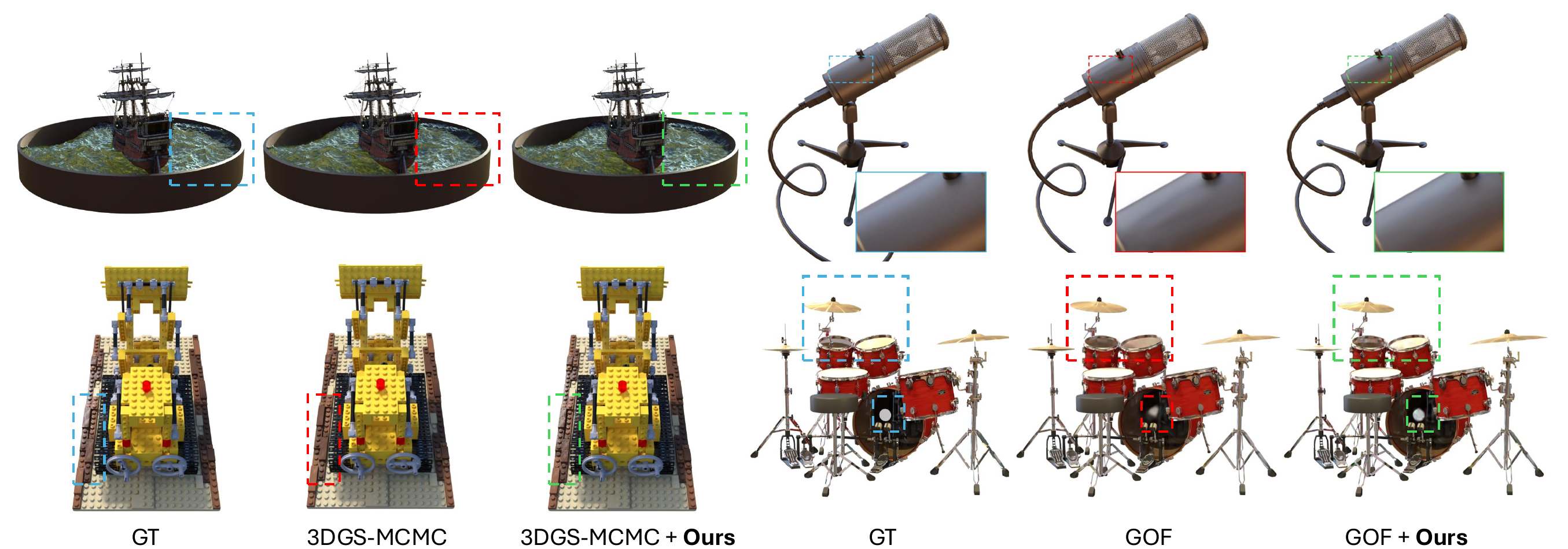}
     \vspace{-0.5em}
    \caption{\textbf{Qualitative Comparison on Blender using 100 input views.} The proposed neural texture representation produces smoother geometry while capturing high-frequency appearance details.}
    \label{fig:dense_blender}
    \vspace{-1.0em}
\end{figure*}

\boldparagraph{Novel View Synthesis}
We evaluate the effectiveness of our method in enhancing novel view synthesis by integrating it with two state-of-the-art baselines, GOF~\cite{Yu2024GOF} and 3DGS-MCMC~\cite{kheradmand20243dgc_mcmc}. Experiments are conducted on both the synthetic Blender dataset~\cite{mildenhall2021nerf} and the real-world MipNeRF360 dataset~\cite{barron2021mipnerf}. 
As shown in Table~\ref{tab:dense_view_all}, our approach consistently improves the performance of both baselines across both benchmarks, with particularly notable gains in PSNR. Moreover, our variants of existing 3DGS methods achieve the best performance among prior works~\cite{barron2022mipnerf360,muller2022instantngp,kerbl20233dgs,chao2024textured,Huang2DGS2024}.
Qualitative results are presented in Fig.~\ref{fig:dense_blender} and Fig.~\ref{fig:mipnerf360}, where our method enhances the baselines’ ability to capture high-frequency, view-dependent effects such as specular highlights. Moreover, it effectively preserves fine-grained structures (e.g., the cable shown in the figure) that are often missed by baseline methods due to their limited local expressiveness.

% We first show the improvement cuased by other method in two baselines GOF and 3DGS-MCMC. We first show results on synthetic dataset blender, our method is able to better capture view-dependet effects

% We further show results on an in-the-wild dataset mipnerf360, showing that our method also extends to outdoor scenes, better captures details and correct specular effects
\boldparagraph{Surface Reconstruction}
We further validate our model’s ability to recover more accurate surfaces by building upon the state-of-the-art 3DGS-based surface reconstruction method, GOF~\cite{Yu2024GOF}.
As benchmarked on the DTU~\cite{aanaes2016DTU} dataset and shown in Table~\ref{tab:dense_view_all}, our approach demonstrates improved surface reconstruction performance compared to the original method.
A detailed per-scene breakdown and additional baseline comparisons are provided in the appendix.
Qualitative results in Fig.~\ref{fig:dtu} demonstrate that our model produces more accurate and smoother surfaces, benefiting from the enhanced local expressiveness of Gaussian primitives for both geometry and appearance. Additionally, our ability to model view-dependent effects is particularly effective in recovering specular surface regions with fewer holes, as shown in the \textit{apple} example.

% % Table: Inference time breakdown (Blender)
% \begin{table}[t]
% \centering
% \caption{\revise{Inference time breakdown on Blender.}}
% \label{tab:inference_breakdown}
% \begin{tabular}{@{}l c@{}}
% \toprule
% Component & Inference Time (ms) \\
% \midrule
% 3DGS-MCMC   & 2.6 \\
% Intersection & 4.2 \\
% Tex Query    & 0.2 \\
% NN forward   & 4.7 \\
% \bottomrule
% \end{tabular}
% \end{table}

% \begin{table}[t]
% \centering
% \footnotesize
% \label{tab:inference_breakdown}
% \begin{tabular}{lcccc}
% \toprule
% Component & 3DGS-MCMC & Intersection & Tex Query & NN forward \\
% \midrule
% Inference Time (ms) & 2.6 & 4.2 & 0.2 & 4.7 \\
% \bottomrule
% \end{tabular}
% \caption{Inference time breakdown on Blender.}

% \end{table}

\begin{table}[t]
\centering
\footnotesize
% \scriptsize
\setlength{\tabcolsep}{2.5pt} 
\caption{\textbf{Breakdown of Inference Time on Blender Dataset.} 
We report average runtimes (ms) of each component when rendering a single $800{\times}800$ image 
from 0.3M primitives. Compared to vanilla 3DGS (Rasterization only), 
the additional computational overhead in our method is primarily due to the ray-Gaussian intersection and Neural Network (NN) forward pass.}
\begin{tabular}{lcccc}
\toprule
Component & Rasterization & Ray-Gaussian Intersection & Texture Query & NN Forward \\
\midrule
Time (ms) & 2.6 & 4.2 & 0.2 & 4.7 \\
\bottomrule
\end{tabular}

\label{tab:inference_breakdown}
  % \vspace{-2.0em}
\end{table}

\subsection{Efficiency Analysis}
We present a comprehensive efficiency analysis by reporting averaged results for baseline methods (3DGS-MCMC~\cite{kheradmand20243dgc_mcmc}, SplatFields~\cite{SplatFields}) and our variants across datasets.
Table~\ref{tab:efficiency_benchmark} summarizes the efficiency statistics, including reconstruction quality measured in PSNR, the number of Gaussians, model size, training GPU memory (VRAM), training time, and rendering FPS.
Under the same number of primitives, our method maintains a comparable model size thanks to the global neural network, which effectively compresses per-primitive local RGBA textures. 
The increase in training GPU memory primarily arises from storing RGBA texture gradients, while the forward rendering stage introduces approximately 1 GB of additional memory usage for millions of primitives. 
As detailed in Table~\ref{tab:inference_breakdown}, the reduction in rendering FPS mainly stems from the ray–Gaussian intersection algorithm adopted from GOF~\cite{Yu2024GOF} and the forward pass of the global neural network. 
Despite the additional overhead, our method achieves superior reconstruction quality and consistently outperforms the baselines at the same model size (see \textit{Ours (Full)$^{\dagger}$} in Table~\ref{tab:ablation}).

\subsection{Ablation Studies}

\begin{table}[!t]
  \centering
  \footnotesize
  \setlength{\tabcolsep}{3pt} % ‘adjust if needed
    \caption{\textbf{Ablation Studies on Model Variants.} 
    We ablate the effect of removing the global neural network (\textit{w/o Neural}) 
    and report the performance of our model at the same model size (\textit{Ours (Full)$^{\dagger}$}) as the baselines.}
  \begin{tabular}{lcc cc cc c}
    \toprule
    \multirow{2}{*}{Method} 
    & \multicolumn{2}{c}{Blender (100 views)} 
    & \multicolumn{2}{c}{Blender (4 views)} 
    & \multicolumn{2}{c}{Owlii (4 views)} 
    & \multicolumn{1}{c}{DTU} \\
    & PSNR$\uparrow$  & SSIM$\uparrow$ & PSNR$\uparrow$  & SSIM$\uparrow$ 
    & PSNR$\uparrow$  & SSIM$\uparrow$ 
    & CD$\downarrow$ \\
    \midrule
    Baseline  
    & 33.44 & 0.9698  
    & 19.16  & 0.8227 
    & 21.95 & 0.9160
    & 0.74 \\
     + Ours (w/o Neural)
    & 33.63 & 0.9696 
    & 19.14 & 0.8144  
    & 22.64 & 0.9072
    &  0.72 \\
    + Ours (Full)$^{\dagger}$ 
    & 34.00 &\textbf{ 0.9721  }
    & \textbf{20.00} & 0.8350
    & 25.01 & 0.9308
    & 0.68 \\
    + Ours (Full) & 
    \textbf{34.09} & 0.9710
    & 19.98 & \textbf{0.8351}
    & \textbf{25.05} & \textbf{0.9319}
    & \textbf{0.67} \\

    %  + Ours*(full)
    %   & \textbf{34.00} & \textbf{0.9721}  
    %   & \textbf{26.52} & \textbf{0.9436}
    %   & \textbf{20.00} & \textbf{0.8350}
    %   & \textbf{0.68} \\
    %  \textbf{+ Ours*(full}
    % & \textbf{34.01} & \textbf{0.9707} 
    % & \textbf{19.98} & \textbf{0.8351}
    % &\textbf{26.47} & \textbf{0.9460}
    % &  \textbf{0.67} \\ 

    \bottomrule
  \end{tabular}
    % \caption{\textbf{Ablation Studies.} We use GOF as the novel view sytnehsisbaseline Blender (100 views) dataset and  DTU dataset. For sparse-view dynamic scene reconstruction on the Dancer sequence, SplatFields is used as the baseline.}

    % \caption{\textbf{Ablation Studies.} 
    % GOF is used as the baseline for novel view synthesis on Blender (100 views) and for geometry reconstruction on DTU. 
    % For sparse-view reconstruction, we select 4 views and 10 view avearge from static Blender and dynamic Owlii, where SplatFields serves as the baseline. 
    % }
  \label{tab:ablation}
  % \vspace{-2.0em}
\end{table}

\begin{table}[t]
\centering
\small
\setlength{\tabcolsep}{4pt} % adjust if needed
\caption{\textbf{Effectiveness of Local RGBA Texture}.
We compare against variants where the global neural network predicts per-primitive RGB values or spherical harmonics with an additional alpha channel.
Our local RGBA texture representation yields consistently better meterics across datasets and backbones (GOF for Blender, 3DGS-MCMC for MipNeRF360).}
\begin{tabular}{@{}l
ccc
ccc@{}}
\toprule
& \multicolumn{3}{c}{Blender} & \multicolumn{3}{c}{MipNeRF360} \\
Method & PSNR$\uparrow$ & SSIM$\uparrow$ & LPIPS$\downarrow$
       & PSNR$\uparrow$ & SSIM$\uparrow$ & LPIPS$\downarrow$ \\
\midrule
Baseline  & 33.44 & 0.9698 & 0.0307 & 28.14 & 0.8427 & 0.1755\\
+ Neural RGBA  & 33.79 & 0.9703 & 0.0300 & 28.17 & 0.8413 & 0.1799 \\
+ Neural SHA   & 33.85 & 0.9703 & 0.0296 & 28.20 & 0.8415 & 0.1806 \\
+ Ours  & \textbf{34.09} & \textbf{0.9710} & \textbf{0.0286} & \textbf{28.24} & \textbf{0.8429} & \textbf{0.1748} \\
\bottomrule

\end{tabular}
% \vspace{-3.0em}

\label{tab:ablation_tex_effectiveness}
% \vspace{-2.00em}
\end{table}

% Table: Ablation studies
\begin{table}[t]
\centering
\small
% \footnotesize
\setlength{\tabcolsep}{5pt}
\caption{\textbf{Ablation Studies on Method Components and Hyperparameters}. 
We evaluate the impact of different components and hyperparameters on the Blender dataset with GOF as the backbone.}
\begin{tabular}{@{}l c c c c@{}}
\toprule
Method & PSNR$\uparrow$ & SSIM$\uparrow$ & LPIPS$\downarrow$ & Model Size (MB) \\
\midrule
Ours              & 34.09 & 0.9710 & 0.0286 & 68.2 \\
W/o CP Decomposition     & 34.03 & 0.9708 & 0.0284 & 75.1 \\
W/o Sparsity Loss        & 34.09 & 0.9708 & 0.0287 & 73.2 \\
W/o View-dependence      & 33.74 & 0.9704 & 0.0293 & 68.5 \\
W/ Tri-plane Resolution 128  & 34.08 & 0.9710 & 0.0286 & 63.1 \\
W/ Tri-plane Resolution 256  & 34.00 & 0.9708 & 0.0287 & 77.7 \\
W/ NN Depth 3          & 34.03 & 0.9710 & 0.0286 & 68.5 \\
W/ NN Width 256        & 34.06 & 0.9709 & 0.0287 & 68.9 \\
\bottomrule
\end{tabular}

% CP decomposition and sparsity loss reduce model size, while view-dependent modeling improves reconstruction quality.
\label{tab:ablation_components}
\vspace{-2.00em}
\end{table}

\boldparagraph{Effect of Global Neural Network} 
We first ablate our key idea of using a global neural network to model local texture fields, as shown in Table~\ref{tab:ablation}.
Compared to naive per-primitive RGBA textures (refer to \textit{w/o Neural} in the table), our neural texture splatting method achieves significant and consistent improvements across all evaluated benchmarks.
In particular, our method achieves notable gains of approximately 3 dB in PSNR for dynamic scene reconstruction on the \textit{Owlii (4 views)} dataset.
This improvement is primarily attributed to the ability of our neural texture modeling to capture time-dependent texture variations, a crucial capability that per-primitive RGBA textures lack.

\boldparagraph{Effect of Local RGBA Textures}
The variant \textit{w/o Neural} in Table~\ref{tab:ablation} shows that our local RGBA texture representation can improve baseline performance across all benchmarks without relying on a global neural network. To further assess its effectiveness, we compare against variants where the global neural network predicts per-primitive RGB values or spherical harmonics augmented with an alpha channel (denoted as \textit{Neural RGBA} and \textit{Neural SHA} in Table~\ref{tab:ablation_tex_effectiveness}). Our local texture fields consistently demonstrate superior results across datasets and backbones, using GOF as the baseline on Blender and 3DGS-MCMC on MipNeRF360.

\boldparagraph{Components and Hyperparameters} 
Table~\ref{tab:ablation_components} reports ablation studies on our method’s components and hyperparameters on the Blender dataset, with GOF as the backbone. The inclusion of CP decomposition (Eq.\eqref{eqn:cp_decompose}) and sparsity loss (Eq.\eqref{eqn:sparsity_loss}) help reduces model size, and view-dependent texture representation improves reconstruction quality. The results also show that our design choices consistently maintain high reconstruction quality across different architectural variations, such as global tri-plane resolution and neural network depth/width.

\section{Conclusion}
\label{sec:conclusion}
We have proposed Neural Texture Splatting (NTS), a novel extension for 3D Gaussian Splatting (3DGS) to enhance its local expressiveness and generalization ability.  We have demonstrated that NTS is most effective when tackling the sparse-view reconstruction problem, where it significantly improves the quality of both static and dynamic scene reconstruction.  We have also shown that NTS can be easily integrated into other 3DGS frameworks, such as geometry
reconstruction~\cite{Yu2024SIGGRAPHASIA} and dense-view novel view synthesis~\cite{kheradmand20243dgc_mcmc}, achieving noticeable improvements over the state-of-the-art on these tasks.  
Despite the improved reconstruction quality, future work could focus on enhancing computational efficiency, for example through more efficient ray–Gaussian intersection algorithms~\cite{bulo2025hardware} and faster neural architectures~\cite{muller2022instantngp}.
We believe that NTS is a promising step towards a more expressive 3DGS framework, and we hope that it will inspire further research in this direction.

% \section*{Acknowledgements}
\begin{acks}
We would like to express our gratitude to Brian Chao and Qin Han for their helpful discussions.
\end{acks}

\bibliographystyle{ACM-Reference-Format}
\bibliography{main}


\begin{thebibliography}{80}

%%% the Web address.

\ifx \showCODEN    \undefined \def \showCODEN     #1{\unskip}     \fi
\ifx \showISBNx    \undefined \def \showISBNx     #1{\unskip}     \fi
\ifx \showISBNxiii \undefined \def \showISBNxiii  #1{\unskip}     \fi
\ifx \showISSN     \undefined \def \showISSN      #1{\unskip}     \fi
\ifx \showLCCN     \undefined \def \showLCCN      #1{\unskip}     \fi
\ifx \shownote     \undefined \def \shownote      #1{#1}          \fi
\ifx \showarticletitle \undefined \def \showarticletitle #1{#1}   \fi
\ifx \showURL      \undefined \def \showURL       {\relax}        \fi
% invisible to TeX
\providecommand\bibfield[2]{#2}
\providecommand\bibinfo[2]{#2}
\providecommand\natexlab[1]{#1}
\providecommand\showeprint[2][]{arXiv:#2}

\bibitem[Aan{\ae}s et~al\mbox{.}(2016)]%
        {aanaes2016DTU}
\bibfield{author}{\bibinfo{person}{Henrik Aan{\ae}s}, \bibinfo{person}{Rasmus~Ramsb{\o}l Jensen}, \bibinfo{person}{George Vogiatzis}, \bibinfo{person}{Engin Tola}, {and} \bibinfo{person}{Anders~Bjorholm Dahl}.} \bibinfo{year}{2016}\natexlab{}.
\newblock \showarticletitle{Large-Scale Data for Multiple-View Stereopsis}.
\newblock \bibinfo{journal}{\emph{International Journal of Computer Vision}} (\bibinfo{year}{2016}), \bibinfo{pages}{1--16}.
\newblock


\bibitem[Barron et~al\mbox{.}(2021)]%
        {barron2021mipnerf}
\bibfield{author}{\bibinfo{person}{Jonathan~T Barron}, \bibinfo{person}{Ben Mildenhall}, \bibinfo{person}{Matthew Tancik}, \bibinfo{person}{Peter Hedman}, \bibinfo{person}{Ricardo Martin-Brualla}, {and} \bibinfo{person}{Pratul~P Srinivasan}.} \bibinfo{year}{2021}\natexlab{}.
\newblock \showarticletitle{Mip-nerf: A multiscale representation for anti-aliasing neural radiance fields}. In \bibinfo{booktitle}{\emph{Proceedings of the IEEE/CVF international conference on computer vision}}. \bibinfo{pages}{5855--5864}.
\newblock


\bibitem[Barron et~al\mbox{.}(2022)]%
        {barron2022mipnerf360}
\bibfield{author}{\bibinfo{person}{Jonathan~T. Barron}, \bibinfo{person}{Ben Mildenhall}, \bibinfo{person}{Dor Verbin}, \bibinfo{person}{Pratul~P. Srinivasan}, {and} \bibinfo{person}{Peter Hedman}.} \bibinfo{year}{2022}\natexlab{}.
\newblock \showarticletitle{Mip-NeRF 360: Unbounded Anti-Aliased Neural Radiance Fields}.
\newblock \bibinfo{journal}{\emph{CVPR}} (\bibinfo{year}{2022}).
\newblock


\bibitem[Broxton et~al\mbox{.}(2020)]%
        {broxton2020immersive}
\bibfield{author}{\bibinfo{person}{Michael Broxton}, \bibinfo{person}{John Flynn}, \bibinfo{person}{Ryan Overbeck}, \bibinfo{person}{Daniel Erickson}, \bibinfo{person}{Peter Hedman}, \bibinfo{person}{Matthew Duvall}, \bibinfo{person}{Jason Dourgarian}, \bibinfo{person}{Jay Busch}, \bibinfo{person}{Matt Whalen}, {and} \bibinfo{person}{Paul Debevec}.} \bibinfo{year}{2020}\natexlab{}.
\newblock \showarticletitle{Immersive light field video with a layered mesh representation}.
\newblock \bibinfo{journal}{\emph{ACM Transactions on Graphics (TOG)}} \bibinfo{volume}{39}, \bibinfo{number}{4} (\bibinfo{year}{2020}), \bibinfo{pages}{86--1}.
\newblock


\bibitem[Bulo et~al\mbox{.}(2025)]%
        {bulo2025hardware}
\bibfield{author}{\bibinfo{person}{Samuel~Rota Bulo}, \bibinfo{person}{Nemanja Bartolovic}, \bibinfo{person}{Lorenzo Porzi}, {and} \bibinfo{person}{Peter Kontschieder}.} \bibinfo{year}{2025}\natexlab{}.
\newblock \showarticletitle{Hardware-Rasterized Ray-Based Gaussian Splatting}. In \bibinfo{booktitle}{\emph{Proceedings of the Computer Vision and Pattern Recognition Conference}}. \bibinfo{pages}{485--494}.
\newblock


\bibitem[Chan et~al\mbox{.}(2022)]%
        {Chan2022CVPR}
\bibfield{author}{\bibinfo{person}{Eric~R. Chan}, \bibinfo{person}{Connor~Z. Lin}, \bibinfo{person}{Matthew~A. Chan}, \bibinfo{person}{Koki Nagano}, \bibinfo{person}{Boxiao Pan}, \bibinfo{person}{Shalini~De Mello}, \bibinfo{person}{Orazio Gallo}, \bibinfo{person}{Leonidas Guibas}, \bibinfo{person}{Jonathan Tremblay}, \bibinfo{person}{Sameh Khamis}, \bibinfo{person}{Tero Karras}, {and} \bibinfo{person}{Gordon Wetzstein}.} \bibinfo{year}{2022}\natexlab{}.
\newblock \showarticletitle{Efficient Geometry-aware {3D} Generative Adversarial Networks}. In \bibinfo{booktitle}{\emph{IEEE Conf. Comput. Vis. Pattern Recog.}}
\newblock


\bibitem[Chao et~al\mbox{.}(2024)]%
        {chao2024textured}
\bibfield{author}{\bibinfo{person}{Brian Chao}, \bibinfo{person}{Hung-Yu Tseng}, \bibinfo{person}{Lorenzo Porzi}, \bibinfo{person}{Chen Gao}, \bibinfo{person}{Tuotuo Li}, \bibinfo{person}{Qinbo Li}, \bibinfo{person}{Ayush Saraf}, \bibinfo{person}{Jia-Bin Huang}, \bibinfo{person}{Johannes Kopf}, \bibinfo{person}{Gordon Wetzstein}, {et~al\mbox{.}}} \bibinfo{year}{2024}\natexlab{}.
\newblock \showarticletitle{Textured Gaussians for Enhanced 3D Scene Appearance Modeling}.
\newblock \bibinfo{journal}{\emph{arXiv preprint arXiv:2411.18625}} (\bibinfo{year}{2024}).
\newblock


\bibitem[Charatan et~al\mbox{.}(2024)]%
        {charatan2024pixelsplat}
\bibfield{author}{\bibinfo{person}{David Charatan}, \bibinfo{person}{Sizhe~Lester Li}, \bibinfo{person}{Andrea Tagliasacchi}, {and} \bibinfo{person}{Vincent Sitzmann}.} \bibinfo{year}{2024}\natexlab{}.
\newblock \showarticletitle{pixelsplat: 3d gaussian splats from image pairs for scalable generalizable 3d reconstruction}. In \bibinfo{booktitle}{\emph{Proceedings of the IEEE/CVF Conference on Computer Vision and Pattern Recognition}}. \bibinfo{pages}{19457--19467}.
\newblock


\bibitem[Chen et~al\mbox{.}(2022)]%
        {Chen2022ECCV}
\bibfield{author}{\bibinfo{person}{Anpei Chen}, \bibinfo{person}{Zexiang Xu}, \bibinfo{person}{Andreas Geiger}, \bibinfo{person}{Jingyi Yu}, {and} \bibinfo{person}{Hao Su}.} \bibinfo{year}{2022}\natexlab{}.
\newblock \showarticletitle{TensoRF: Tensorial Radiance Fields}. In \bibinfo{booktitle}{\emph{Eur. Conf. Comput. Vis.}}
\newblock


\bibitem[Chen et~al\mbox{.}(2023)]%
        {Chen2023ARXIV}
\bibfield{author}{\bibinfo{person}{Hanlin Chen}, \bibinfo{person}{Chen Li}, \bibinfo{person}{Yunsong Wang}, {and} \bibinfo{person}{Gim~Hee Lee}.} \bibinfo{year}{2023}\natexlab{}.
\newblock \showarticletitle{NeuSG: Neural Implicit Surface Reconstruction with 3D Gaussian Splatting Guidance}.
\newblock \bibinfo{journal}{\emph{arXiv.org}}  \bibinfo{volume}{2312.00846} (\bibinfo{year}{2023}).
\newblock


\bibitem[Chen et~al\mbox{.}(2024)]%
        {chen2024mvsplat}
\bibfield{author}{\bibinfo{person}{Yuedong Chen}, \bibinfo{person}{Haofei Xu}, \bibinfo{person}{Chuanxia Zheng}, \bibinfo{person}{Bohan Zhuang}, \bibinfo{person}{Marc Pollefeys}, \bibinfo{person}{Andreas Geiger}, \bibinfo{person}{Tat-Jen Cham}, {and} \bibinfo{person}{Jianfei Cai}.} \bibinfo{year}{2024}\natexlab{}.
\newblock \showarticletitle{Mvsplat: Efficient 3d gaussian splatting from sparse multi-view images}. In \bibinfo{booktitle}{\emph{European Conference on Computer Vision}}. Springer.
\newblock


\bibitem[Chung et~al\mbox{.}(2024)]%
        {chung2024depthreg}
\bibfield{author}{\bibinfo{person}{Jaeyoung Chung}, \bibinfo{person}{Jeongtaek Oh}, {and} \bibinfo{person}{Kyoung~Mu Lee}.} \bibinfo{year}{2024}\natexlab{}.
\newblock \showarticletitle{Depth-regularized optimization for 3d gaussian splatting in few-shot images}. In \bibinfo{booktitle}{\emph{Proceedings of the IEEE/CVF Conference on Computer Vision and Pattern Recognition}}. \bibinfo{pages}{811--820}.
\newblock


\bibitem[Dai et~al\mbox{.}(2024a)]%
        {Dai2024SIGGRAPH}
\bibfield{author}{\bibinfo{person}{Pinxuan Dai}, \bibinfo{person}{Jiamin Xu}, \bibinfo{person}{Wenxiang Xie}, \bibinfo{person}{Xinguo Liu}, \bibinfo{person}{Huamin Wang}, {and} \bibinfo{person}{Weiwei Xu}.} \bibinfo{year}{2024}\natexlab{a}.
\newblock \showarticletitle{High-quality Surface Reconstruction using Gaussian Surfels}. In \bibinfo{booktitle}{\emph{Proc. of SIGGRAPH}}.
\newblock


\bibitem[Dai et~al\mbox{.}(2024b)]%
        {Dai2024GaussianSurfels}
\bibfield{author}{\bibinfo{person}{Pinxuan Dai}, \bibinfo{person}{Jiamin Xu}, \bibinfo{person}{Wenxiang Xie}, \bibinfo{person}{Xinguo Liu}, \bibinfo{person}{Huamin Wang}, {and} \bibinfo{person}{Weiwei Xu}.} \bibinfo{year}{2024}\natexlab{b}.
\newblock \showarticletitle{High-quality Surface Reconstruction using Gaussian Surfels}. In \bibinfo{booktitle}{\emph{ACM SIGGRAPH 2024 Conference Papers}}. \bibinfo{publisher}{Association for Computing Machinery}, Article \bibinfo{articleno}{22}, \bibinfo{numpages}{11}~pages.
\newblock


\bibitem[Fan et~al\mbox{.}(2025)]%
        {fan2025advances}
\bibfield{author}{\bibinfo{person}{Jinlong Fan}, \bibinfo{person}{Xuepu Zeng}, \bibinfo{person}{Jing Zhang}, \bibinfo{person}{Mingming Gong}, \bibinfo{person}{Yuxiang Yang}, {and} \bibinfo{person}{Dacheng Tao}.} \bibinfo{year}{2025}\natexlab{}.
\newblock \showarticletitle{Advances in Radiance Field for Dynamic Scene: From Neural Field to Gaussian Field}.
\newblock \bibinfo{journal}{\emph{arXiv preprint arXiv:2505.10049}} (\bibinfo{year}{2025}).
\newblock


\bibitem[Fan et~al\mbox{.}(2023)]%
        {fan2023lightgaussian}
\bibfield{author}{\bibinfo{person}{Zhiwen Fan}, \bibinfo{person}{Kevin Wang}, \bibinfo{person}{Kairun Wen}, \bibinfo{person}{Zehao Zhu}, \bibinfo{person}{Dejia Xu}, {and} \bibinfo{person}{Zhangyang Wang}.} \bibinfo{year}{2023}\natexlab{}.
\newblock \bibinfo{title}{LightGaussian: Unbounded 3D Gaussian Compression with 15x Reduction and 200+ FPS}.
\newblock
\showeprint[arxiv]{2311.17245}~[cs.CV]


\bibitem[Flynn et~al\mbox{.}(2016)]%
        {flynn2016deepstereo}
\bibfield{author}{\bibinfo{person}{John Flynn}, \bibinfo{person}{Ivan Neulander}, \bibinfo{person}{James Philbin}, {and} \bibinfo{person}{Noah Snavely}.} \bibinfo{year}{2016}\natexlab{}.
\newblock \showarticletitle{Deepstereo: Learning to predict new views from the world's imagery}. In \bibinfo{booktitle}{\emph{Proceedings of the IEEE conference on computer vision and pattern recognition}}. \bibinfo{pages}{5515--5524}.
\newblock


\bibitem[Guangcong et~al\mbox{.}(2023)]%
        {wang2023sparsenerf}
\bibfield{author}{\bibinfo{person}{Guangcong}, \bibinfo{person}{Zhaoxi Chen}, \bibinfo{person}{Chen~Change Loy}, {and} \bibinfo{person}{Ziwei Liu}.} \bibinfo{year}{2023}\natexlab{}.
\newblock \showarticletitle{SparseNeRF: Distilling Depth Ranking for Few-shot Novel View Synthesis}.
\newblock \bibinfo{journal}{\emph{IEEE/CVF International Conference on Computer Vision (ICCV)}} (\bibinfo{year}{2023}).
\newblock


\bibitem[Gu{\'e}don and Lepetit(2024a)]%
        {Guedon2024CVPR}
\bibfield{author}{\bibinfo{person}{Antoine Gu{\'e}don} {and} \bibinfo{person}{Vincent Lepetit}.} \bibinfo{year}{2024}\natexlab{a}.
\newblock \showarticletitle{SuGaR: Surface-Aligned Gaussian Splatting for Efficient 3D Mesh Reconstruction and High-Quality Mesh Rendering}. In \bibinfo{booktitle}{\emph{IEEE Conf. Comput. Vis. Pattern Recog.}}
\newblock


\bibitem[Gu{\'e}don and Lepetit(2024b)]%
        {guedon2024sugar}
\bibfield{author}{\bibinfo{person}{Antoine Gu{\'e}don} {and} \bibinfo{person}{Vincent Lepetit}.} \bibinfo{year}{2024}\natexlab{b}.
\newblock \showarticletitle{Sugar: Surface-aligned gaussian splatting for efficient 3d mesh reconstruction and high-quality mesh rendering}. In \bibinfo{booktitle}{\emph{Proceedings of the IEEE/CVF Conference on Computer Vision and Pattern Recognition}}. \bibinfo{pages}{5354--5363}.
\newblock


\bibitem[Huang et~al\mbox{.}(2024)]%
        {Huang2DGS2024}
\bibfield{author}{\bibinfo{person}{Binbin Huang}, \bibinfo{person}{Zehao Yu}, \bibinfo{person}{Anpei Chen}, \bibinfo{person}{Andreas Geiger}, {and} \bibinfo{person}{Shenghua Gao}.} \bibinfo{year}{2024}\natexlab{}.
\newblock \showarticletitle{2D Gaussian Splatting for Geometrically Accurate Radiance Fields}. In \bibinfo{booktitle}{\emph{SIGGRAPH 2024 Conference Papers}}. \bibinfo{publisher}{Association for Computing Machinery}.
\newblock
\href{https://doi.org/10.1145/3641519.3657428}{doi:\nolinkurl{10.1145/3641519.3657428}}


\bibitem[Huang and Gong(2024)]%
        {huang2024texturedGS}
\bibfield{author}{\bibinfo{person}{Zhentao Huang} {and} \bibinfo{person}{Minglun Gong}.} \bibinfo{year}{2024}\natexlab{}.
\newblock \showarticletitle{Textured-GS: Gaussian Splatting with Spatially Defined Color and Opacity}.
\newblock \bibinfo{journal}{\emph{arXiv preprint arXiv:2407.09733}} (\bibinfo{year}{2024}).
\newblock


\bibitem[Kerbl et~al\mbox{.}(2023)]%
        {kerbl20233dgs}
\bibfield{author}{\bibinfo{person}{Bernhard Kerbl}, \bibinfo{person}{Georgios Kopanas}, \bibinfo{person}{Thomas Leimk{\"u}hler}, {and} \bibinfo{person}{George Drettakis}.} \bibinfo{year}{2023}\natexlab{}.
\newblock \showarticletitle{3d gaussian splatting for real-time radiance field rendering.}
\newblock \bibinfo{journal}{\emph{ACM Trans. Graph.}} \bibinfo{volume}{42}, \bibinfo{number}{4} (\bibinfo{year}{2023}), \bibinfo{pages}{139--1}.
\newblock


\bibitem[Kheradmand et~al\mbox{.}(2024)]%
        {kheradmand20243dgc_mcmc}
\bibfield{author}{\bibinfo{person}{Shakiba Kheradmand}, \bibinfo{person}{Daniel Rebain}, \bibinfo{person}{Gopal Sharma}, \bibinfo{person}{Weiwei Sun}, \bibinfo{person}{Yang-Che Tseng}, \bibinfo{person}{Hossam Isack}, \bibinfo{person}{Abhishek Kar}, \bibinfo{person}{Andrea Tagliasacchi}, {and} \bibinfo{person}{Kwang~Moo Yi}.} \bibinfo{year}{2024}\natexlab{}.
\newblock \showarticletitle{3d gaussian splatting as markov chain monte carlo}.
\newblock \bibinfo{journal}{\emph{Advances in Neural Information Processing Systems}}  \bibinfo{volume}{37} (\bibinfo{year}{2024}), \bibinfo{pages}{80965--80986}.
\newblock


\bibitem[Lei et~al\mbox{.}(2024)]%
        {lei2024mosca}
\bibfield{author}{\bibinfo{person}{Jiahui Lei}, \bibinfo{person}{Yijia Weng}, \bibinfo{person}{Adam Harley}, \bibinfo{person}{Leonidas Guibas}, {and} \bibinfo{person}{Kostas Daniilidis}.} \bibinfo{year}{2024}\natexlab{}.
\newblock \showarticletitle{Mosca: Dynamic gaussian fusion from casual videos via 4d motion scaffolds}.
\newblock \bibinfo{journal}{\emph{arXiv preprint arXiv:2405.17421}} (\bibinfo{year}{2024}).
\newblock


\bibitem[Li et~al\mbox{.}(2025)]%
        {Li2025CVPR}
\bibfield{author}{\bibinfo{person}{Haolin Li}, \bibinfo{person}{Jinyang Liu}, \bibinfo{person}{Mario Sznaier}, {and} \bibinfo{person}{Octavia Camps}.} \bibinfo{year}{2025}\natexlab{}.
\newblock \showarticletitle{3d-hgs: 3d half-gaussian splatting}. In \bibinfo{booktitle}{\emph{IEEE Conf. Comput. Vis. Pattern Recog.}}
\newblock


\bibitem[Li et~al\mbox{.}(2022)]%
        {li2022neural3DV}
\bibfield{author}{\bibinfo{person}{Tianye Li}, \bibinfo{person}{Mira Slavcheva}, \bibinfo{person}{Michael Zollhoefer}, \bibinfo{person}{Simon Green}, \bibinfo{person}{Christoph Lassner}, \bibinfo{person}{Changil Kim}, \bibinfo{person}{Tanner Schmidt}, \bibinfo{person}{Steven Lovegrove}, \bibinfo{person}{Michael Goesele}, \bibinfo{person}{Richard Newcombe}, {et~al\mbox{.}}} \bibinfo{year}{2022}\natexlab{}.
\newblock \showarticletitle{Neural 3d video synthesis from multi-view video}. In \bibinfo{booktitle}{\emph{Proceedings of the IEEE/CVF Conference on Computer Vision and Pattern Recognition}}. \bibinfo{pages}{5521--5531}.
\newblock


\bibitem[Li et~al\mbox{.}(2023)]%
        {Li2023CVPR}
\bibfield{author}{\bibinfo{person}{Zhaoshuo Li}, \bibinfo{person}{Thomas M\"uller}, \bibinfo{person}{Alex Evans}, \bibinfo{person}{Russell~H Taylor}, \bibinfo{person}{Mathias Unberath}, \bibinfo{person}{Ming-Yu Liu}, {and} \bibinfo{person}{Chen-Hsuan Lin}.} \bibinfo{year}{2023}\natexlab{}.
\newblock \showarticletitle{Neuralangelo: High-Fidelity Neural Surface Reconstruction}. In \bibinfo{booktitle}{\emph{IEEE Conf. Comput. Vis. Pattern Recog.}}
\newblock


\bibitem[Lu et~al\mbox{.}(2024b)]%
        {scaffoldgs}
\bibfield{author}{\bibinfo{person}{Tao Lu}, \bibinfo{person}{Mulin Yu}, \bibinfo{person}{Linning Xu}, \bibinfo{person}{Yuanbo Xiangli}, \bibinfo{person}{Limin Wang}, \bibinfo{person}{Dahua Lin}, {and} \bibinfo{person}{Bo Dai}.} \bibinfo{year}{2024}\natexlab{b}.
\newblock \showarticletitle{Scaffold-gs: Structured 3d gaussians for view-adaptive rendering}. In \bibinfo{booktitle}{\emph{Proceedings of the IEEE/CVF Conference on Computer Vision and Pattern Recognition}}. \bibinfo{pages}{20654--20664}.
\newblock


\bibitem[Lu et~al\mbox{.}(2024a)]%
        {lu20243d}
\bibfield{author}{\bibinfo{person}{Zhicheng Lu}, \bibinfo{person}{Xiang Guo}, \bibinfo{person}{Le Hui}, \bibinfo{person}{Tianrui Chen}, \bibinfo{person}{Min Yang}, \bibinfo{person}{Xiao Tang}, \bibinfo{person}{Feng Zhu}, {and} \bibinfo{person}{Yuchao Dai}.} \bibinfo{year}{2024}\natexlab{a}.
\newblock \showarticletitle{3d geometry-aware deformable gaussian splatting for dynamic view synthesis}. In \bibinfo{booktitle}{\emph{Proceedings of the IEEE/CVF Conference on Computer Vision and Pattern Recognition}}. \bibinfo{pages}{8900--8910}.
\newblock


\bibitem[Mihajlovic et~al\mbox{.}(2024a)]%
        {mihajlovic2024ResFields}
\bibfield{author}{\bibinfo{person}{Marko Mihajlovic}, \bibinfo{person}{Sergey Prokudin}, \bibinfo{person}{Marc Pollefeys}, {and} \bibinfo{person}{Siyu Tang}.} \bibinfo{year}{2024}\natexlab{a}.
\newblock \showarticletitle{{ResFields}: Residual Neural Fields for Spatiotemporal Signals}. In \bibinfo{booktitle}{\emph{International Conference on Learning Representations (ICLR)}}.
\newblock


\bibitem[Mihajlovic et~al\mbox{.}(2024b)]%
        {SplatFields}
\bibfield{author}{\bibinfo{person}{Marko Mihajlovic}, \bibinfo{person}{Sergey Prokudin}, \bibinfo{person}{Siyu Tang}, \bibinfo{person}{Robert Maier}, \bibinfo{person}{Federica Bogo}, \bibinfo{person}{Tony Tung}, {and} \bibinfo{person}{Edmond Boyer}.} \bibinfo{year}{2024}\natexlab{b}.
\newblock \showarticletitle{SplatFields: Neural Gaussian Splats for Sparse 3D and 4D Reconstruction}. In \bibinfo{booktitle}{\emph{European Conference on Computer Vision (ECCV)}}. Springer.
\newblock


\bibitem[Mildenhall et~al\mbox{.}(2020)]%
        {mildenhall2021nerf}
\bibfield{author}{\bibinfo{person}{Ben Mildenhall}, \bibinfo{person}{Pratul~P Srinivasan}, \bibinfo{person}{Matthew Tancik}, \bibinfo{person}{Jonathan~T Barron}, \bibinfo{person}{Ravi Ramamoorthi}, {and} \bibinfo{person}{Ren Ng}.} \bibinfo{year}{2020}\natexlab{}.
\newblock \showarticletitle{Nerf: Representing scenes as neural radiance fields for view synthesis}. In \bibinfo{booktitle}{\emph{Eur. Conf. Comput. Vis.}}
\newblock


\bibitem[M{\"u}ller et~al\mbox{.}(2022)]%
        {muller2022instantngp}
\bibfield{author}{\bibinfo{person}{Thomas M{\"u}ller}, \bibinfo{person}{Alex Evans}, \bibinfo{person}{Christoph Schied}, {and} \bibinfo{person}{Alexander Keller}.} \bibinfo{year}{2022}\natexlab{}.
\newblock \showarticletitle{Instant neural graphics primitives with a multiresolution hash encoding}.
\newblock \bibinfo{journal}{\emph{ACM transactions on graphics (TOG)}} \bibinfo{volume}{41}, \bibinfo{number}{4} (\bibinfo{year}{2022}), \bibinfo{pages}{1--15}.
\newblock


\bibitem[Oechsle et~al\mbox{.}(2021)]%
        {Oechsle2021ICCV}
\bibfield{author}{\bibinfo{person}{Michael Oechsle}, \bibinfo{person}{Songyou Peng}, {and} \bibinfo{person}{Andreas Geiger}.} \bibinfo{year}{2021}\natexlab{}.
\newblock \showarticletitle{UNISURF: Unifying Neural Implicit Surfaces and Radiance Fields for Multi-View Reconstruction}. In \bibinfo{booktitle}{\emph{Int. Conf. Comput. Vis.}}
\newblock


\bibitem[Park et~al\mbox{.}(2021a)]%
        {park2021nerfies}
\bibfield{author}{\bibinfo{person}{Keunhong Park}, \bibinfo{person}{Utkarsh Sinha}, \bibinfo{person}{Jonathan~T Barron}, \bibinfo{person}{Sofien Bouaziz}, \bibinfo{person}{Dan~B Goldman}, \bibinfo{person}{Steven~M Seitz}, {and} \bibinfo{person}{Ricardo Martin-Brualla}.} \bibinfo{year}{2021}\natexlab{a}.
\newblock \showarticletitle{Nerfies: Deformable neural radiance fields}. In \bibinfo{booktitle}{\emph{Proceedings of the IEEE/CVF international conference on computer vision}}. \bibinfo{pages}{5865--5874}.
\newblock


\bibitem[Park et~al\mbox{.}(2021b)]%
        {park2021hypernerf}
\bibfield{author}{\bibinfo{person}{Keunhong Park}, \bibinfo{person}{Utkarsh Sinha}, \bibinfo{person}{Peter Hedman}, \bibinfo{person}{Jonathan~T. Barron}, \bibinfo{person}{Sofien Bouaziz}, \bibinfo{person}{Dan~B Goldman}, \bibinfo{person}{Ricardo Martin-Brualla}, {and} \bibinfo{person}{Steven~M. Seitz}.} \bibinfo{year}{2021}\natexlab{b}.
\newblock \showarticletitle{HyperNeRF: A Higher-Dimensional Representation for Topologically Varying Neural Radiance Fields}.
\newblock \bibinfo{journal}{\emph{ACM Trans. Graph.}} \bibinfo{volume}{40}, \bibinfo{number}{6}, Article \bibinfo{articleno}{238} (\bibinfo{date}{dec} \bibinfo{year}{2021}).
\newblock


\bibitem[Peng et~al\mbox{.}(2020)]%
        {Peng2020ECCV}
\bibfield{author}{\bibinfo{person}{Songyou Peng}, \bibinfo{person}{Michael Niemeyer}, \bibinfo{person}{Lars Mescheder}, \bibinfo{person}{Marc Pollefeys}, {and} \bibinfo{person}{Andreas Geiger}.} \bibinfo{year}{2020}\natexlab{}.
\newblock \showarticletitle{Convolutional Occupancy Networks}. In \bibinfo{booktitle}{\emph{Eur. Conf. Comput. Vis.}}
\newblock


\bibitem[Pumarola et~al\mbox{.}(2021)]%
        {pumarola2021dnerf}
\bibfield{author}{\bibinfo{person}{Albert Pumarola}, \bibinfo{person}{Enric Corona}, \bibinfo{person}{Gerard Pons-Moll}, {and} \bibinfo{person}{Francesc Moreno-Noguer}.} \bibinfo{year}{2021}\natexlab{}.
\newblock \showarticletitle{D-nerf: Neural radiance fields for dynamic scenes}. In \bibinfo{booktitle}{\emph{Proceedings of the IEEE/CVF conference on computer vision and pattern recognition}}. \bibinfo{pages}{10318--10327}.
\newblock


\bibitem[Qian et~al\mbox{.}(2024)]%
        {qian20243dgs}
\bibfield{author}{\bibinfo{person}{Zhiyin Qian}, \bibinfo{person}{Shaofei Wang}, \bibinfo{person}{Marko Mihajlovic}, \bibinfo{person}{Andreas Geiger}, {and} \bibinfo{person}{Siyu Tang}.} \bibinfo{year}{2024}\natexlab{}.
\newblock \showarticletitle{3dgs-avatar: Animatable avatars via deformable 3d gaussian splatting}. In \bibinfo{booktitle}{\emph{Proceedings of the IEEE/CVF conference on computer vision and pattern recognition}}. \bibinfo{pages}{5020--5030}.
\newblock


\bibitem[Reiser et~al\mbox{.}(2024)]%
        {reiser2024binary}
\bibfield{author}{\bibinfo{person}{Christian Reiser}, \bibinfo{person}{Stephan Garbin}, \bibinfo{person}{Pratul Srinivasan}, \bibinfo{person}{Dor Verbin}, \bibinfo{person}{Richard Szeliski}, \bibinfo{person}{Ben Mildenhall}, \bibinfo{person}{Jonathan Barron}, \bibinfo{person}{Peter Hedman}, {and} \bibinfo{person}{Andreas Geiger}.} \bibinfo{year}{2024}\natexlab{}.
\newblock \showarticletitle{Binary opacity grids: Capturing fine geometric detail for mesh-based view synthesis}.
\newblock \bibinfo{journal}{\emph{ACM Transactions on Graphics (TOG)}} \bibinfo{volume}{43}, \bibinfo{number}{4} (\bibinfo{year}{2024}), \bibinfo{pages}{1--14}.
\newblock


\bibitem[Reiser et~al\mbox{.}(2021)]%
        {reiser2021kilonerf}
\bibfield{author}{\bibinfo{person}{Christian Reiser}, \bibinfo{person}{Songyou Peng}, \bibinfo{person}{Yiyi Liao}, {and} \bibinfo{person}{Andreas Geiger}.} \bibinfo{year}{2021}\natexlab{}.
\newblock \showarticletitle{Kilonerf: Speeding up neural radiance fields with thousands of tiny mlps}. In \bibinfo{booktitle}{\emph{Proceedings of the IEEE/CVF international conference on computer vision}}. \bibinfo{pages}{14335--14345}.
\newblock


\bibitem[Rong et~al\mbox{.}(2024)]%
        {rong2024gstex}
\bibfield{author}{\bibinfo{person}{Victor Rong}, \bibinfo{person}{Jingxiang Chen}, \bibinfo{person}{Sherwin Bahmani}, \bibinfo{person}{Kiriakos~N Kutulakos}, {and} \bibinfo{person}{David~B Lindell}.} \bibinfo{year}{2024}\natexlab{}.
\newblock \showarticletitle{Gstex: Per-primitive texturing of 2d gaussian splatting for decoupled appearance and geometry modeling}.
\newblock \bibinfo{journal}{\emph{arXiv preprint arXiv:2409.12954}} (\bibinfo{year}{2024}).
\newblock


\bibitem[Rosu and Behnke(2023)]%
        {Alexandru2023CVPR}
\bibfield{author}{\bibinfo{person}{Radu~Alexandru Rosu} {and} \bibinfo{person}{Sven Behnke}.} \bibinfo{year}{2023}\natexlab{}.
\newblock \showarticletitle{PermutoSDF: Fast Multi-View Reconstruction with Implicit Surfaces using Permutohedral Lattices}. In \bibinfo{booktitle}{\emph{IEEE Conf. Comput. Vis. Pattern Recog.}}
\newblock


\bibitem[Shi et~al\mbox{.}(2024)]%
        {Shi2024CVPR}
\bibfield{author}{\bibinfo{person}{Ruoxi Shi}, \bibinfo{person}{Xinyue Wei}, \bibinfo{person}{Cheng Wang}, {and} \bibinfo{person}{Hao Su}.} \bibinfo{year}{2024}\natexlab{}.
\newblock \showarticletitle{ZeroRF: Fast Sparse View 360deg Reconstruction with Zero Pretraining}. In \bibinfo{booktitle}{\emph{IEEE Conf. Comput. Vis. Pattern Recog.}}
\newblock


\bibitem[Sitzmann et~al\mbox{.}(2020)]%
        {sitzmann2019siren}
\bibfield{author}{\bibinfo{person}{Vincent Sitzmann}, \bibinfo{person}{Julien~N.P. Martel}, \bibinfo{person}{Alexander~W. Bergman}, \bibinfo{person}{David~B. Lindell}, {and} \bibinfo{person}{Gordon Wetzstein}.} \bibinfo{year}{2020}\natexlab{}.
\newblock \showarticletitle{Implicit Neural Representations with Periodic Activation Functions}. In \bibinfo{booktitle}{\emph{Advances in Neural Information Processing Systems (NeurIPS)}}.
\newblock


\bibitem[Sitzmann et~al\mbox{.}(2019)]%
        {sitzmann2019scene}
\bibfield{author}{\bibinfo{person}{Vincent Sitzmann}, \bibinfo{person}{Michael Zollh{\"o}fer}, {and} \bibinfo{person}{Gordon Wetzstein}.} \bibinfo{year}{2019}\natexlab{}.
\newblock \showarticletitle{Scene representation networks: Continuous 3d-structure-aware neural scene representations}. In \bibinfo{booktitle}{\emph{Advances in Neural Information Processing Systems (NeurIPS)}}.
\newblock


\bibitem[Song et~al\mbox{.}(2024)]%
        {song2024hdgstextured2dgaussian}
\bibfield{author}{\bibinfo{person}{Yunzhou Song}, \bibinfo{person}{Heguang Lin}, \bibinfo{person}{Jiahui Lei}, \bibinfo{person}{Lingjie Liu}, {and} \bibinfo{person}{Kostas Daniilidis}.} \bibinfo{year}{2024}\natexlab{}.
\newblock \bibinfo{title}{HDGS: Textured 2D Gaussian Splatting for Enhanced Scene Rendering}.
\newblock
\showeprint[arxiv]{2412.01823}~[cs.CV]
\urldef\tempurl%
\url{https://arxiv.org/abs/2412.01823}
\showURL{%
\tempurl}


\bibitem[Sun et~al\mbox{.}(2022)]%
        {sun2022direct}
\bibfield{author}{\bibinfo{person}{Cheng Sun}, \bibinfo{person}{Min Sun}, {and} \bibinfo{person}{Hwann-Tzong Chen}.} \bibinfo{year}{2022}\natexlab{}.
\newblock \showarticletitle{Direct voxel grid optimization: Super-fast convergence for radiance fields reconstruction}. In \bibinfo{booktitle}{\emph{Proceedings of the IEEE/CVF conference on computer vision and pattern recognition}}. \bibinfo{pages}{5459--5469}.
\newblock


\bibitem[Sun et~al\mbox{.}(2024)]%
        {sun20243dgstream}
\bibfield{author}{\bibinfo{person}{Jiakai Sun}, \bibinfo{person}{Han Jiao}, \bibinfo{person}{Guangyuan Li}, \bibinfo{person}{Zhanjie Zhang}, \bibinfo{person}{Lei Zhao}, {and} \bibinfo{person}{Wei Xing}.} \bibinfo{year}{2024}\natexlab{}.
\newblock \showarticletitle{3dgstream: On-the-fly training of 3d gaussians for efficient streaming of photo-realistic free-viewpoint videos}. In \bibinfo{booktitle}{\emph{Proceedings of the IEEE/CVF Conference on Computer Vision and Pattern Recognition}}. \bibinfo{pages}{20675--20685}.
\newblock


\bibitem[Svitov et~al\mbox{.}(2024)]%
        {svitov2024billboard}
\bibfield{author}{\bibinfo{person}{David Svitov}, \bibinfo{person}{Pietro Morerio}, \bibinfo{person}{Lourdes Agapito}, {and} \bibinfo{person}{Alessio Del~Bue}.} \bibinfo{year}{2024}\natexlab{}.
\newblock \showarticletitle{BillBoard Splatting (BBSplat): Learnable Textured Primitives for Novel View Synthesis}.
\newblock \bibinfo{journal}{\emph{arXiv preprint arXiv:2411.08508}} (\bibinfo{year}{2024}).
\newblock


\bibitem[Tang et~al\mbox{.}(2024)]%
        {tang2024mv}
\bibfield{author}{\bibinfo{person}{Zhenggang Tang}, \bibinfo{person}{Yuchen Fan}, \bibinfo{person}{Dilin Wang}, \bibinfo{person}{Hongyu Xu}, \bibinfo{person}{Rakesh Ranjan}, \bibinfo{person}{Alexander Schwing}, {and} \bibinfo{person}{Zhicheng Yan}.} \bibinfo{year}{2024}\natexlab{}.
\newblock \showarticletitle{MV-DUSt3R+: Single-Stage Scene Reconstruction from Sparse Views In 2 Seconds}.
\newblock \bibinfo{journal}{\emph{arXiv preprint arXiv:2412.06974}} (\bibinfo{year}{2024}).
\newblock


\bibitem[Tewari et~al\mbox{.}(2022)]%
        {tewari2022advances}
\bibfield{author}{\bibinfo{person}{Ayush Tewari}, \bibinfo{person}{Justus Thies}, \bibinfo{person}{Ben Mildenhall}, \bibinfo{person}{Pratul Srinivasan}, \bibinfo{person}{Edgar Tretschk}, \bibinfo{person}{Wang Yifan}, \bibinfo{person}{Christoph Lassner}, \bibinfo{person}{Vincent Sitzmann}, \bibinfo{person}{Ricardo Martin-Brualla}, \bibinfo{person}{Stephen Lombardi}, {et~al\mbox{.}}} \bibinfo{year}{2022}\natexlab{}.
\newblock \showarticletitle{Advances in neural rendering}. In \bibinfo{booktitle}{\emph{Computer Graphics Forum}}.
\newblock


\bibitem[Turkulainen et~al\mbox{.}(2024)]%
        {Turkulainen2024WACV}
\bibfield{author}{\bibinfo{person}{Matias Turkulainen}, \bibinfo{person}{Xuqian Ren}, \bibinfo{person}{Iaroslav Melekhov}, \bibinfo{person}{Otto Seiskari}, \bibinfo{person}{Esa Rahtu}, {and} \bibinfo{person}{Juho Kannala}.} \bibinfo{year}{2024}\natexlab{}.
\newblock \showarticletitle{DN-Splatter: Depth and Normal Priors for Gaussian Splatting and Meshing}. In \bibinfo{booktitle}{\emph{IEEE Winter Conference on Applications of Computer Vision (WACV)}}.
\newblock


\bibitem[Wang et~al\mbox{.}(2021)]%
        {Wang2021NEURIPS}
\bibfield{author}{\bibinfo{person}{Peng Wang}, \bibinfo{person}{Lingjie Liu}, \bibinfo{person}{Yuan Liu}, \bibinfo{person}{Christian Theobalt}, \bibinfo{person}{Taku Komura}, {and} \bibinfo{person}{Wenping Wang}.} \bibinfo{year}{2021}\natexlab{}.
\newblock \showarticletitle{NeuS: Learning Neural Implicit Surfaces by Volume Rendering for Multi-view Reconstruction}. In \bibinfo{booktitle}{\emph{Advances in Neural Information Processing Systems (NeurIPS)}}.
\newblock


\bibitem[Wang et~al\mbox{.}(2023)]%
        {Wang2023ICCV}
\bibfield{author}{\bibinfo{person}{Yiming Wang}, \bibinfo{person}{Qin Han}, \bibinfo{person}{Marc Habermann}, \bibinfo{person}{Kostas Daniilidis}, \bibinfo{person}{Christian Theobalt}, {and} \bibinfo{person}{Lingjie Liu}.} \bibinfo{year}{2023}\natexlab{}.
\newblock \showarticletitle{NeuS2: Fast Learning of Neural Implicit Surfaces for Multi-view Reconstruction}. In \bibinfo{booktitle}{\emph{Int. Conf. Comput. Vis.}}
\newblock


\bibitem[Wang et~al\mbox{.}(2004)]%
        {SSIM}
\bibfield{author}{\bibinfo{person}{Zhou Wang}, \bibinfo{person}{Alan~C Bovik}, \bibinfo{person}{Hamid~R Sheikh}, {and} \bibinfo{person}{Eero~P Simoncelli}.} \bibinfo{year}{2004}\natexlab{}.
\newblock \showarticletitle{Image quality assessment: from error visibility to structural similarity}.
\newblock \bibinfo{journal}{\emph{IEEE transactions on image processing}} \bibinfo{volume}{13}, \bibinfo{number}{4} (\bibinfo{year}{2004}), \bibinfo{pages}{600--612}.
\newblock


\bibitem[Wu et~al\mbox{.}(2024a)]%
        {Wu2024CVPR}
\bibfield{author}{\bibinfo{person}{Guanjun Wu}, \bibinfo{person}{Taoran Yi}, \bibinfo{person}{Jiemin Fang}, \bibinfo{person}{Lingxi Xie}, \bibinfo{person}{Xiaopeng Zhang}, \bibinfo{person}{Wei Wei}, \bibinfo{person}{Wenyu Liu}, \bibinfo{person}{Qi Tian}, {and} \bibinfo{person}{Xinggang Wang}.} \bibinfo{year}{2024}\natexlab{a}.
\newblock \showarticletitle{4D Gaussian Splatting for Real-Time Dynamic Scene Rendering}. In \bibinfo{booktitle}{\emph{IEEE Conf. Comput. Vis. Pattern Recog.}}
\newblock


\bibitem[Wu et~al\mbox{.}(2024b)]%
        {wu20244dgaussian}
\bibfield{author}{\bibinfo{person}{Guanjun Wu}, \bibinfo{person}{Taoran Yi}, \bibinfo{person}{Jiemin Fang}, \bibinfo{person}{Lingxi Xie}, \bibinfo{person}{Xiaopeng Zhang}, \bibinfo{person}{Wei Wei}, \bibinfo{person}{Wenyu Liu}, \bibinfo{person}{Qi Tian}, {and} \bibinfo{person}{Xinggang Wang}.} \bibinfo{year}{2024}\natexlab{b}.
\newblock \showarticletitle{4d gaussian splatting for real-time dynamic scene rendering}. In \bibinfo{booktitle}{\emph{Proceedings of the IEEE/CVF conference on computer vision and pattern recognition}}. \bibinfo{pages}{20310--20320}.
\newblock


\bibitem[Wu et~al\mbox{.}(2024c)]%
        {Wu_2024_CVPR}
\bibfield{author}{\bibinfo{person}{Guanjun Wu}, \bibinfo{person}{Taoran Yi}, \bibinfo{person}{Jiemin Fang}, \bibinfo{person}{Lingxi Xie}, \bibinfo{person}{Xiaopeng Zhang}, \bibinfo{person}{Wei Wei}, \bibinfo{person}{Wenyu Liu}, \bibinfo{person}{Qi Tian}, {and} \bibinfo{person}{Xinggang Wang}.} \bibinfo{year}{2024}\natexlab{c}.
\newblock \showarticletitle{4D Gaussian Splatting for Real-Time Dynamic Scene Rendering}. In \bibinfo{booktitle}{\emph{Proceedings of the IEEE/CVF Conference on Computer Vision and Pattern Recognition (CVPR)}}. \bibinfo{pages}{20310--20320}.
\newblock


\bibitem[Xu et~al\mbox{.}(2024a)]%
        {xu2024supergaussians}
\bibfield{author}{\bibinfo{person}{Rui Xu}, \bibinfo{person}{Wenyue Chen}, \bibinfo{person}{Jiepeng Wang}, \bibinfo{person}{Yuan Liu}, \bibinfo{person}{Peng Wang}, \bibinfo{person}{Lin Gao}, \bibinfo{person}{Shiqing Xin}, \bibinfo{person}{Taku Komura}, \bibinfo{person}{Xin Li}, {and} \bibinfo{person}{Wenping Wang}.} \bibinfo{year}{2024}\natexlab{a}.
\newblock \showarticletitle{SuperGaussians: Enhancing Gaussian Splatting Using Primitives with Spatially Varying Colors}.
\newblock \bibinfo{journal}{\emph{arXiv preprint arXiv:2411.18966}} (\bibinfo{year}{2024}).
\newblock


\bibitem[Xu et~al\mbox{.}(2024b)]%
        {xu2024texture}
\bibfield{author}{\bibinfo{person}{Tian-Xing Xu}, \bibinfo{person}{Wenbo Hu}, \bibinfo{person}{Yu-Kun Lai}, \bibinfo{person}{Ying Shan}, {and} \bibinfo{person}{Song-Hai Zhang}.} \bibinfo{year}{2024}\natexlab{b}.
\newblock \showarticletitle{Texture-gs: Disentangling the geometry and texture for 3d gaussian splatting editing}. In \bibinfo{booktitle}{\emph{European Conference on Computer Vision}}. Springer, \bibinfo{pages}{37--53}.
\newblock


\bibitem[Xu et~al\mbox{.}(2017)]%
        {xu2017owlii}
\bibfield{author}{\bibinfo{person}{Yi Xu}, \bibinfo{person}{Yao Lu}, {and} \bibinfo{person}{Ziyu Wen}.} \bibinfo{year}{2017}\natexlab{}.
\newblock \showarticletitle{Owlii Dynamic human mesh sequence dataset}. In \bibinfo{booktitle}{\emph{ISO/IEC JTC1/SC29/WG11 m41658, 120th MPEG Meeting}}.
\newblock


\bibitem[Yang et~al\mbox{.}(2023)]%
        {yang2023deformable3dgs}
\bibfield{author}{\bibinfo{person}{Ziyi Yang}, \bibinfo{person}{Xinyu Gao}, \bibinfo{person}{Wen Zhou}, \bibinfo{person}{Shaohui Jiao}, \bibinfo{person}{Yuqing Zhang}, {and} \bibinfo{person}{Xiaogang Jin}.} \bibinfo{year}{2023}\natexlab{}.
\newblock \showarticletitle{Deformable 3D Gaussians for High-Fidelity Monocular Dynamic Scene Reconstruction}.
\newblock \bibinfo{journal}{\emph{arXiv preprint arXiv:2309.13101}} (\bibinfo{year}{2023}).
\newblock


\bibitem[Yang et~al\mbox{.}(2024a)]%
        {yang2024deformable}
\bibfield{author}{\bibinfo{person}{Ziyi Yang}, \bibinfo{person}{Xinyu Gao}, \bibinfo{person}{Wen Zhou}, \bibinfo{person}{Shaohui Jiao}, \bibinfo{person}{Yuqing Zhang}, {and} \bibinfo{person}{Xiaogang Jin}.} \bibinfo{year}{2024}\natexlab{a}.
\newblock \showarticletitle{Deformable 3d gaussians for high-fidelity monocular dynamic scene reconstruction}. In \bibinfo{booktitle}{\emph{Proceedings of the IEEE/CVF conference on computer vision and pattern recognition}}. \bibinfo{pages}{20331--20341}.
\newblock


\bibitem[Yang et~al\mbox{.}(2024b)]%
        {yang2023gs4d}
\bibfield{author}{\bibinfo{person}{Zeyu Yang}, \bibinfo{person}{Hongye Yang}, \bibinfo{person}{Zijie Pan}, {and} \bibinfo{person}{Li Zhang}.} \bibinfo{year}{2024}\natexlab{b}.
\newblock \showarticletitle{Real-time Photorealistic Dynamic Scene Representation and Rendering with 4D Gaussian Splatting}. In \bibinfo{booktitle}{\emph{International Conference on Learning Representations (ICLR)}}.
\newblock


\bibitem[Yariv et~al\mbox{.}(2021)]%
        {Yariv2021NEURIPS}
\bibfield{author}{\bibinfo{person}{Lior Yariv}, \bibinfo{person}{Jiatao Gu}, \bibinfo{person}{Yoni Kasten}, {and} \bibinfo{person}{Yaron Lipman}.} \bibinfo{year}{2021}\natexlab{}.
\newblock \showarticletitle{Volume rendering of neural implicit surfaces}. In \bibinfo{booktitle}{\emph{Advances in Neural Information Processing Systems (NeurIPS)}}.
\newblock


\bibitem[Yariv et~al\mbox{.}(2023)]%
        {Yariv2023SIGGRAPH}
\bibfield{author}{\bibinfo{person}{Lior Yariv}, \bibinfo{person}{Peter Hedman}, \bibinfo{person}{Christian Reiser}, \bibinfo{person}{Dor Verbin}, \bibinfo{person}{Pratul~P. Srinivasan}, \bibinfo{person}{Richard Szeliski}, \bibinfo{person}{Jonathan~T. Barron}, {and} \bibinfo{person}{Ben Mildenhall}.} \bibinfo{year}{2023}\natexlab{}.
\newblock \showarticletitle{BakedSDF: Meshing Neural SDFs for Real-Time View Synthesis}. In \bibinfo{booktitle}{\emph{Proc. of SIGGRAPH}}.
\newblock


\bibitem[Yu et~al\mbox{.}(2024b)]%
        {yu2024cogs}
\bibfield{author}{\bibinfo{person}{Heng Yu}, \bibinfo{person}{Joel Julin}, \bibinfo{person}{Zolt{\'a}n~{\'A} Milacski}, \bibinfo{person}{Koichiro Niinuma}, {and} \bibinfo{person}{L{\'a}szl{\'o}~A Jeni}.} \bibinfo{year}{2024}\natexlab{b}.
\newblock \showarticletitle{Cogs: Controllable gaussian splatting}. In \bibinfo{booktitle}{\emph{Proceedings of the IEEE/CVF Conference on Computer Vision and Pattern Recognition}}.
\newblock


\bibitem[Yu et~al\mbox{.}(2024c)]%
        {Yu2024NEURIPS}
\bibfield{author}{\bibinfo{person}{Mulin Yu}, \bibinfo{person}{Tao Lu}, \bibinfo{person}{Linning Xu}, \bibinfo{person}{Lihan Jiang}, \bibinfo{person}{Yuanbo Xiangli}, {and} \bibinfo{person}{Bo Dai}.} \bibinfo{year}{2024}\natexlab{c}.
\newblock \showarticletitle{GSDF: 3DGS Meets SDF for Improved Rendering and Reconstruction}. In \bibinfo{booktitle}{\emph{Advances in Neural Information Processing Systems (NeurIPS)}}.
\newblock


\bibitem[Yu et~al\mbox{.}(2024a)]%
        {Yu2023MipSplatting}
\bibfield{author}{\bibinfo{person}{Zehao Yu}, \bibinfo{person}{Anpei Chen}, \bibinfo{person}{Binbin Huang}, \bibinfo{person}{Torsten Sattler}, {and} \bibinfo{person}{Andreas Geiger}.} \bibinfo{year}{2024}\natexlab{a}.
\newblock \showarticletitle{Mip-Splatting: Alias-free 3D Gaussian Splatting}.
\newblock \bibinfo{journal}{\emph{Conference on Computer Vision and Pattern Recognition (CVPR)}} (\bibinfo{year}{2024}).
\newblock


\bibitem[Yu et~al\mbox{.}(2024d)]%
        {Yu2024GOF}
\bibfield{author}{\bibinfo{person}{Zehao Yu}, \bibinfo{person}{Torsten Sattler}, {and} \bibinfo{person}{Andreas Geiger}.} \bibinfo{year}{2024}\natexlab{d}.
\newblock \showarticletitle{Gaussian Opacity Fields: Efficient Adaptive Surface Reconstruction in Unbounded Scenes}.
\newblock \bibinfo{journal}{\emph{ACM Transactions on Graphics}} (\bibinfo{year}{2024}).
\newblock


\bibitem[Yu et~al\mbox{.}(2024e)]%
        {Yu2024SIGGRAPHASIA}
\bibfield{author}{\bibinfo{person}{Zehao Yu}, \bibinfo{person}{Torsten Sattler}, {and} \bibinfo{person}{Andreas Geiger}.} \bibinfo{year}{2024}\natexlab{e}.
\newblock \showarticletitle{Gaussian Opacity Fields: Efficient Adaptive Surface Reconstruction in Unbounded Scenes}.
\newblock \bibinfo{journal}{\emph{ACM Trans. Graph.}} \bibinfo{volume}{43}, \bibinfo{number}{6} (\bibinfo{year}{2024}), \bibinfo{pages}{271:1--271:13}.
\newblock


\bibitem[Zhang et~al\mbox{.}(2025b)]%
        {zhang2025transplat}
\bibfield{author}{\bibinfo{person}{Chuanrui Zhang}, \bibinfo{person}{Yingshuang Zou}, \bibinfo{person}{Zhuoling Li}, \bibinfo{person}{Minmin Yi}, {and} \bibinfo{person}{Haoqian Wang}.} \bibinfo{year}{2025}\natexlab{b}.
\newblock \showarticletitle{Transplat: Generalizable 3d gaussian splatting from sparse multi-view images with transformers}. In \bibinfo{booktitle}{\emph{Proceedings of the AAAI Conference on Artificial Intelligence}}.
\newblock


\bibitem[Zhang et~al\mbox{.}(2018)]%
        {zhang2018lpips}
\bibfield{author}{\bibinfo{person}{Richard Zhang}, \bibinfo{person}{Phillip Isola}, \bibinfo{person}{Alexei~A Efros}, \bibinfo{person}{Eli Shechtman}, {and} \bibinfo{person}{Oliver Wang}.} \bibinfo{year}{2018}\natexlab{}.
\newblock \showarticletitle{The unreasonable effectiveness of deep features as a perceptual metric}. In \bibinfo{booktitle}{\emph{Proceedings of the IEEE conference on computer vision and pattern recognition}}. \bibinfo{pages}{586--595}.
\newblock


\bibitem[Zhang et~al\mbox{.}(2025a)]%
        {zhang2025nest}
\bibfield{author}{\bibinfo{person}{Xin Zhang}, \bibinfo{person}{Anpei Chen}, \bibinfo{person}{Jincheng Xiong}, \bibinfo{person}{Pinxuan Dai}, \bibinfo{person}{Yujun Shen}, {and} \bibinfo{person}{Weiwei Xu}.} \bibinfo{year}{2025}\natexlab{a}.
\newblock \showarticletitle{Neural Shell Texture Splatting: More Details and Fewer Primitives}. In \bibinfo{booktitle}{\emph{Proceedings of the IEEE/CVF International Conference on Computer Vision (ICCV)}}.
\newblock


\bibitem[Zhou et~al\mbox{.}(2018)]%
        {zhou2018rel10k}
\bibfield{author}{\bibinfo{person}{Tinghui Zhou}, \bibinfo{person}{Richard Tucker}, \bibinfo{person}{John Flynn}, \bibinfo{person}{Graham Fyffe}, {and} \bibinfo{person}{Noah Snavely}.} \bibinfo{year}{2018}\natexlab{}.
\newblock \showarticletitle{Stereo Magnification: Learning View Synthesis using Multiplane Images}. In \bibinfo{booktitle}{\emph{SIGGRAPH}}.
\newblock


\bibitem[Zhu et~al\mbox{.}(2025)]%
        {Zhu2025CVPR}
\bibfield{author}{\bibinfo{person}{Jialin Zhu}, \bibinfo{person}{Jiangbei Yue}, \bibinfo{person}{Feixiang He}, {and} \bibinfo{person}{He Wang}.} \bibinfo{year}{2025}\natexlab{}.
\newblock \showarticletitle{3D Student Splatting and Scooping}. In \bibinfo{booktitle}{\emph{IEEE Conf. Comput. Vis. Pattern Recog.}}
\newblock


\bibitem[Zhu et~al\mbox{.}(2024)]%
        {zhu2024motiongs}
\bibfield{author}{\bibinfo{person}{Ruijie Zhu}, \bibinfo{person}{Yanzhe Liang}, \bibinfo{person}{Hanzhi Chang}, \bibinfo{person}{Jiacheng Deng}, \bibinfo{person}{Jiahao Lu}, \bibinfo{person}{Wenfei Yang}, \bibinfo{person}{Tianzhu Zhang}, {and} \bibinfo{person}{Yongdong Zhang}.} \bibinfo{year}{2024}\natexlab{}.
\newblock \showarticletitle{Motiongs: Exploring explicit motion guidance for deformable 3d gaussian splatting}. In \bibinfo{booktitle}{\emph{Advances in Neural Information Processing Systems (NeurIPS)}}.
\newblock


\bibitem[Zwicker et~al\mbox{.}(2002)]%
        {zwicker2002ewa}
\bibfield{author}{\bibinfo{person}{Matthias Zwicker}, \bibinfo{person}{Hanspeter Pfister}, \bibinfo{person}{Jeroen Van~Baar}, {and} \bibinfo{person}{Markus Gross}.} \bibinfo{year}{2002}\natexlab{}.
\newblock \showarticletitle{EWA splatting}.
\newblock \bibinfo{journal}{\emph{IEEE Transactions on Visualization and Computer Graphics}} \bibinfo{volume}{8}, \bibinfo{number}{3} (\bibinfo{year}{2002}), \bibinfo{pages}{223--238}.
\newblock


\end{thebibliography}

\clearpage
\newpage

\begin{figure*}
    \centering
    \includegraphics[width=0.9\linewidth]{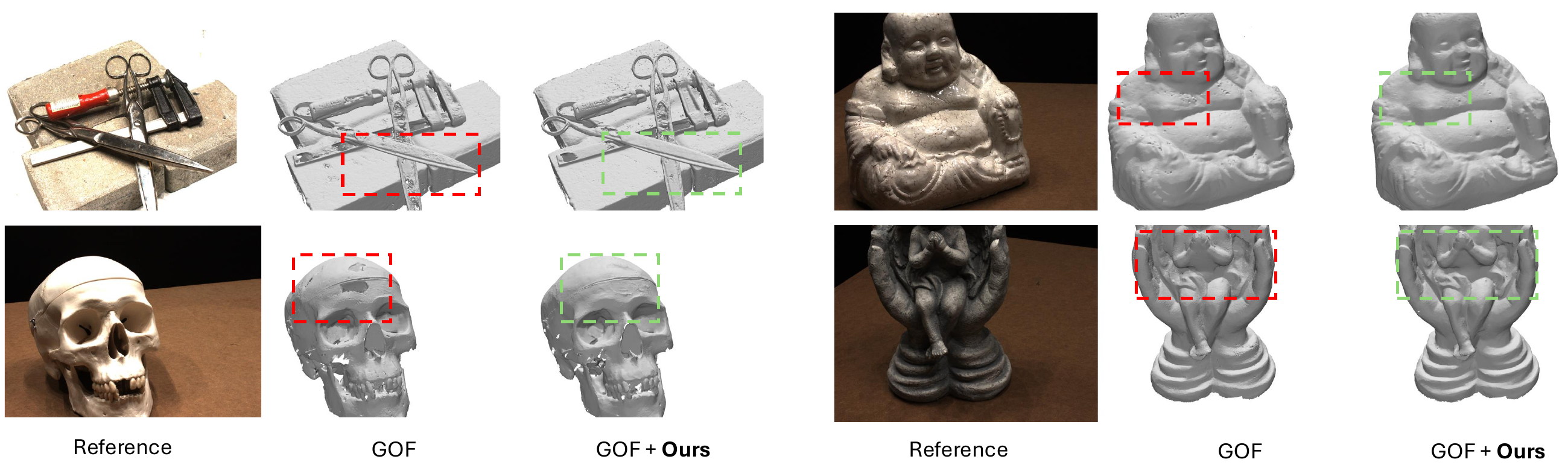}
    \vspace{-1.0em}
    \caption{\textbf{Qualitative comparison of surface reconstruction results on the DTU dataset. }
    Our method improves the mesh quality for both surface extraction approaches used in GOF: TSDF Fusion (first row) and GOF’s proposed method (second row).}

    \label{fig:dtu}
\end{figure*}

\clearpage
\newpage
% \setcounter{page}{1}
% \maketitlesupplementary
% \section*{Appendix}
% \twocolumn[{
% \vspace*{-0.5em}
% \begin{center}
% {\Large \sffamily Supplemental Material for}\\[0.4em]
% {\huge \bfseries \sffamily Physics-Informed Learning of Characteristic Trajectories\\
% for Smoke Reconstruction}\\[0.8em]
% {\large \sffamily
% Yiming Wang\textsuperscript{1}\quad
% Siyu Tang\textsuperscript{1}\quad
% Mengyu Chu\textsuperscript{2}\\[0.25em]
% \textsuperscript{1}ETH Zurich, Switzerland\quad
% \textsuperscript{2}Peking University \& SKL of General AI, China
% }
% \end{center}
% \vspace{0.6em}
% % \hrule height0.4pt
% \vspace{1.0em}
% }]

% \twocolumn[
% \begin{center}
% {\Huge \sffamily
%     Neural Texture Splatting: Expressive 3D Gaussian Splatting  for \newline View Synthesis, Geometry, and Dynamic Reconstruction - Supplementary	}\newline
%         \vspace{-20pt}
% 		% {\LARGE \newline SUBMISSION ID: 880}
% 		\subsection*{ 
%         \begin{center}
% 		{\LARGE Yiming Wang,} ETH Zurich, Switzerland
%           \end{center}
%          \begin{center}
%         {\LARGE Shaofei Wang,} ETH Zurich, Switzerland
%         \end{center}
%          \begin{center}
%         {\LARGE Marko Mihajlovic,} ETH Zurich, Switzerland
%         \end{center}
%          \begin{center}
%         {\LARGE Siyu Tang,} ETH Zurich, Switzerland
%         \end{center}
% 		}
        
% \end{center}
%   \vspace{12pt}
% ]

\twocolumn[
\begin{center}
{\Huge \sffamily
    Neural Texture Splatting: Expressive 3D Gaussian Splatting \\
    for View Synthesis, Geometry, and Dynamic Reconstruction -- Supplementary
}\vspace{10pt}

{\Large
Yiming Wang, Shaofei Wang, Marko Mihajlovic, Siyu Tang \\[4pt]
\textit{ETH Zurich, Switzerland}
}
\end{center}
\vspace{12pt}
]

\appendix
\section*{Appendix}
%\label{app:detail}

% \noindent

% \clearpage
% \setcounter{page}{1}
% \maketitlesupplementary  % 显示“Supplementary Material”页眉（如果模板支持）

% \appendix
% \section*{Appendix}
% \addcontentsline{toc}{section}{Appendix}

\section{Implementation details}
\subsection{Model Parameters}
% All other hyperparameters are kept identical to the backbone. Our method is trained alongside each baseline for the same number of iterations to ensure a fair comparison.
% For our local tri-plane texture resolution $\tau$, we set the resolution to 8 for Blender, and 4 for both Mip-NeRF360 and DTU datasets.
We use a tri-plane of size $192 \times 192 \times 16$ across all scenes for the global neural field.  The associated neural network has a width of 128, consists of two hidden layers.
To enhance representational capacity in dense-view reconstruction tasks, we use separate learnable global tri-planes and distinct neural network architectures for modeling RGB textures and alpha textures.  We adopt a standard linear MLP with ReLU activation for color decoding, while a SIREN network~\cite{sitzmann2019siren} is adopted for alpha decoding.
For sparse-view reconstruction of static and dynamic scenes, we build upon SplatFields by employing a CNN-based tri-plane generator~\cite{Shi2024CVPR} to provide a deep structural prior, and utilize the ResFields MLP~\cite{mihajlovic2024ResFields} for decoding of color and alpha values.
% \boldparagraph{Implementation Details}
%
For our local tri-plane texture resolution $\tau$, we set $\tau = 8$ when evaluating on the Blender dataset~\cite{mildenhall2021nerf} and DTU dataset~\cite{aanaes2016DTU}, and $\tau = 4$ for all other experiments.

\subsection{Training Details}

% To ensure a fair comparison, our method is trained alongside each baseline for the same number of iterations.
We run all experiments on  NVIDIA RTX 4090 GPUs, except for running Mip-NeRF360 on NVIDIA H200 due to additional memory costs.

\boldparagraph{Sparse-view Scene Reconstruction}
We adopt the training setup and hyperparameter configurations from SplatFields to ensure a fair and consistent comparison across all experiments.
For the Blender dataset, both SplatFields and our method are trained for 40k iterations using identical configurations. Our model is initialized from the SplatFields checkpoint at 20k iterations and jointly trained for the remaining 20k iterations.
For dynamic scene reconstruction on the Owlii dataset, both SplatFields and our method are trained for 200k iterations. Specifically, our model is initialized from the SplatFields checkpoint at 100k iterations and trained jointly for an additional 100k iterations. 

\boldparagraph{Dense-view Scene Reconstruction}
Following the standard training protocol commonly adopted in 3DGS-based methods, we train all models for 30k iterations on the Blender and Mip-NeRF360 datasets. In our setup, we first pretrain the backbone network for 5k iterations, followed by 25k iterations of joint training with our proposed module. All experiments use the same hyperparameters and optimization settings to ensure a fair comparison.

\section{Supplementary Ablation Results}

\boldparagraph{Modeling the RGBA Texture Field}
We explore a simplified approximation of the RGBA volume using a single 2D plane, aligned with the splat's principal axis. 
As shown in Tab.~\ref{tab:dense_view_blender}, when combined with our Neural Texture method, both representations improve upon the baseline.
We adopt the 3D tri-plane representation for modeling the RGBA texture field throughout this paper.

\begin{table}[t]
  \centering
  % \footnotesize\
  \caption{\textbf{Comparison of 2D Texture and 3D Texture modelling on Blender.} 
  Modeling our RGBA Texture Field with either 2D or 3D textures leads to improvements over the baseline.
  }
  \setlength{\tabcolsep}{8pt}
  \begin{tabular}{lccc}
    \toprule
    Method & PSNR$\uparrow$ & SSIM$\uparrow$ & LPIPS$\downarrow$ \\
    \midrule
    GoF & 33.44 & 0.9698 & 0.0307 \\
    + Ours (2D Texture) & 33.93 & 0.9705 & 0.0290 \\
    + Ours (3D Texture) & \textbf{34.09} & \textbf{0.9710} & \textbf{0.0286} \\
    \bottomrule
  \end{tabular}

  \vspace{-1.0em}
  \label{tab:dense_view_blender}
\end{table}

\begin{table}[t]
\centering
\setlength{\tabcolsep}{1.5pt} 
\small

\caption{
\textbf{Efficiency analysis on GOF.} We report the average training time and rendering speed on the Blender dataset using the original GOF* implementation as released in the paper. 
Note that the latest version of GOF* merges CUDA operations, achieving a 2× speed-up, an optimization also applicable to our method. 
While our proposed module introduces additional computational overhead compared to GOF*, this trade-off yields significantly improved performance in both novel view synthesis metrics on Blender and surface reconstruction accuracy (CD) on DTU.
}
% \vspace{-0.5em}
\begin{tabular}{lcccccc}
\toprule
Method & CD$\downarrow$ & PSNR$\uparrow$  & SSIM$\uparrow$ & LPIPS$\downarrow$ & Training (min) & FPS \\
% \midrule
\hline
GOF* & 0.74 &33.44 & 0.9698 & 0.0307 & 39.50 & 28.32\\
GOF* + \textbf{Ours}  & \textbf{0.67} & \textbf{34.09} & \textbf{0.9710} & \textbf{0.0286} & 61.78 & 20.82 \\
% \hdashline
% 2DGS + Triplane Encoded Tex & \textbf{31.73} & \textbf{27.5}\\
\bottomrule
\end{tabular}

\vspace{-1.0em}
\label{tab:runtime}
% \vspace{-0.5em}
\end{table}

% \begin{table}[!t]
%   \centering
%   \footnotesize
%   \setlength{\tabcolsep}{4pt} % adjust if needed
%   \begin{tabular}{lcccccccc}
%     \toprule
%     \multirow{2}{*}{Method} 
%     & \multicolumn{4}{c}{Blender} 
%     & \multicolumn{4}{c}{Mip-NeRF360} \\
%     \cmidrule(lr){2-5} \cmidrule(lr){6-9}
%     & PSNR$\uparrow$  & SSIM$\uparrow$ & LPIPS$\downarrow$ & Disk Storage (MB)$\downarrow$
%     & PSNR$\uparrow$  & SSIM$\uparrow$ & LPIPS$\downarrow$ & Disk Storage (MB)$\downarrow$ \\
%     \midrule
%     Textured Gaussians       & 33.24 & 0.9674 & 0.0428 & ~51M   & 27.35    & 0.8274     & 0.1858   & ~854M \\
%     GOF + Ours               & 34.01 & 0.9707 & 0.0291 & ~15M     & 27.69    & 0.8256     & 0.1977     & ~15M \\
%     3DGS-MCMC + Ours         & 34.06 & 0.9724 & 0.0275 & ~15M     & 28.24   & 0.8429     & 0.1748     & ~15M \\
%     \bottomrule
%   \end{tabular}
%   \caption{\textbf{Comparison with Textured-GS on Blender and Mip-NeRF360.} We show PSNR, SSIM, LPIPS, and disk storage cost for different methods.}
%   \label{tab:dense_view_all}
% \end{table}

\begin{table}[!t]
  \centering
  \small
  \setlength{\tabcolsep}{1pt} % adjust if needed
    \caption{\textbf{Comparison with State-of-the-art per-primitive texture methods Textured-GS on Blender and Mip-NeRF360.}
  We report the disk storage required for storing textures. Our NTS achieves higher PSNR while using less storage. }
  \begin{tabular}{lcccc}
    \toprule
    \multirow{2}{*}{Method} 
    & \multicolumn{2}{c}{Blender} 
    & \multicolumn{2}{c}{MipNeRF360} \\
    % \cmidrule(lr){2-3} \cmidrule(lr){4-5}
    & PSNR$\uparrow$  & Model Size (MB)
    & PSNR$\uparrow$  & Model Size (MB) \\
    \midrule
    Textured Gaussians       & 33.24 & $\sim$51M   & 27.35 & $\sim$854M \\
    GOF + \textbf{Ours}               & \textbf{34.09} & $\sim$15M   & 27.69 & $\sim$15M \\
    3DGS-MCMC + \textbf{Ours}         & 34.06 & $\sim$15M   & \textbf{28.24 }& $\sim$15M \\
    \bottomrule
  \end{tabular}

  \label{tab:cmp_with_textured_gs}
  % \vspace{-3.5em}
  \vspace{-1.0em}
\end{table}

\boldparagraph{Efficiency Analysis}
An efficiency analysis of our method is presented in Table~\ref{tab:runtime}, with experiments conducted on an NVIDIA RTX 4090 GPU. While our approach significantly improves upon the baseline performance, it comes with increased computational cost during training and a slight decrease in rendering speed. A more detailed discussion is provided in the Limitations section.

\begin{table}[]
  \centering
  \scriptsize
  \setlength{\tabcolsep}{4pt}

\caption{
\textbf{Quantitative Comparison on Sparse-View Dynamic Scene Reconstruction (10 Input Views).}
% We conduct evaluation on 4 sequences from the Owlii dataset, each using 10 input views. 
SSIM values are scaled by a factor of 100 for readability; coloring \colorbox[RGB]{\colorfirst}{1st}, \colorbox[RGB]{\colorsecond}{2nd}, and \colorbox[RGB]{\colorthird}{3rd} highlight best results.
}
\resizebox{\linewidth}{!}{%
\begin{tabular}{@{}clccccc@{}}
\toprule
\multirow{2}{*}{Method}
&& \multicolumn{5}{c}{PSNR $\uparrow$} \\
&& \textit{Mean} & Dancer & Exercise & Model & Basketball \\
\midrule
 \multirow{5}{*}{\rotatebox{90}{4D NeRFs}}
& DyNeRF & 29.70 & 28.22 & 30.64 & 29.95 & 30.00 \\
& TNeRF & 30.39 & 29.12 & 31.00 & 30.71 & 30.71 \\
& DNeRF & 30.25 & 29.39 & 30.63 & 30.63 & 30.35 \\
& Nerfies & \cellcolor[RGB]{\colorthird}30.70 & 29.57 & \cellcolor[RGB]{\colorthird}31.08 & 30.53 & \cellcolor[RGB]{\colorsecond}31.60 \\
& HyperNeRF & 30.36 & \cellcolor[RGB]{\colorthird}30.09 & 30.39 & \cellcolor[RGB]{\colorthird}30.88 & 30.08 \\
\midrule
\multirow{5}{*}{\rotatebox{90}{Splatting}} 
& 4D-GS & 28.05 & 28.11 & 29.09 & 29.06 & 25.94 \\
& Deformable3DGS & 27.76 & 27.86 & 28.78 & 26.47 & 27.95 \\
& 4DGaussians & 29.80 & 28.46 & 30.21 & 30.69 & 29.82 \\
& SplatFields4D & \cellcolor[RGB]{\colorsecond}31.32 & \cellcolor[RGB]{\colorsecond}31.05 & \cellcolor[RGB]{\colorsecond}31.16 & \cellcolor[RGB]{\colorsecond}31.50 & \cellcolor[RGB]{\colorthird}31.58 \\
% & SplatFields4D + \textbf{Ours} & \cellcolor[RGB]{\colorfirst}31.60 & \cellcolor[RGB]{\colorfirst}31.44 & \cellcolor[RGB]{\colorfirst}31.42 & \cellcolor[RGB]{\colorfirst}31.94 & \cellcolor[RGB]{\colorfirst}31.61 \\
& SplatFields4D + \textbf{Ours} & \cellcolor[RGB]{\colorfirst}32.09 & \cellcolor[RGB]{\colorfirst}31.79 & \cellcolor[RGB]{\colorfirst}31.84 & \cellcolor[RGB]{\colorfirst}32.41 & \cellcolor[RGB]{\colorfirst}32.30 \\
\midrule
&& \multicolumn{5}{c}{SSIM $\uparrow$} \\
&& \textit{Mean} & Dancer & Exercise & Model & Basketball \\
\midrule
\multirow{5}{*}{\rotatebox{90}{}} 
% \multirow{5}{*}{\rotatebox{90}{Splatting}} 
& 4D-GS & 95.34 & \cellcolor[RGB]{\colorthird}95.31 & 95.96 & 94.92 & 95.16 \\
& Deformable3DGS & 93.80 & 94.10 & 95.09 & 91.58 & 94.43 \\
& 4DGaussians & \cellcolor[RGB]{\colorthird}95.91 & 95.19 & \cellcolor[RGB]{\colorthird}96.47 & \cellcolor[RGB]{\colorthird}95.71 & \cellcolor[RGB]{\colorthird}96.28 \\
& SplatFields4D & \cellcolor[RGB]{\colorsecond}96.81 & \cellcolor[RGB]{\colorsecond}96.76 & \cellcolor[RGB]{\colorsecond}96.92 & \cellcolor[RGB]{\colorsecond}96.32 & \cellcolor[RGB]{\colorsecond}97.23 \\
% & SplatFields4D + \textbf{Ours} & \cellcolor[RGB]{\colorfirst}97.11 & \cellcolor[RGB]{\colorfirst}97.10 & \cellcolor[RGB]{\colorfirst}97.11 & \cellcolor[RGB]{\colorfirst}96.75 & \cellcolor[RGB]{\colorfirst}97.47 \\
& SplatFields4D + \textbf{Ours} & \cellcolor[RGB]{\colorfirst}97.34 & \cellcolor[RGB]{\colorfirst}97.31 & \cellcolor[RGB]{\colorfirst}97.31 & \cellcolor[RGB]{\colorfirst}97.10 & \cellcolor[RGB]{\colorfirst}97.65 \\
\midrule
&& \multicolumn{5}{c}{LPIPS $\downarrow$} \\
&& \textit{Mean} & Dancer & Exercise & Model & Basketball \\
\midrule
\multirow{2}{*}{\rotatebox{90}{}}   
& SplatFields4D & 6.01 & 5.90 & 6.62 & 6.30 & 5.21 \\
& SplatFields4D + \textbf{Ours} & \textbf{5.45} & \textbf{5.20} & \textbf{6.20} & \textbf{5.60} & \textbf{4.79}\\ 
\bottomrule
\end{tabular}
}
% \caption{
% \textbf{Sparse-view Dynamic Scene Reconstruction.} We evaluate on 4 sequences from the Owlii dataset, each using 10 input views.
% }

\label{tab:owlii_10views}
\end{table}
\begin{table}[]
  \centering
  \scriptsize
  \setlength{\tabcolsep}{4pt}
\caption{
\textbf{Quantitative Comparison on Sparse-View Dynamic Scene Reconstruction (8 Input Views).} SSIM values are scaled by a factor of 100 for readability; coloring \colorbox[RGB]{\colorfirst}{1st}, \colorbox[RGB]{\colorsecond}{2nd}, and \colorbox[RGB]{\colorthird}{3rd} highlight best results.}
\resizebox{\linewidth}{!}{%
\begin{tabular}{@{}lccccc@{}}
\toprule
\multirow{2}{*}{Method} 
& \multicolumn{5}{c}{PSNR $\uparrow$} \\
& \textit{Mean} & Dancer & Exercise & Model & Basketball \\
\midrule
 4D-GS & 26.20 & 26.99 & 26.34 & 27.41 & 24.07 \\
 Deformable3DGS & 26.06 & 26.77 & 26.24 & 25.61 & 25.62 \\
 4DGaussians & \cellcolor[RGB]{\colorthird}28.16 & \cellcolor[RGB]{\colorthird}27.34 & \cellcolor[RGB]{\colorthird}28.10 & \cellcolor[RGB]{\colorthird}29.60 & \cellcolor[RGB]{\colorthird}27.62 \\
 SplatFields4D & \cellcolor[RGB]{\colorsecond}29.84 & \cellcolor[RGB]{\colorsecond}29.92 & \cellcolor[RGB]{\colorsecond}29.16 & \cellcolor[RGB]{\colorsecond}30.10 & \cellcolor[RGB]{\colorsecond}30.18 \\
 % SplatFields4D + \textbf{Ours} & \cellcolor[RGB]{\colorfirst}30.96 & \cellcolor[RGB]{\colorfirst}31.01 & \cellcolor[RGB]{\colorfirst}30.43 & \cellcolor[RGB]{\colorfirst}31.41 & \cellcolor[RGB]{\colorfirst}31.00 \\
 SplatFields4D + \textbf{Ours} & \cellcolor[RGB]{\colorfirst}30.67 & \cellcolor[RGB]{\colorfirst}30.61 & \cellcolor[RGB]{\colorfirst}30.16 & \cellcolor[RGB]{\colorfirst}31.19 & \cellcolor[RGB]{\colorfirst}30.70 \\ 

\midrule
& \multicolumn{5}{c}{SSIM $\uparrow$} \\
& \textit{Mean} & Dancer & Exercise & Model & Basketball \\
\midrule
 4D-GS & 93.71 & 94.19 & 93.94 & 93.29 & 93.40 \\
Deformable3DGS & 92.37 & 93.24 & 93.29 & 90.33 & 92.62 \\
4DGaussians & \cellcolor[RGB]{\colorthird}95.00 & \cellcolor[RGB]{\colorthird}94.39 & \cellcolor[RGB]{\colorthird}95.45 & \cellcolor[RGB]{\colorthird}94.92 & \cellcolor[RGB]{\colorthird}95.26 \\
 SplatFields4D & \cellcolor[RGB]{\colorsecond}96.28 & \cellcolor[RGB]{\colorsecond}96.31 & \cellcolor[RGB]{\colorsecond}96.26 & \cellcolor[RGB]{\colorsecond}95.78 & \cellcolor[RGB]{\colorsecond}96.77 \\
 % SplatFields4D + \textbf{Ours} & \cellcolor[RGB]{\colorfirst}96.97 & \cellcolor[RGB]{\colorfirst}97.01 & \cellcolor[RGB]{\colorfirst}96.91 & \cellcolor[RGB]{\colorfirst}96.64 & \cellcolor[RGB]{\colorfirst}97.30 \\
 SplatFields4D + \textbf{Ours} & \cellcolor[RGB]{\colorfirst}96.75 & \cellcolor[RGB]{\colorfirst}96.85 & \cellcolor[RGB]{\colorfirst}96.65 & \cellcolor[RGB]{\colorfirst}96.56 & \cellcolor[RGB]{\colorfirst}96.94 \\ 
\midrule
& \multicolumn{5}{c}{LPIPS $\downarrow$} \\
& \textit{Mean} & Dancer & Exercise & Model & Basketball \\
\midrule
SplatFields4D & 6.26 & 6.14 & 6.93 & 6.53 & 5.44  \\
SplatFields4D + \textbf{Ours} & \textbf{6.06} & \textbf{5.70} & \textbf{6.82} & \textbf{6.12} & \textbf{5.60}\\ 
\bottomrule
\end{tabular}
}
% \caption{
% \textbf{Sparse-view Dynamic Scene Reconstruction.} We conduct evaluation on 4 sequences from the Owlii dataset, each using 8 input views. SSIM values are scaled by a factor of 100 for readability.
% }

\label{tab:owlii_8views}
\end{table}

\begin{table}[]
  \centering
  \scriptsize
  \setlength{\tabcolsep}{4pt}
\caption{
\textbf{Quantitative Comparison on Sparse-View Dynamic Scene Reconstruction (6 Input Views).} SSIM values are scaled by a factor of 100 for readability; coloring \colorbox[RGB]{\colorfirst}{1st}, \colorbox[RGB]{\colorsecond}{2nd}, and \colorbox[RGB]{\colorthird}{3rd} highlight best results.}
\resizebox{\linewidth}{!}{%
\begin{tabular}{@{}lccccc@{}}
\toprule
\multirow{2}{*}{Method} 
& \multicolumn{5}{c}{PSNR $\uparrow$} \\
& \textit{Mean} & Dancer & Exercise & Model & Basketball \\
\midrule
4D-GS & 21.42 & 22.89 & 20.80 & 21.60 & 20.40 \\
 Deformable3DGS & 24.46 & 25.37 & 24.31 & 24.12 & 24.02 \\
 4DGaussians & \cellcolor[RGB]{\colorthird}26.52 & \cellcolor[RGB]{\colorthird}26.13 & \cellcolor[RGB]{\colorthird}26.27 & \cellcolor[RGB]{\colorthird}27.34 & \cellcolor[RGB]{\colorthird}26.36 \\
SplatFields4D & \cellcolor[RGB]{\colorsecond}28.36 & \cellcolor[RGB]{\colorsecond}28.84 & \cellcolor[RGB]{\colorsecond}27.54 & \cellcolor[RGB]{\colorsecond}28.67 & \cellcolor[RGB]{\colorsecond}28.39 \\
 SplatFields4D + \textbf{Ours} & \cellcolor[RGB]{\colorfirst}29.19 &  \cellcolor[RGB]{\colorfirst}29.58 &  \cellcolor[RGB]{\colorfirst}28.42 & \cellcolor[RGB]{\colorfirst}29.59 &  \cellcolor[RGB]{\colorfirst}29.19 \\

\midrule
& \multicolumn{5}{c}{SSIM $\uparrow$} \\
& \textit{Mean} & Dancer & Exercise & Model & Basketball \\
\midrule
4D-GS & 87.23 & 89.52 & 87.05 & 85.16 & 87.20 \\
 Deformable3DGS & 90.95 & 91.73 & 91.48 & 89.11 & 91.48 \\
 4DGaussians & \cellcolor[RGB]{\colorthird}93.87 & \cellcolor[RGB]{\colorthird}93.58 & \cellcolor[RGB]{\colorthird}94.45 & \cellcolor[RGB]{\colorthird}93.05 & \cellcolor[RGB]{\colorthird}94.40 \\
 SplatFields4D & \cellcolor[RGB]{\colorsecond}95.69 & \cellcolor[RGB]{\colorsecond}95.99 & \cellcolor[RGB]{\colorsecond}95.71 & \cellcolor[RGB]{\colorsecond}94.85 & \cellcolor[RGB]{\colorsecond}96.21 \\
 SplatFields4D + \textbf{Ours} &  \cellcolor[RGB]{\colorfirst}96.16 &  \cellcolor[RGB]{\colorfirst}96.47 &  \cellcolor[RGB]{\colorfirst}96.05 &  \cellcolor[RGB]{\colorfirst}95.62 &  \cellcolor[RGB]{\colorfirst}96.49 \\ 
\midrule
& \multicolumn{5}{c}{LPIPS $\downarrow$} \\
& \textit{Mean} & Dancer & Exercise & Model & Basketball \\
\midrule
SplatFields4D & 6.60 & 6.26 & 7.16 & 7.27 & 5.73 \\
SplatFields4D + \textbf{Ours} & \textbf{6.36} & \textbf{5.88} & \textbf{7.08} & \textbf{6.75} & \textbf{5.72} \\ 
\bottomrule
\end{tabular}
}

\label{tab:owlii_6views}
\end{table}

\begin{table}[]
 \centering
  \scriptsize
  \setlength{\tabcolsep}{4pt}
\caption{
\textbf{Quantitative Comparison on Sparse-View Dynamic Scene Reconstruction (4 Input Views).}
% We conduct evaluation on 4 sequences from the Owlii dataset, each using 10 input views. 
SSIM and LPIPS values are scaled by a factor of 100 for readability; coloring \colorbox[RGB]{\colorfirst}{1st}, \colorbox[RGB]{\colorsecond}{2nd}, and \colorbox[RGB]{\colorthird}{3rd} highlight best results.
}
\resizebox{\linewidth}{!}{%
\begin{tabular}{@{}lccccc@{}}
\toprule
\multirow{2}{*}{Method} & \multicolumn{5}{c}{PSNR $\uparrow$} \\
& \textit{Mean} & Dancer & Exercise & Model & Basketball \\
\midrule
4D-GS                      & 17.40 & 17.70 & 16.86 & 18.35 & 16.71 \\
Deformable3DGS & 20.04 & 21.42 & 19.56 & 19.71 & 19.45 \\
4DGaussians   & \cellcolor[RGB]{\colorthird}21.31 & \cellcolor[RGB]{\colorthird}21.49 & \cellcolor[RGB]{\colorsecond}21.05 & \cellcolor[RGB]{\colorsecond}21.90 & \cellcolor[RGB]{\colorthird}20.80 \\
SplatFields4D & \cellcolor[RGB]{\colorsecond}21.95 & \cellcolor[RGB]{\colorsecond}22.83 & \cellcolor[RGB]{\colorthird}20.76 & \cellcolor[RGB]{\colorthird}21.83 & \cellcolor[RGB]{\colorsecond}22.39 \\
SplatFields4D + \textbf{Ours} &   \cellcolor[RGB]{\colorfirst}25.05 &  \cellcolor[RGB]{\colorfirst}26.19 &  \cellcolor[RGB]{\colorfirst}24.54 &  \cellcolor[RGB]{\colorfirst}23.65 &  \cellcolor[RGB]{\colorfirst} 25.81 \\
\midrule
& \multicolumn{5}{c}{SSIM $\uparrow$} \\
& \textit{Mean} & Dancer & Exercise & Model & Basketball \\
\midrule
4D-GS                     & 78.94 & 80.81 & 78.72 & 78.09 & 78.15 \\
Deformable3DGS & 87.10 & 89.04 & 87.67 & 85.05 & 86.63 \\
4DGaussians  & \cellcolor[RGB]{\colorthird}89.50 & \cellcolor[RGB]{\colorthird}90.34 & \cellcolor[RGB]{\colorthird}90.67 & \cellcolor[RGB]{\colorthird}87.47 & \cellcolor[RGB]{\colorthird}89.51 \\
SplatFields4D & \cellcolor[RGB]{\colorsecond}91.60 & \cellcolor[RGB]{\colorsecond}92.89 & \cellcolor[RGB]{\colorsecond}92.07 & \cellcolor[RGB]{\colorsecond}89.02 & \cellcolor[RGB]{\colorsecond}92.41 \\
SplatFields4D + \textbf{Ours} &  \cellcolor[RGB]{\colorfirst}93.19 &  \cellcolor[RGB]{\colorfirst}94.50 &  \cellcolor[RGB]{\colorfirst}93.62 &  \cellcolor[RGB]{\colorfirst}90.61 &  \cellcolor[RGB]{\colorfirst}94.02 \\

\midrule
& \multicolumn{5}{c}{LPIPS $\downarrow$} \\
& \textit{Mean} & Dancer & Exercise & Model & Basketball \\
\midrule
SplatFields4D & 9.86 & 8.82 & 10.23 & 11.27 & 9.14 \\
SplatFields4D + \textbf{Ours} & \textbf{ 8.91} & \textbf{7.63} & \textbf{9.60} & \textbf{10.23} & \textbf{8.20 }\\ 
\bottomrule
\end{tabular}
}

% \vspace{-12.5em}
\label{tab:owlii_4views}
\end{table}

\boldparagraph{Comparison with TexturedGaussians}
We further demonstrate the advantages of our neural texture modeling by comparing it to a concurrent method, Textured Gaussians~\cite{chao2024textured}, which employs per-primitive 2D RGBA textures to enhance 3DGS performance.
We report its performance metrics as stated in the original paper and estimate its model size based on the reported number of primitives and texture resolution.
As shown in Table~\ref{tab:cmp_with_textured_gs}, our method achieves significantly higher performance while requiring a much smaller model size, resulting in substantial reductions in storage overhead.

\section{Additional Quantitative Results}
We provided complete benchmarking for sparse-view reconstruction of the static scene on the Blender and Owlii datasets.

We also provide a detailed per-scene breakdown and additional baseline comparisons for surface reconstruction on DTU~\cite{aanaes2016DTU} dataset in Table~\ref{tab:dtu_result}. Our method achieves state-of-the-art performance among baselines~\cite{kerbl20233dgs, guedon2024sugar, Dai2024GaussianSurfels, Huang2DGS2024, Yu2024GOF}.

\begin{table}[]
\centering
\scriptsize
\setlength{\tabcolsep}{2.5pt}
\caption{\textbf{Quantitative Comparison on Sparse-View Novel View Synthesis in Blender (10 Input Views)}. SSIM values are scaled by a factor of 100 for readability; coloring \colorbox[RGB]{\colorfirst}{1st}, \colorbox[RGB]{\colorsecond}{2nd}, and \colorbox[RGB]{\colorthird}{3rd} highlight best results. 
}
\begin{tabular}{@{}lccccccccc@{}}
\toprule
\multirow{2}{*}{Method} & \multicolumn{9}{c}{PSNR $\uparrow$} \\
& \textit{mean} & Toy & Ficus & Hotdog & Chair & Mic & Ship & Drums & Materials \\
\midrule
SparseNeRF & - & 22.64 & 18.27 & - & 25.30 & 23.27 & 20.29 & 18.61 & 19.72 \\
SparseNeRF wo. depth & 22.58 & 23.89 & 18.75 & 27.56 & \cellcolor[RGB]{\colorthird}26.42 & 23.23 & 21.68 & 18.20 & \cellcolor[RGB]{\colorthird}20.87 \\
SuGaR & 21.10 & 22.78 & 22.42 & 23.60 & 24.25 & 17.93 & 20.35 & 19.11 & 18.40 \\
ScaffoldGS & 22.63 & 21.98 & 22.68 & 24.37 & 24.15 & \cellcolor[RGB]{\colorsecond}27.76 & 20.39 & 19.64 & 20.08 \\
Mip-Splatting & 23.65 & 23.49 & 24.97 & 25.27 & 24.49 & \cellcolor[RGB]{\colorthird}27.69 & 21.38 & 21.23 & 20.66 \\
3DGS & 24.11 & 23.79 & 25.54 & 26.16 & 25.28 & \cellcolor[RGB]{\colorfirst}28.39 & \cellcolor[RGB]{\colorthird}21.87 & 21.34 & 20.51 \\
LightGaussian & 24.21 & 23.94 & \cellcolor[RGB]{\colorfirst}26.95 & 25.62 & 25.91 & 27.45 & 21.82 & 21.38 & 20.60 \\
2DGS & \cellcolor[RGB]{\colorthird}24.42 & \cellcolor[RGB]{\colorthird}24.06 & 25.17 & \cellcolor[RGB]{\colorthird}27.92 & \cellcolor[RGB]{\colorfirst}26.96 & 27.53 & 21.83 & \cellcolor[RGB]{\colorthird}21.58 & 20.27 \\
SplatFields3D & \cellcolor[RGB]{\colorsecond}24.94 & \cellcolor[RGB]{\colorsecond}26.51 & \cellcolor[RGB]{\colorthird}25.59 & \cellcolor[RGB]{\colorsecond}28.29 & 25.92 & 27.36 & \cellcolor[RGB]{\colorsecond}23.12 & \cellcolor[RGB]{\colorsecond}21.86 & \cellcolor[RGB]{\colorsecond}20.88 \\
% SplatFields3D + \textbf{Ours} & \cellcolor[RGB]{\colorfirst}25.26 & \cellcolor[RGB]{\colorfirst}26.73 & 25.43 & \cellcolor[RGB]{\colorfirst}28.96 & \cellcolor[RGB]{\colorsecond}26.83 & 27.30 & \cellcolor[RGB]{\colorfirst}23.39 & \cellcolor[RGB]{\colorfirst}22.13 & \cellcolor[RGB]{\colorfirst}21.33 \\
SplatFields3D + \textbf{Ours} & \cellcolor[RGB]{\colorfirst}25.43 & \cellcolor[RGB]{\colorfirst}27.11 & \cellcolor[RGB]{\colorsecond}26.09 & \cellcolor[RGB]{\colorfirst}28.77 &  \cellcolor[RGB]{\colorsecond}26.84 & 27.48 & \cellcolor[RGB]{\colorfirst}23.63 & \cellcolor[RGB]{\colorfirst}22.22 & \cellcolor[RGB]{\colorfirst}21.30 \\
\midrule
& \multicolumn{9}{c}{SSIM $\uparrow$} \\
& \textit{mean} & Toy & Ficus & Hotdog & Chair & Mic & Ship & Drums & Materials \\
\midrule
SparseNeRF & - & 85.65 & 84.30 & - & 89.57 & 92.27 & 75.51 & 82.52 & 83.82 \\
SparseNeRF wo. depth & 86.95 & \cellcolor[RGB]{\colorthird}88.45 & 84.65 & \cellcolor[RGB]{\colorthird}94.06 & 91.26 & 92.39 & \cellcolor[RGB]{\colorthird}77.25 & 82.12 & \cellcolor[RGB]{\colorsecond}85.42 \\
SuGaR & 83.83 & 84.27 & 88.55 & 90.52 & 88.77 & 84.39 & 74.27 & 80.78 & 79.10 \\
ScaffoldGS & 85.45 & 83.27 & 90.14 & 89.66 & 88.56 & 95.77 & 70.66 & 83.46 & 82.10 \\
Mip-Splatting & 88.00 & 85.68 & 93.15 & 91.76 & 90.34 & \cellcolor[RGB]{\colorsecond}96.44 & 73.98 & 88.22 & 84.41 \\
3DGS & 88.27 & 86.30 & 93.65 & 92.19 & 91.23 & \cellcolor[RGB]{\colorfirst}96.49 & 73.89 & 88.48 & 83.93 \\
LightGaussian & 88.76 & 86.66 & \cellcolor[RGB]{\colorfirst}94.82 & 92.41 & \cellcolor[RGB]{\colorsecond}91.70 & 96.27 & 74.37 & 88.75 & 85.12 \\
2DGS & \cellcolor[RGB]{\colorthird}89.51 & 88.42 & 93.61 & 93.79 & \cellcolor[RGB]{\colorfirst}93.26 & \cellcolor[RGB]{\colorthird}96.37 & 76.52 & \cellcolor[RGB]{\colorthird}89.38 & 84.72 \\
SplatFields3D & \cellcolor[RGB]{\colorsecond}90.32 & \cellcolor[RGB]{\colorsecond}91.34 & \cellcolor[RGB]{\colorthird}93.70 & \cellcolor[RGB]{\colorsecond}95.09 & 91.14 & 95.95 & \cellcolor[RGB]{\colorsecond}80.25 & \cellcolor[RGB]{\colorsecond}89.85 & \cellcolor[RGB]{\colorthird}85.26 \\
% SplatFields3D + \textbf{Ours} & \cellcolor[RGB]{\colorfirst}90.57 & \cellcolor[RGB]{\colorsecond}90.96 & \cellcolor[RGB]{\colorsecond}93.86 & \cellcolor[RGB]{\colorfirst}95.34 & \cellcolor[RGB]{\colorsecond}92.23 & 95.86 & \cellcolor[RGB]{\colorfirst}80.37 & \cellcolor[RGB]{\colorfirst}90.12 & \cellcolor[RGB]{\colorfirst}85.85 \\
SplatFields3D + \textbf{Ours} & \cellcolor[RGB]{\colorfirst}90.73 & \cellcolor[RGB]{\colorfirst}91.75 & \cellcolor[RGB]{\colorsecond} 94.25 & \cellcolor[RGB]{\colorfirst}95.26 & \cellcolor[RGB]{\colorthird}91.49 & 95.98 & \cellcolor[RGB]{\colorfirst}80.90 & \cellcolor[RGB]{\colorfirst}90.27 & \cellcolor[RGB]{\colorfirst}85.94 \\
\midrule
& \multicolumn{9}{c}{LPIPS $\downarrow$} \\
& \textit{mean} & Toy & Ficus & Hotdog & Chair & Mic & Ship & Drums & Materials \\
\midrule
SplatFields3D & \textbf{10.96} &\textbf{ 9.28} & 6.29 &\textbf{ 7.57} & 9.74 & \textbf{4.15} & \textbf{26.07} & \textbf{9.56} & 15.03 \\ 
SplatFields3D + \textbf{Ours} & 11.05 & 9.74 & \textbf{5.38 }& 7.92 & \textbf{9.68} & 4.27 & 26.67 & 9.98 & \textbf{14.78 }\\ 
\bottomrule
\end{tabular}
% \caption{Quantitative comparisons on sparse-view novel view synthesis in Blender (10 input views). SSIM values are scaled by a factor of 100 for readability.}

\label{tab:sparseview_results_10views}
\end{table}

\begin{table}[]
\centering
\scriptsize
\setlength{\tabcolsep}{2.5pt}
\caption{\textbf{Quantitative Comparison on Sparse-View Novel View Synthesis in Blender (8 Input Views).} SSIM values are scaled by a factor of 100 for readability; coloring \colorbox[RGB]{\colorfirst}{1st}, \colorbox[RGB]{\colorsecond}{2nd}, and \colorbox[RGB]{\colorthird}{3rd} highlight best results.
}
\begin{tabular}{@{}lccccccccc@{}}
\toprule
\multirow{2}{*}{Method} & \multicolumn{9}{c}{PSNR $\uparrow$} \\
& \textit{mean} & Toy & Ficus & Hotdog & Chair & Mic & Ship & Drums & Materials \\
\midrule
SparseNeRF & - & 22.33 & 17.97 & - & 23.81 & 23.01 & 19.85 & 17.85 & \cellcolor[RGB]{\colorthird}20.02 \\
SparseNeRF wo. depth & 22.20 & \cellcolor[RGB]{\colorthird}24.06 & 18.42 & \cellcolor[RGB]{\colorthird}27.09 & 25.12 & 23.04 & \cellcolor[RGB]{\colorthird}21.23 & 17.94 & \cellcolor[RGB]{\colorfirst}20.74 \\
SuGaR & 20.62 & 21.91 & 22.33 & 23.01 & 23.30 & 18.60 & 19.59 & 18.66 & 17.55 \\
ScaffoldGS & 21.53 & 20.95 & 21.35 & 23.77 & 22.77 & 26.40 & 18.88 & 18.96 & 19.17 \\
Mip-Splatting & 22.37 & 22.05 & 23.23 & 24.24 & 23.57 & 26.32 & 19.91 & 20.10 & 19.55 \\
3DGS & 22.93 & 22.55 & \cellcolor[RGB]{\colorthird}23.69 & 25.57 & 24.43 & \cellcolor[RGB]{\colorfirst}27.37 & 19.98 & 20.33 & 19.49 \\
LightGaussian & 22.98 & 22.67 & \cellcolor[RGB]{\colorfirst}24.98 & 24.79 & 24.40 & \cellcolor[RGB]{\colorsecond}26.59 & 20.60 & 20.41 & 19.41 \\
2DGS & \cellcolor[RGB]{\colorthird}23.04 & 22.19 & 23.63 & 26.76 & \cellcolor[RGB]{\colorsecond}25.46 & 26.24 & 20.16 & \cellcolor[RGB]{\colorthird}20.60 & 19.25 \\
SplatFields3D & \cellcolor[RGB]{\colorsecond}23.98 & \cellcolor[RGB]{\colorsecond}24.71 & \cellcolor[RGB]{\colorsecond}23.97 & \cellcolor[RGB]{\colorsecond}27.87 & \cellcolor[RGB]{\colorfirst}25.64 & 26.49 & \cellcolor[RGB]{\colorsecond}22.15 & \cellcolor[RGB]{\colorsecond}21.12 & 19.85 \\
SplatFields3D + \textbf{Ours} & \cellcolor[RGB]{\colorfirst}24.09 & \cellcolor[RGB]{\colorfirst}25.40 & 23.16 & \cellcolor[RGB]{\colorfirst}27.97 & \cellcolor[RGB]{\colorthird}25.30 & \cellcolor[RGB]{\colorthird}26.54 & \cellcolor[RGB]{\colorfirst}22.40 & \cellcolor[RGB]{\colorfirst}21.24 & \cellcolor[RGB]{\colorsecond}20.67 \\
\midrule
& \multicolumn{9}{c}{SSIM $\uparrow$} \\
& \textit{mean} & Toy & Ficus & Hotdog & Chair & Mic & Ship & Drums & Materials \\
\midrule
SparseNeRF & - & 84.85 & 83.98 & - & 88.39 & 92.06 & 74.32 & 81.44 & \cellcolor[RGB]{\colorthird}83.66 \\
SparseNeRF wo. depth & 86.30 & \cellcolor[RGB]{\colorsecond}88.27 & 83.97 & \cellcolor[RGB]{\colorthird}93.65 & 90.23 & 92.15 & \cellcolor[RGB]{\colorthird}76.01 & 81.38 & \cellcolor[RGB]{\colorsecond}84.78 \\
SuGaR & 82.74 & 82.04 & 87.95 & 89.41 & 87.60 & 85.08 & 72.80 & 79.84 & 77.23 \\
ScaffoldGS & 83.62 & 80.54 & 88.25 & 89.04 & 86.23 & 95.01 & 68.49 & 81.49 & 79.91 \\
Mip-Splatting& 86.24 & 82.93 & 91.03 & 90.84 & 89.21 & \cellcolor[RGB]{\colorsecond}95.77 & 71.86 & 86.15 & 82.14 \\
3DGS & 86.63 & 84.05 & 91.56 & 91.63 & \cellcolor[RGB]{\colorthird}90.49 & \cellcolor[RGB]{\colorfirst}95.99 & 70.96 & 86.62 & 81.76 \\
LightGaussian & 87.11 & 84.44 & \cellcolor[RGB]{\colorfirst}93.01 & 91.47 & 90.17 & \cellcolor[RGB]{\colorthird}95.76 & 72.16 & 86.97 & 82.92 \\
2DGS & \cellcolor[RGB]{\colorthird}87.72 & 84.80 & \cellcolor[RGB]{\colorsecond}91.74 & 93.13 & \cellcolor[RGB]{\colorfirst}91.79 & 95.72 & 74.06 & \cellcolor[RGB]{\colorthird}87.89 & 82.65 \\
SplatFields3D & \cellcolor[RGB]{\colorsecond}88.94 & \cellcolor[RGB]{\colorthird}88.04 & \cellcolor[RGB]{\colorthird}91.69 & \cellcolor[RGB]{\colorsecond}94.68 & \cellcolor[RGB]{\colorsecond}90.91 & 95.48 & \cellcolor[RGB]{\colorsecond}78.76 & \cellcolor[RGB]{\colorsecond}88.50 & 83.46 \\
SplatFields3D + \textbf{Ours} & \cellcolor[RGB]{\colorfirst}89.14 & \cellcolor[RGB]{\colorfirst}88.71 & 91.01 & \cellcolor[RGB]{\colorfirst}94.77 & 90.38 & 95.27 & \cellcolor[RGB]{\colorfirst}78.79 & \cellcolor[RGB]{\colorfirst}88.73 & \cellcolor[RGB]{\colorfirst}85.47 \\
\midrule
& \multicolumn{9}{c}{LPIPS $\downarrow$} \\
& \textit{mean} & Toy & Ficus & Hotdog & Chair & Mic & Ship & Drums & Materials \\
\midrule
SplatFields3D & \textbf{11.92} & 12.14 &\textbf{ 7.93} & \textbf{8.29} & \textbf{10.08} &\textbf{ 4.52} &\textbf{ 25.91} & \textbf{10.28} &\textbf{ 16.21} \\ 
SplatFields3D + \textbf{Ours} & 12.70 & \textbf{12.02} & 9.29 & 8.45 & 11.07 & 5.23 & 27.96 & 10.83 & 16.78 \\
\bottomrule
\end{tabular}
% \caption{Quantitative comparisons on sparse-view novel view synthesis in Blender (8 input views). SSIM values are scaled by a factor of 100 for readability.}

\label{tab:sparseview_results_8views}
\end{table}

\begin{table}[]
\centering
\scriptsize
\setlength{\tabcolsep}{2.5pt}
\caption{\textbf{Quantitative Comparison on Sparse-View Novel View Synthesis in Blender (6 Input Views)}. SSIM values are scaled by a factor of 100 for readability; coloring \colorbox[RGB]{\colorfirst}{1st}, \colorbox[RGB]{\colorsecond}{2nd}, and \colorbox[RGB]{\colorthird}{3rd} highlight best results.
}
\begin{tabular}{@{}lccccccccc@{}}
\toprule
\multirow{2}{*}{Method} & \multicolumn{9}{c}{PSNR $\uparrow$} \\
& \textit{mean} & Toy & Ficus & Hotdog & Chair & Mic & Ship & Drums & Materials \\
\midrule
SparseNeRF & - & 20.86 & 18.03 & - & 22.75 & 22.40 & \cellcolor[RGB]{\colorthird}19.33 & 16.24 & \cellcolor[RGB]{\colorsecond}19.54 \\
SparseNeRF wo. depth & \cellcolor[RGB]{\colorthird}20.86 & \cellcolor[RGB]{\colorsecond}22.62 & 17.63 & \cellcolor[RGB]{\colorthird}25.84 & 22.65 & 20.72 & \cellcolor[RGB]{\colorsecond}19.85 & 17.25 & \cellcolor[RGB]{\colorfirst}20.30 \\
SuGaR & 19.07 & 19.89 & 20.61 & 20.80 & 21.92 & 18.26 & 17.72 & 16.86 & 16.53 \\
ScaffoldGS & 19.65 & 18.21 & 20.72 & 19.48 & 22.20 & 24.31 & 16.47 & 17.21 & 18.62 \\
Mip-Splatting & 20.04 & 19.39 & 21.81 & 19.70 & 21.72 & 24.44 & 17.02 & 17.72 & 18.52 \\
3DGS & 20.62 & 19.80 & 22.25 & 21.16 & 22.75 & \cellcolor[RGB]{\colorfirst}25.21 & 17.58 & 17.77 & 18.48 \\
LightGaussian & 20.76 & 20.25 & \cellcolor[RGB]{\colorfirst}23.12 & 20.66 & 22.69 & \cellcolor[RGB]{\colorthird}24.89 & 17.83 & \cellcolor[RGB]{\colorthird}18.02 & 18.63 \\
2DGS & 20.74 & 19.38 & 21.93 & 23.85 & \cellcolor[RGB]{\colorthird}23.26 & 24.48 & 16.92 & 17.91 & 18.17 \\
SplatFields3D & \cellcolor[RGB]{\colorsecond}22.26 & \cellcolor[RGB]{\colorthird}22.41 & \cellcolor[RGB]{\colorthird}22.26 & \cellcolor[RGB]{\colorsecond}26.19 & \cellcolor[RGB]{\colorfirst}25.03 & 24.84 & \cellcolor[RGB]{\colorthird}19.33 & \cellcolor[RGB]{\colorsecond}18.97 & 19.05 \\
SplatFields3D + \textbf{Ours} & \cellcolor[RGB]{\colorfirst}22.72 & \cellcolor[RGB]{\colorfirst}23.50 & \cellcolor[RGB]{\colorsecond}22.45 & \cellcolor[RGB]{\colorfirst}26.94 & \cellcolor[RGB]{\colorsecond}24.31 & \cellcolor[RGB]{\colorsecond}25.05 & \cellcolor[RGB]{\colorfirst}21.19 & \cellcolor[RGB]{\colorfirst}19.19 & \cellcolor[RGB]{\colorthird}19.12 \\
\midrule
& \multicolumn{9}{c}{SSIM $\uparrow$} \\
& \textit{mean} & Toy & Ficus & Hotdog & Chair & Mic & Ship & Drums & Materials \\
\midrule
SparseNeRF & - & 83.25 & 83.70 & - & 87.76 & 91.53 & \cellcolor[RGB]{\colorthird}72.96 & 78.58 & \cellcolor[RGB]{\colorsecond}83.47 \\
SparseNeRF wo. depth & 84.42 & \cellcolor[RGB]{\colorsecond}85.68 & 82.33 & \cellcolor[RGB]{\colorthird}92.51 & 87.47 & 90.72 & 72.50 & 79.58 & \cellcolor[RGB]{\colorfirst}84.59 \\
SuGaR & 79.85 & 77.67 & 85.51 & 86.44 & 84.98 & 84.24 & 68.93 & 76.35 & 74.68 \\
ScaffoldGS & 80.34 & 74.13 & 87.01 & 82.79 & 84.99 & 93.10 & 62.49 & 78.61 & 79.59 \\
Mip-Splatting & 83.09 & 77.68 & 89.34 & 86.81 & 86.46 & 94.58 & 66.45 & 82.08 & 81.34 \\
3DGS & 83.56 & 78.58 & \cellcolor[RGB]{\colorthird}89.79 & 87.81 & 87.35 & \cellcolor[RGB]{\colorfirst}94.81 & 66.71 & 82.45 & 80.94 \\
LightGaussian & 84.34 & 79.64 & \cellcolor[RGB]{\colorfirst}91.08 & 88.67 & 87.49 & \cellcolor[RGB]{\colorsecond}94.76 & 67.63 & 83.09 & 82.33 \\
2DGS &  \cellcolor[RGB]{\colorthird}84.43 & 78.54 & 89.71 & 90.36 & \cellcolor[RGB]{\colorthird}88.16 & \cellcolor[RGB]{\colorthird}94.59 & 68.63 & \cellcolor[RGB]{\colorthird}83.87 & 81.61 \\
% 3DGS w. Moran & \cellcolor[RGB]{\colorthird}84.54 & 80.20 & 90.84 & 90.15 & 87.76 & 94.50 & 67.49 & 83.47 & 81.89 \\
SplatFields3D & \cellcolor[RGB]{\colorsecond}86.62 & \cellcolor[RGB]{\colorthird}84.05 & 89.56 & \cellcolor[RGB]{\colorsecond}93.62 & \cellcolor[RGB]{\colorfirst}89.53 & 94.50 & \cellcolor[RGB]{\colorsecond}74.14 & \cellcolor[RGB]{\colorsecond}85.03 & \cellcolor[RGB]{\colorthird}82.55 \\
SplatFields3D + \textbf{Ours} & \cellcolor[RGB]{\colorfirst}87.40 & \cellcolor[RGB]{\colorfirst}86.22 & \cellcolor[RGB]{\colorsecond}89.95 & \cellcolor[RGB]{\colorfirst}94.00 & \cellcolor[RGB]{\colorsecond}89.23 & 94.49 & \cellcolor[RGB]{\colorfirst}77.27 & \cellcolor[RGB]{\colorfirst}85.52 & 82.53 \\
\midrule
& \multicolumn{9}{c}{LPIPS $\downarrow$} \\
& \textit{mean} & Toy & Ficus & Hotdog & Chair & Mic & Ship & Drums & Materials \\
\midrule
SplatFields3D & 13.50 & 14.46 & 9.47 & \textbf{9.64} & \textbf{10.53} & 5.48 & 28.32 & 13.36 & \textbf{16.76} \\
SplatFields3D + \textbf{Ours} & \textbf{13.32 }& \textbf{13.87 }& \textbf{8.86} & 9.72 & 11.02 & \textbf{5.44} & \textbf{27.64} & \textbf{13.05} & 16.95  \\ 
\bottomrule

\end{tabular}
% \caption{Quantitative comparisons on sparse-view novel view synthesis in Blender (6 input views). SSIM values are scaled by a factor of 100 for readability.}

\label{tab:sparseview_results_6views}
\end{table}
\begin{table}[]
     \centering
\scriptsize
\setlength{\tabcolsep}{2.5pt}
\caption{\textbf{Quantitative Comparison on Sparse-View Novel View Synthesis in Blender (4 Input Views)}.  SSIM and LPIPS values are scaled by a factor of 100 for readability; coloring \colorbox[RGB]{\colorfirst}{1st}, \colorbox[RGB]{\colorsecond}{2nd}, and \colorbox[RGB]{\colorthird}{3rd} highlight best results.
% SSIM values are scaled by a factor of 100 for readability.
}
\begin{tabular}{@{}lccccccccc@{}}
\toprule
\multirow{2}{*}{Method} & \multicolumn{9}{c}{PSNR $\uparrow$} \\
& \textit{mean} & Toy & Ficus & Hotdog & Chair & Mic & Ship & Drums & Materials \\
\midrule
SparseNeRF & - & \cellcolor[RGB]{\colorfirst}20.94 & 17.48 & \cellcolor[RGB]{\colorsecond}23.81 & \cellcolor[RGB]{\colorfirst}21.41 & \cellcolor[RGB]{\colorthird}21.52 & - & 15.37 & \cellcolor[RGB]{\colorthird}17.03 \\
SparseNeRF \textit{wo.} depth & \cellcolor[RGB]{\colorthird}17.87 & \cellcolor[RGB]{\colorsecond}19.31 & 17.05 & 23.54 & \cellcolor[RGB]{\colorsecond}20.26 & 11.56 & \cellcolor[RGB]{\colorfirst}17.86 & 13.64 & \cellcolor[RGB]{\colorfirst}19.77 \\
SuGaR & 16.94 & 16.96 & 19.30 & 19.36 & 19.07 & 17.47 & 15.22 & 14.73 & 13.38 \\
ScaffoldGS & 16.86 & 15.40 & 19.58 & 17.31 & 18.40 & 20.54 & 14.70 & 15.27 & 13.69 \\
Mip-Splatting & 16.94 & 16.23 & 19.60 & 16.98 & 18.38 & 20.56 & 14.64 & 14.92 & 14.21 \\
3DGS & 17.37 & 16.44 & 19.72 & 18.65 & 18.72 & 20.75 & 15.43 & 15.08 & 14.15 \\
LightGaussian & 17.70 & 16.94 & \cellcolor[RGB]{\colorsecond}20.35 & 18.56 & 18.96 & 21.53 & 15.67 & \cellcolor[RGB]{\colorthird}15.44 & 14.19 \\
2DGS & 17.58 & 16.32 & 19.69 & 20.67 & 19.39 & 21.17 & 14.45 & 14.84 & 14.14 \\
SplatFields3D & \cellcolor[RGB]{\colorsecond}19.16 & 18.89 & \cellcolor[RGB]{\colorthird}20.19 & \cellcolor[RGB]{\colorfirst}24.31 & 19.31 & \cellcolor[RGB]{\colorsecond}21.73 & \cellcolor[RGB]{\colorthird}16.83 & \cellcolor[RGB]{\colorsecond}16.35 & 15.69 \\
SplatFields3D + \textbf{Ours} & \cellcolor[RGB]{\colorfirst}19.98 & \cellcolor[RGB]{\colorthird}18.94 & \cellcolor[RGB]{\colorfirst}20.39 & \cellcolor[RGB]{\colorthird}23.79 & \cellcolor[RGB]{\colorthird}20.26 & \cellcolor[RGB]{\colorfirst}23.75 & \cellcolor[RGB]{\colorsecond}17.34 & \cellcolor[RGB]{\colorfirst}17.38 & \cellcolor[RGB]{\colorsecond}17.97 \\
\midrule
& \multicolumn{9}{c}{SSIM $\uparrow$} \\
& \textit{mean} & Toy & Ficus & Hotdog & Chair & Mic & Ship & Drums & Materials \\
\midrule
SparseNeRF & - & \cellcolor[RGB]{\colorfirst}83.38 & 82.98 & \cellcolor[RGB]{\colorthird}90.95 & \cellcolor[RGB]{\colorsecond}85.14 & 90.49 & - & \cellcolor[RGB]{\colorthird}76.46 & \cellcolor[RGB]{\colorthird}79.48 \\
SparseNeRF \textit{wo.} depth & 78.66 & \cellcolor[RGB]{\colorsecond}78.80 & 80.86 & 90.57 & 82.26 & 72.23 & \cellcolor[RGB]{\colorthird}69.29 & 71.97 & \cellcolor[RGB]{\colorfirst}83.28 \\
SuGaR & 75.61 & 72.07 & 83.08 & 83.68 & 80.66 & 82.48 & 63.46 & 70.73 & 68.69 \\
ScaffoldGS & 74.99 & 68.04 & 84.68 & 76.92 & 77.58 & 90.24 & 59.60 & 71.20 & 71.68 \\
Mip-Splatting & 77.67 & 71.47 & 85.73 & 82.68 & 81.00 & 91.42 & 61.17 & 74.27 & 73.62 \\
3DGS & 78.12 & 72.17 & 85.97 & 83.78 & 81.49 & 91.55 & 62.05 & 74.80 & 73.18 \\
LightGaussian & \cellcolor[RGB]{\colorthird}79.38 & 73.58 & \cellcolor[RGB]{\colorsecond}86.93 & 85.98 & 81.91 & \cellcolor[RGB]{\colorsecond}92.25 & 63.27 & 75.88 & 75.22 \\
2DGS & 79.26 & 72.84 & 86.04 & 86.62 & 82.54 & \cellcolor[RGB]{\colorthird}92.04 & 63.57 & 76.25 & 74.14 \\
SplatFields3D & \cellcolor[RGB]{\colorsecond}82.26 & \cellcolor[RGB]{\colorthird}78.23 & \cellcolor[RGB]{\colorthird}86.17 & \cellcolor[RGB]{\colorfirst}92.10 & \cellcolor[RGB]{\colorthird}83.85 & 91.92 & \cellcolor[RGB]{\colorsecond}70.40 & \cellcolor[RGB]{\colorsecond}78.67 & 76.77 \\
SplatFields3D + \textbf{Ours} & \cellcolor[RGB]{\colorfirst}83.51 & 77.33 & \cellcolor[RGB]{\colorfirst}87.48 & \cellcolor[RGB]{\colorsecond}91.92 & \cellcolor[RGB]{\colorfirst}85.24 & \cellcolor[RGB]{\colorfirst}93.46 & \cellcolor[RGB]{\colorfirst}71.51 & \cellcolor[RGB]{\colorfirst}80.33 & \cellcolor[RGB]{\colorsecond}80.85 \\
\midrule
& \multicolumn{9}{c}{LPIPS $\downarrow$} \\
& \textit{mean} & Toy & Ficus & Hotdog & Chair & Mic & Ship & Drums & Materials \\
\midrule
SplatFields3D & 17.56 & \textbf{19.56} & 12.20 & \textbf{11.41} & 16.16 & 8.35 & \textbf{31.97} & 19.06 & 21.76 \\
SplatFields3D + \textbf{Ours} & \textbf{16.93} & 20.49 & \textbf{9.34} & 12.10 & \textbf{15.81} & \textbf{6.89} & 32.72 & \textbf{17.82} & \textbf{ 20.28} \\  
\bottomrule
\end{tabular}

\label{tab:sparseview_results_4views}
\end{table}

\begin{table*}[h]
\setlength\tabcolsep{0.5em}
\centering
\small
\begin{tabular}{@{}lccccccccccccccccc@{}}

\toprule
& \multicolumn{15}{c}{Chamfer Distance per Scan $\downarrow$} &  \\
Method & 24 & 37 & 40 & 55 & 63 & 65 & 69 & 83 & 97 & 105 & 106 & 110 & 114 & 118 & 122 & \textit{Mean}\\
\midrule
% \cline{1-17}
3DGS & 2.14 & 1.53 & 2.08 & 1.68 & 3.49 & 2.21 & 1.43 & 2.07 & 2.22 & 1.75 & 1.79 & 2.55 & 1.53 & 1.52 & 1.50 & 1.96 \\
SuGaR & 1.47 & 1.33 & 1.13 & 0.61 & 2.25 & 1.71 & 1.15 & 1.63 & 1.62 & 1.07 & 0.79 & 2.45 & 0.98 & 0.88 & 0.79 & 1.33 \\
GaussianSurfels & 0.66 & 0.93 & 0.54 & 0.41 & \cellcolor[RGB]{\colorthird}1.06 & 1.14 & 0.85 & \cellcolor[RGB]{\colorthird}1.29 & 1.53 & 0.79 & 0.82 & 1.58 & 0.45 & \cellcolor[RGB]{\colorsecond}0.66 & 0.53 & 0.88 \\
2DGS & \cellcolor[RGB]{\colorsecond}0.48 & \cellcolor[RGB]{\colorthird}0.91 & \cellcolor[RGB]{\colorthird}0.39 & \cellcolor[RGB]{\colorthird}0.39 & \cellcolor[RGB]{\colorsecond}1.01 & \cellcolor[RGB]{\colorthird}0.83 & \cellcolor[RGB]{\colorthird}0.81 & 1.36 & \cellcolor[RGB]{\colorsecond}1.27 & \cellcolor[RGB]{\colorthird}0.76 & \cellcolor[RGB]{\colorsecond}0.70 & \cellcolor[RGB]{\colorthird}1.40 & \cellcolor[RGB]{\colorsecond}0.40 & 0.76 & \cellcolor[RGB]{\colorthird}0.52 & \cellcolor[RGB]{\colorthird}0.80 \\
GOF & \cellcolor[RGB]{\colorthird}0.50 & \cellcolor[RGB]{\colorsecond}0.82 & \cellcolor[RGB]{\colorfirst}0.37 & \cellcolor[RGB]{\colorfirst}0.37 & 1.12 & \cellcolor[RGB]{\colorfirst}0.74 & \cellcolor[RGB]{\colorsecond}0.73 & \cellcolor[RGB]{\colorsecond}1.18 & \cellcolor[RGB]{\colorthird}1.29 & \cellcolor[RGB]{\colorsecond}0.68 & \cellcolor[RGB]{\colorthird}0.77 & \cellcolor[RGB]{\colorsecond}0.90 & \cellcolor[RGB]{\colorthird}0.42 & \cellcolor[RGB]{\colorsecond}0.66 & \cellcolor[RGB]{\colorsecond}0.49 & \cellcolor[RGB]{\colorsecond}0.74 \\
GOF + \textbf{Ours} & \cellcolor[RGB]{\colorfirst}0.44 & \cellcolor[RGB]{\colorfirst}0.70 & \cellcolor[RGB]{\colorsecond}0.38 & \cellcolor[RGB]{\colorfirst}0.37 & \cellcolor[RGB]{\colorfirst}0.92 & \cellcolor[RGB]{\colorsecond}0.77 & \cellcolor[RGB]{\colorfirst}0.67 & \cellcolor[RGB]{\colorfirst}1.14 & \cellcolor[RGB]{\colorfirst}1.24 & \cellcolor[RGB]{\colorfirst}0.64 & \cellcolor[RGB]{\colorfirst}0.62 & \cellcolor[RGB]{\colorfirst}0.78 & \cellcolor[RGB]{\colorfirst}0.37 & \cellcolor[RGB]{\colorfirst}0.59 & \cellcolor[RGB]{\colorfirst}0.45 & \cellcolor[RGB]{\colorfirst}0.67 \\
% \hline
\bottomrule
\end{tabular}
\caption{\textbf{Quantitative comparison on the DTU Dataset.} Our method enhances GOF and achieves state-of-the-art performance.} 
\vspace{-2.0em}
\label{tab:dtu_result}
\end{table*}

\section{Limitation}
Our method introduces additional computational overhead, primarily from two sources: (1) the neural network used to encode per-primitive texture fields, and (2) the extra computation required for ray-Gaussian intersection. Future work could focus on optimizing these two components to further improve efficiency.

Another limitation of our method lies in its reduced effectiveness on outdoor scenes. While it achieves strong performance on indoor datasets, the improvement is less pronounced on outdoor benchmarks such as Mip-NeRF 360. We attribute this to the limitations of the current bounded tri-plane encoding in capturing complex, large-scale outdoor variations. Future directions may explore more expressive or hierarchical neural encoding schemes to address this challenge.

\end{document}